\pdfoutput=1
\documentclass[namedreferences,oneside,a4paper]{solarphysics}

\usepackage{amsmath}
\usepackage[hyperref,optionalrh]{spr-sola-addons} % For Solar Physics 
\usepackage{color}           % For color text: \color command
\usepackage{breakurl}        % For breaking URLs easily trough lines
   
\usepackage{varwidth}% define the fonts for the URLs
\usepackage{emptypage}
\usepackage{geometry}
\usepackage{adjustbox}
\usepackage{boldline}
\usepackage{caption}
\usepackage{pifont}
\usepackage{enumitem}
\usepackage{threeparttable}
\usepackage{placeins}
\usepackage{float}
\usepackage{lipsum}
\usepackage{fancyhdr}
\usepackage{xcolor}
\usepackage{soul}
\usepackage{color, colortbl}
\usepackage{comment}
\usepackage{multirow}% http://ctan.org/pkg/hhline

\newcommand{\cmark}{\ding{52}}%
\newcommand{\xmark}{\ding{54}}%

\newcommand{\circles}{\ding{108}}%
\definecolor{Gray}{gray}{0.9}
	
\definecolor{LightCyan}{rgb}{0.88,1,1}
\chardef\us=`\_
\DeclareUnicodeCharacter{2212}{\textendash}
\begin{document}
\begin{article}
\begin{opening}

\title{A comprehensive review of Binary Neural Network}

\author[addressref={aff1},email={cyuan1@gradcenter.cuny.edu}]{\inits{Chunyu }\fnm{Chunyu}~\lnm{Yuan}}
%\orcid{123-456-7890}}

\author[addressref={aff1,aff2},email={sos.agaian@csi.cuny.edu}]{\inits{Sos S.}\fnm{Sos S.}~\lnm{Agaian}}
%\author[addressref=aff2,corref,email={e-mail.c@mail.com}]{\inits{F.}\fnm{First~Names}~\lnm{Last~Name~Author-c}\orcid{987-654-3210}}
%\author[addressref=aff3]{\inits{T.}\fnm{First~Names}~\lnm{Last~Name~Author-d}}
%\author{\inits{}\fnm{}~\lnm{}\orcid{}}
%   NOTE:  Just one corresponding author [corref]
%   \institute{$^{1}$ First affiliation
%                     email: \url{e.mail-a} email: \url{e.mail-b}\\ 
%              $^{2}$ Second affiliation
%                     email: \url{e.mail-c} \\
%             \textit{}
\address[id=aff1]{
The Graduate Center, City University of New York}
\address[id=aff2]{
College of Staten Island, City University of New York}
%\address[id=aff2]{Institution, City, State, Country}
%\address[id=aff3]{Third affiliation and address}

\runningauthor{Author-a et al.}
\runningtitle{\textit{}} % Example Article

\begin{abstract}
\normalsize
\\
%Since the first proposal of Binary Neural Network(BNN) was %published in 2016, the BNN techniques have drawn increasing %research interests because of their capability for deploying %models on resource-limited devices. Different from traditional %Convolution Neural Network(CNNs), binarization of activations and %weights can largely save the expensive storage. And, applying %bits' XNOR and popcount operations to do matrix multiplication %can extremely decrease the cost of computation. However, due to %binarization unavoidably causes severe information loss, there %still exists the accuracy performance gap compared to the full %precision CNNs. To address this issue, recently, a variety of %algorithms and solutions have been proposed, and the satisfying %progress is continuously achieved. Based on the research articles %published in peer-review journals and conference from 1996 to %till date, this paper trends to present a systematic review of %BNN including optimized methods to BNN, the diverse %applications using BNN and the open-source frameworks for BNN %model design and development. Moreover, this paper finally %discusses the unresolved challenges and future research %opportunities. 

Deep learning (DL) has recently changed the development of intelligent systems and is widely adopted in many real-life applications. Despite their various benefits and potentials, there is a high demand for DL processing in different computationally limited and energy-constrained devices. It is natural to study game-changing technologies such as Binary  Neural Networks (BNN) to increase deep learning capabilities. Recently remarkable progress has been made in BNN since they can be implemented and embedded on tiny restricted devices and save a significant amount of storage, computation cost, and energy consumption. However, nearly all BNN acts trade with extra memory,  computation cost, and higher performance. This article provides a complete overview of recent developments in BNN. This article focuses exclusively on 1-bit activations and weights 1-bit convolution networks, contrary to previous surveys in which low-bit works are mixed in. It conducted a complete investigation of BNN's development -from their predecessors to the latest BNN algorithms/techniques, presenting a broad design pipeline and discussing each module's variants. Along the way, it examines BNN (a) purpose: their early successes and challenges; (b) BNN optimization: selected representative works that contain essential optimization techniques; (c) deployment: open-source frameworks for BNN modeling and development; (d) terminal: efficient computing architectures and devices for BNN and (e) applications: diverse applications with BNN. Moreover, this paper discusses potential directions and future research opportunities in each section.
\end{abstract}

\keywords{Binary Neural Network, Convolution Neural Network, Model compression and acceleration, Binarization, Quantization }
\end{opening}

\section{Introduction}
     \label{S-Introduction} 
  \subsection{Background}
Artificial intelligence (AI) means the simulation of human intelligence in machines. Because of increased volume of big data, continually advanced algorithms, and incessant improvements in hardware, AI is growing to be  one of the most popular topics in today's world. In recent years, AI holds tremendous promise to power nearly all aspects of society. In AI community, convolution neural networks (CNN) is one common method to solve vision problems such as image classification\citep{lu2007survey, nath2014survey, li2018deep, deepa2011survey, wang2019development}, object detection\citep{zou2019object, liu2020deep, borji2019salient, shantaiya2013survey, jiao2019survey} and object recognition\citep{sukanya2016survey, goyal2014object, jafri2014computer, campbell2001survey, zhou2021rgb}.

 \begin{figure}[!h]    %%%%%%%%%%%%%%%%%% FIGURE 1 
 
   \centering{\includegraphics[width=0.6\textwidth,clip=]{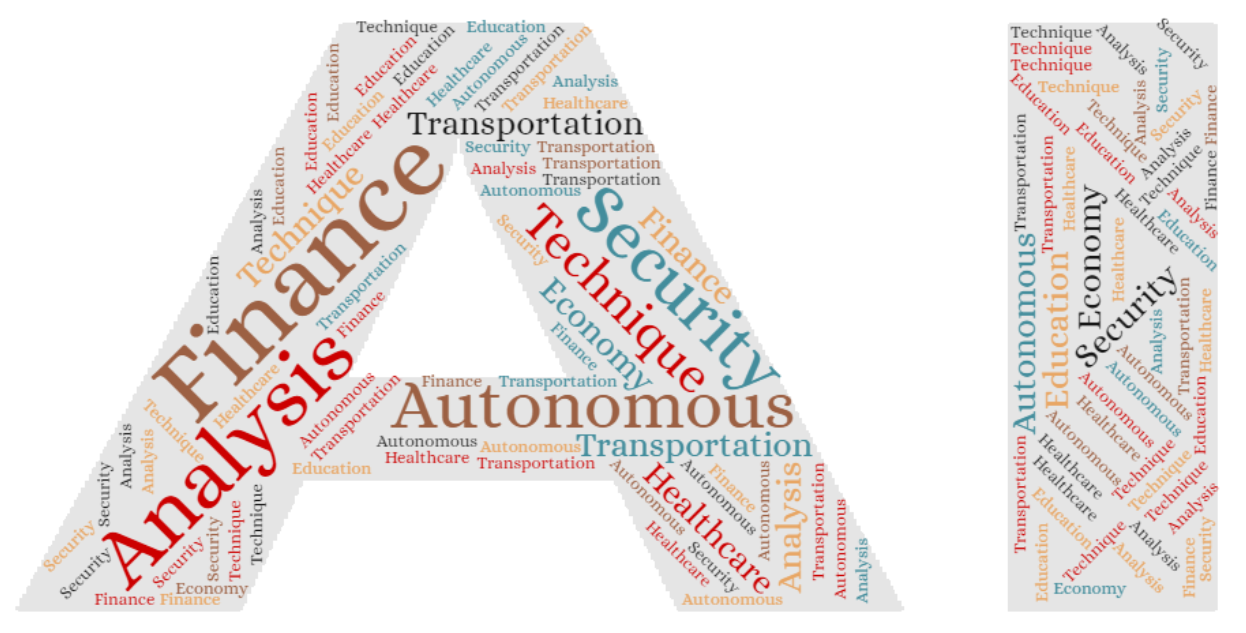}
              }
              \vspace{0pt}
\captionsetup{justification=raggedright,singlelinecheck=false}
   \caption{   Popular topics using AI}
    \label{F-appendix}
    \vspace{-2pt}
  \end{figure}

% \begin{figure}[!h]    %%%%%%%%%%%%%%%%%% FIGURE 1 
%   \centerline{\includegraphics[width=0.5\textwidth,clip=]{survey1.PNG%}
%              }
%   \caption{Popular applications using CNNs}
%    \label{F-appendix}
%  \end{figure}
 
 Although new CNN models were continually presented and advanced, such as ResNeXt\citep{xie2017aggregated}, SE-Net\citep{hu2018squeeze} and SK-Net\citep{li2019selective}, the CNN architectures don't change too much compared to that before 2017. And as CNN models become larger so that they require more computational power and storage, they cannot be equipped on resource-constraint platforms such as smart phones and tiny Internet of Things (IoT) devices. Therefore, it is reasonable to generate a new open problem, which is how to develop more compact, lightweight and efficient power networks which can simultaneously maintain acceptable accuracy. So that trained models can be effectively  utilized  on devices  that  billions  of  customers  use  in  their  everyday  lives.  
 %%\begin{figure}    %%%%%%%%%%%%%%%%%% FIGURE 1 
 %%  \centerline{\includegraphics[width=1\textwidth,clip=]{cnnde%%velop.PNG}
 %%             }
 %%  \caption{Milestones for CNNs}
 %%   \label{F-appendix}
 %% \end{figure}\\
  %%

\indent Model Compression and Acceleration for deep neural networks is one type of solution to solve the problem mentioned above, trying to save the memory and reduce computational costs of CNN while still offering similar capabilities of full-precision CNN models. Based on solutions’ properties, this type can be subdivided into five major categories: parameters quantization, parameters pruning, low-rank matrix factorization, transferred/compact convolutional filters and knowledge distillation.
  \begin{figure}[!h]    %%%%%%%%%%%%%%%%%% FIGURE 2 
   \centerline{\includegraphics[width=1\textwidth,clip=]{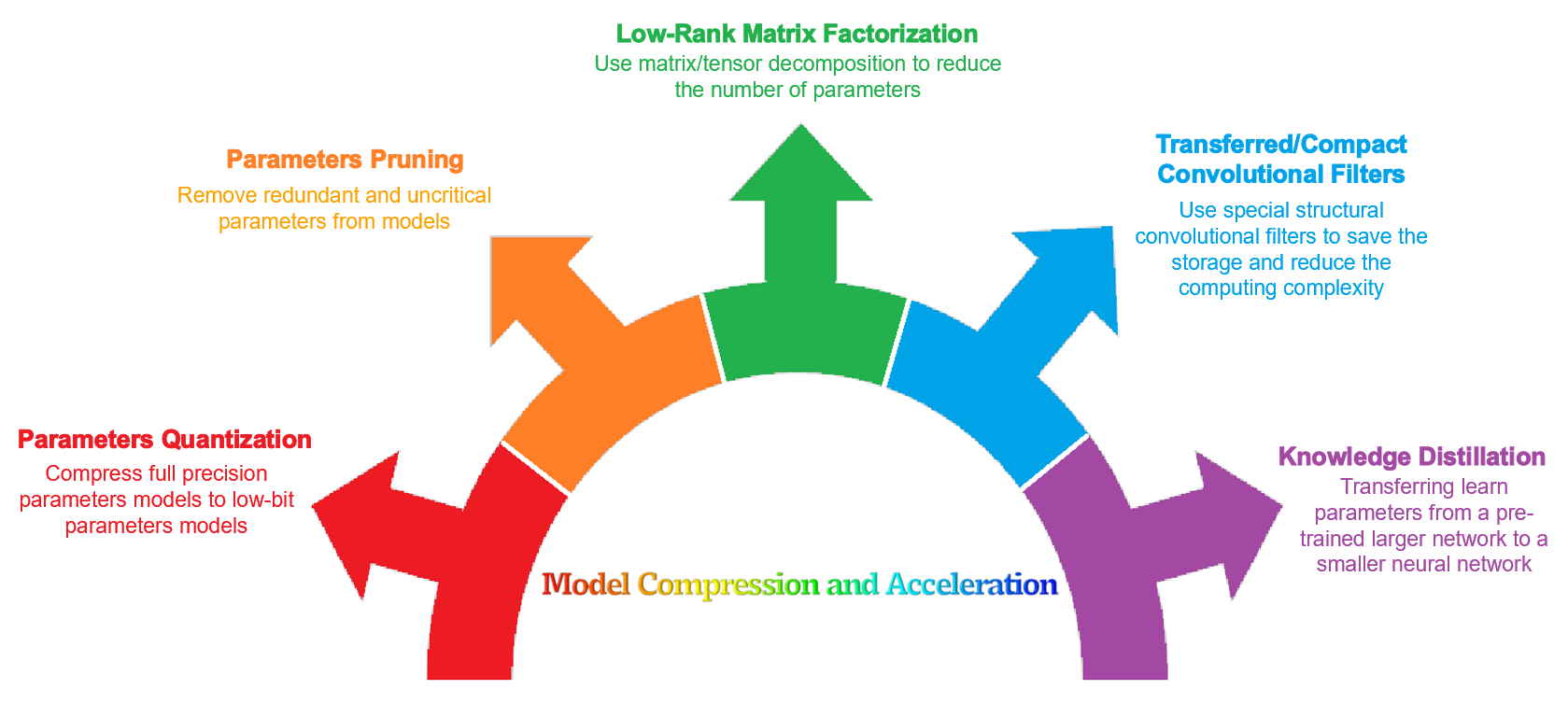}
              }
   
   \caption{   Topics of Model Compression and Acceleration}
    \label{F-appendix}
  \end{figure}
 % Parameters quantization means to compress  full  precision  parameters  models  to  low-bits parameters  models.  Parameters pruning points to removing redundant and uncritical parameters from models. Low-rank matrix factorization tries to use matrix/tensor decomposition to reduce the number of parameters.Transferred/compact convolutional filters use special structural convolutional filters to save the storage and  reduce  the  computation  complexity.  The  knowledge  distillation is  the  target smaller neural network learning parameters from a pre-trained larger network. 
  
  \vspace{-0.5cm}
  \subsection{Motivation}
   It is known that current deep neural network models are computationally expensive and use memory resources extensively. For example, \citep{krizhevsky2012imagenet} designed AlexNet, which contains 60 million float-point parameters and 650,000 neurons, on the ImagetNet dataset. \citep{simonyan2014very}'s model VGG-16, when applied to the same dataset, has over 130 million float-point parameters. Large-weight models are difficult to deploy on tiny devices with limited resources. It is natural to study game-changing technologies such as BNN to increase deep learning capabilities. BNN method is an extreme case of parameters quantization methods. Completing activations and weights to 1-bit values can theoretically have 32 times lower memory storage and 58 times faster inference speed than traditional 32-bit CNN. BNN can be used in a variety of problems including classification \citep{chen2021bnn, qin2020bipointnet}, pattern recognition \citep{qiao2020stbnn}, computer vision \citep{frickenstein2020binary}, natural language
processing (NLP) \citep{xiang2017binary,qian2019binary,9516334}, etc. Because of BNN's incredible advantages with fewer size parameters and faster inference speed can be easily applied and embedded on resource-limited devices such as wearable devices and tiny sensors. In recent years, with the increasing trend on lightweight and practical networks, more and more researchers are turning their attention to BNN. In 2021, a workshop spotlights BNN called binary networks for computer vision held by Computer Vision and Pattern Recognition (CVPR). BNN has become one popular and important research topic in the AI community.

    \subsection{Related Work}
  
   There are a few published surveys on BNN that are \citep{simons2019review} and \citep{qin2020binary}. However, by carefully checking those surveys, we find some of the previous representative BNN techniques were not reviewed and discussed. Even worse, some benchmark results in the literature were wrongly collected and cited. For example, the results on the dataset COCO-2017 \citep{caesar2018cvpr} from a 4-bit quantization network called  FQN \citep{li2019fully} were mistaken as the BNN in \citep{qin2020binary}. Efficiency indicators measured on FPGA were incorrectly cited in both of \citep{simons2019review} and \citep{qin2020binary}.  Besides, in the most recent year, there is a great number of new BNN methods published. Some of them have already been crucially improved BNN performance and have generated new directions that were not included and discussed in those previous survey works. For those reasons, we present a new extensive review of BNN which covers all the BNN design pipeline topics including algorithms, deployment, and applications. Especially, different from prior surveys that mixed low-bit networks reviews,  we only focus on reviewing the pure and truthful BNN where has 1-bit activations and weights in the convolution. We accurately summarize the BNN's major optimization techniques and subdivided them into five categories to discuss. More importantly,  we noticed that each previous work may contain several optimization techniques. So we don't simply relegate each work to one category. More carefully, we utilize tables to list all the associated works with their contributions in that category. To the best of our ability, we collect all previous BNN works which were published in realizable conferences and journals till the date. We believe this work can serve as educational materials on the topic of BNN, and as reference information for professional researchers in this field.

     \begin{figure}[!h]   %%%%%%%%%%%%%%%%%% FIGURE 2 
   \centering{\includegraphics[width=0.7\textwidth,clip=]{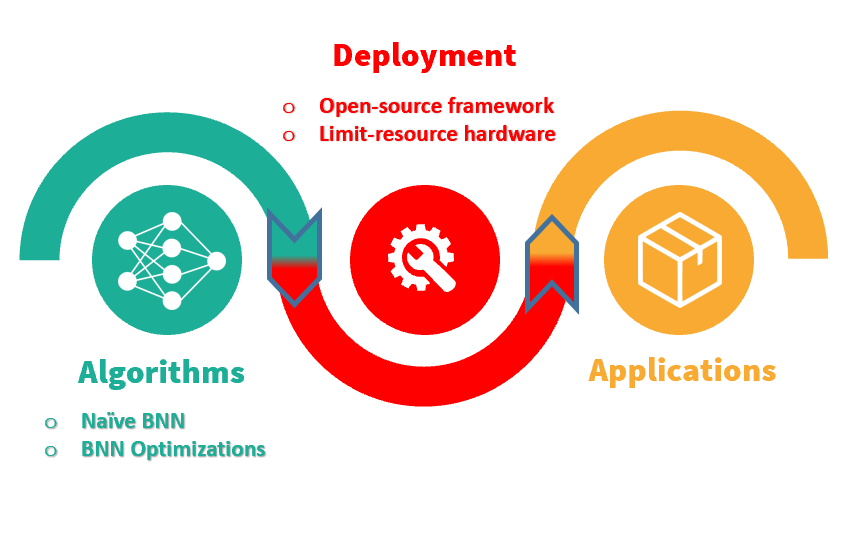}
              }
              \vspace{-0.4cm}
               \captionsetup{justification=centering}
   \caption{   Topics covered in this survey paper}
 
  \end{figure}

    \vspace{-0.4cm}
   \subsection{Organization}
     %(1)quantization error minimization,(2) network loss function improvement,(3) gradient er-ror decrease,(4) network topology and(5) training strategy and tricks.  
The organization of this work is structured as follows. Section 2 introduces the basic principles of BNN and their early successes and challenges. Section 3 mainly reviews the optimization methods for BNN based on selected representative works that contain vital optimization techniques. Section 4 reviews the open-source frameworks for the BNN  modeling. Section 5 introduces popular efficiency hardware platforms and their definitions of common terms. Section 6 presents the recent BNN applications, including the associated tables with the performance progress history. Section 7 is our conclusion and summary.

\vspace{-0.4cm}

%%%%%%%%%%%%%%%%%%%%%%%%%%%%%%%%%%%%%%%%%%%%%%%%%%%%%%%%%%%%%%%%%%%%%%%%%%%%

\section{Binary Neural Network}
\subsection{What is BNN?}
\label{S-BNN}
 \begin{figure}[!h]    %%%%%%%%%%%%%%%%%% FIGURE 2 
   \centerline{\includegraphics[width=0.8\textwidth,clip=]{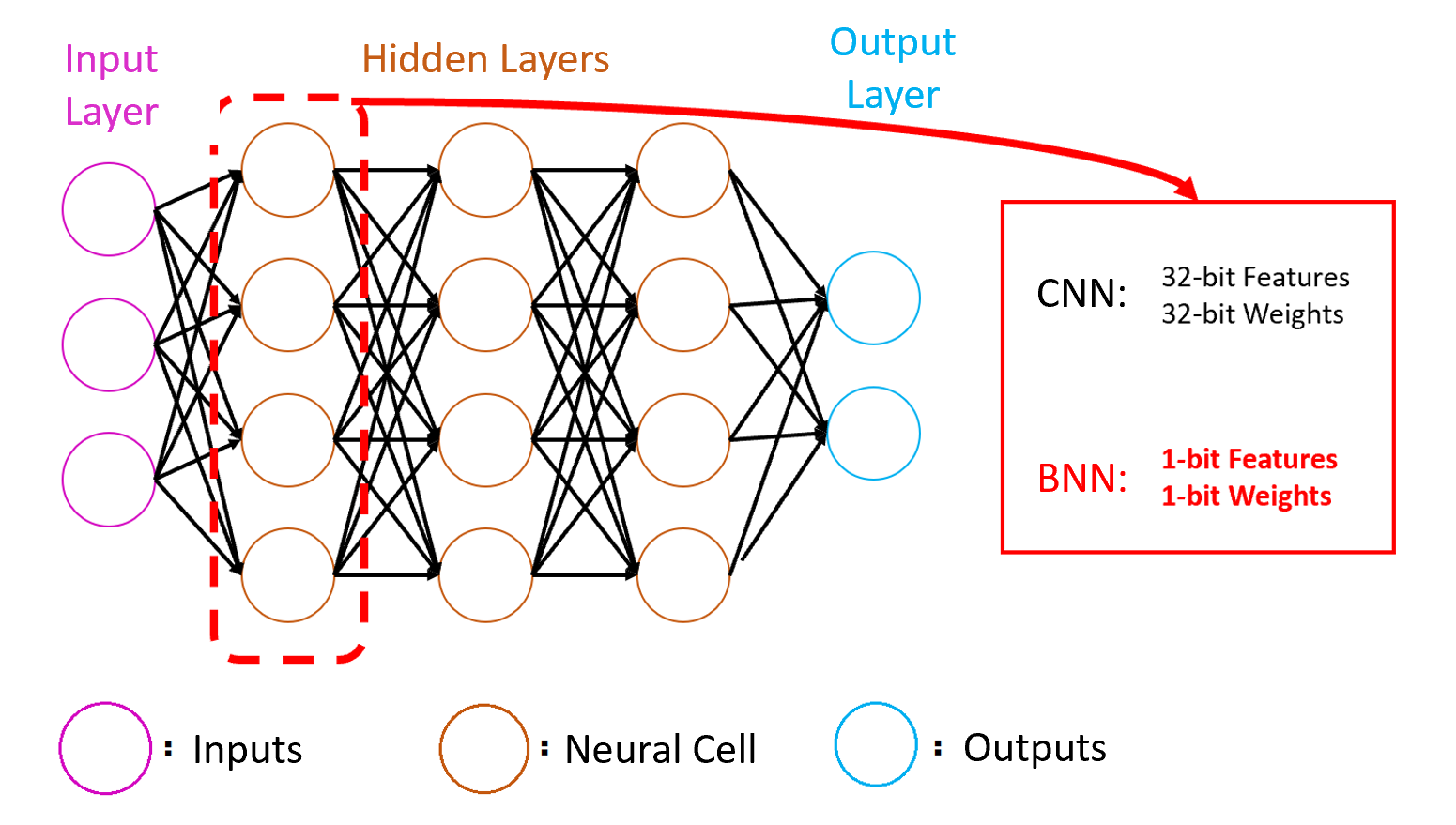}
              }
            
   \caption{   An artificial neural network }
    \label{F-appendix}
    
  \end{figure}
\noindent BNN is a type of neural network that activations(or called features) and weights are 1-bit values in all the hidden layers (except the input and output layers). In a few words, BNN is an extremely compacted case of CNN. Because BNN and CNN have the same structures except for the different precision activations and weights. BNN also specifically refers to BNN techniques that compact 32-bit activations and weights to 1-bit values. The process of compacting 32-bit to 1-bit values is binarization. The purpose of binarization not only can save the expensive model's storage, but also reduce the matrix computation costs by using XNOR and popcount operations.  \citep{rastegari2016xnor} reported that BNN can have 32 times lower memory saving and 58 times faster convolution operations than 32-bit CNN. In traditional CNN, the vast majority of computing cost spent in matrix multiplication inside the convolution operation. The basic convolution operation without bias can be expressed as:
\begin{equation}
    {Z = I * W}
\end{equation}
\noindent
where I and W represent activations and weights, respectively, Z is the output of the convolution operation with matrix multiplication. Because in such a multiplication operation, it contains a large of floating-point operations, including floating-point multiplication and floating-point addition, which is the reason for low-speed performance in the neural network inference. To resolve this issue, \citet{courbariaux2016binarized} and \citet{kim2016bitwise} separately proposed their vanilla BNN architectures.

\begin{table}[!h]

\label{T-simple}
\caption{Naive BNN 
}
    \begin{threeparttable}

\begin{tabular}{ll}
 %\hline
 %\multicolumn{4}{|c|}{Country List} \\
\hlineB{2.5} 
  \rowcolor{Gray}
 BNN Name &Key Techniques  \\
\hlineB{2.5}

 \rowcolor{LightCyan}
Binarized Neural Networks \citep{courbariaux2016binarized}  & \vtop{\hbox{\strut FP: sign(x)}\hbox{\strut BP: clip(x,-1,1)=max(-1,min(1,x))}} \\

Bitwise Neural Networks \citep{kim2016bitwise} &  \vtop{\hbox{\strut FP: sign(x)}\hbox{\strut BP: two steps,}\hbox {\strut ~~~~~~similar to clip(x,-1,1)=max(-1,min(1,x))}}  \\

 \hline

\end{tabular}

   \begin{tablenotes}
          %\footnotesize   %% If you want them smaller like foot notes
          \item Note: \textbf{FP}: Forward Propagation, \textbf{BP}: Backward Propagation .
        \end{tablenotes}
    \end{threeparttable}
\end{table}
\vspace{-0.6cm}
An artificial neural network consists of two processes: forward propagation and backward propagation. Forward propagation is the process of moving from the input layer (left) to the output layer (right) in the figure. 4, which also refers model inference. Backward propagation is the process of moving from the output layer (right) to the input layer (left) in the figure. 4, which represent the process of fine-tuning the model's weights. Subsection 2.2 and 2.3 discuss how BNN works in forward propagation and backward propagation.

\subsection{Forward Propagation}

\noindent The neural cell is a basic computation structure in the forward path of a neural network.  Different from the 32-bit CNN, neural cell in the BNN's forward path adds the binarization steps to the activations I and weights W before convolution operation. The binarization steps' purpose is to represent the floating-point activations and weights using 1-bit. Figure. 5 presents the difference in computation steps inside a neural cell along the forward path between naive BNN and 32-bit CNN.

 \begin{figure}[!h]    %%%%%%%%%%%%%%%%%% FIGURE 4
   \centerline{\includegraphics[width=0.7\textwidth,clip=]{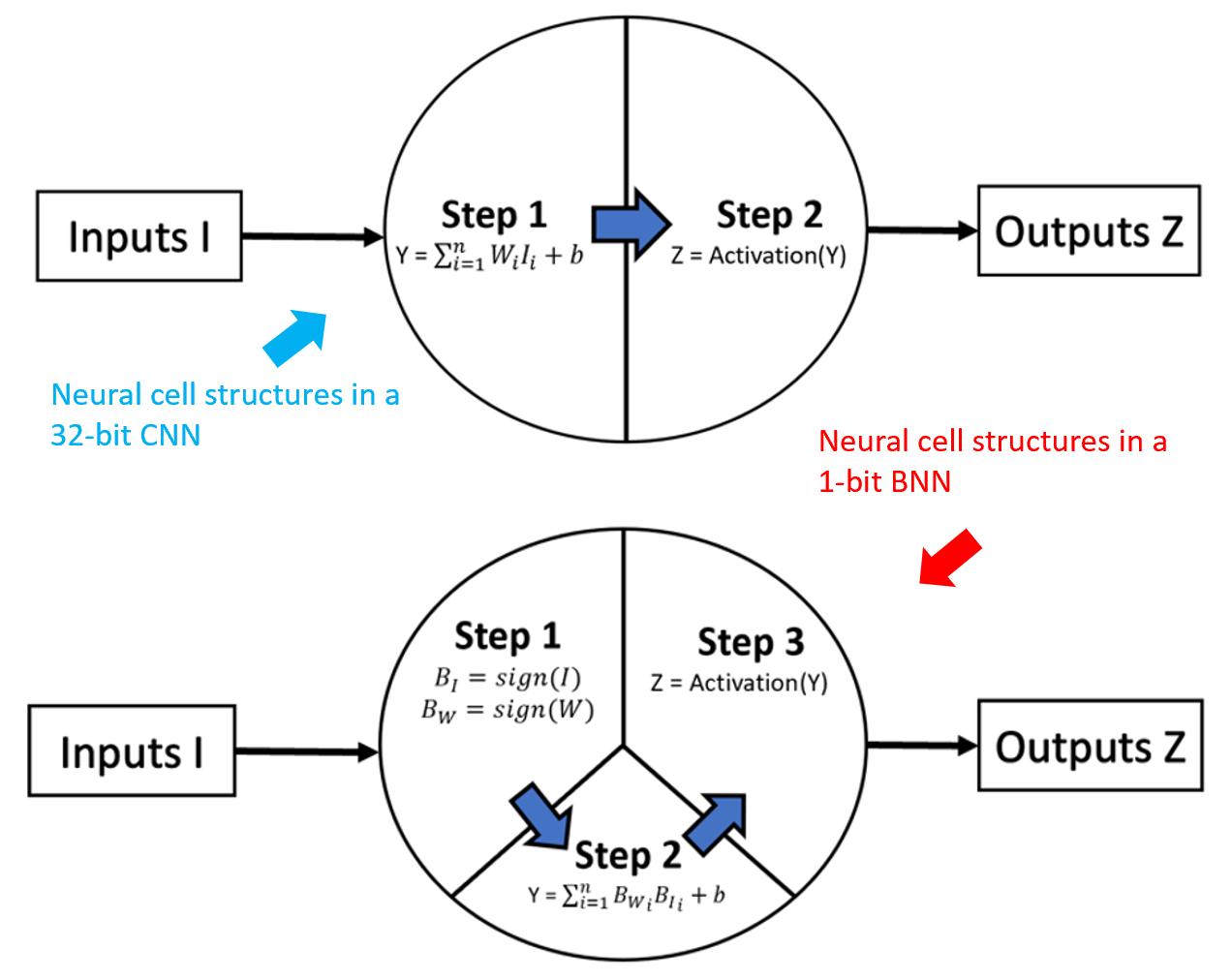}
              }
   \caption{   Neural cell structures of native BNN  and 32-bit CNN}
    \label{F-appendix}
  \end{figure}
 The sign function is widely used for binarization:
\begin{equation}
     {Sign(x) = } {\begin{cases}
  +1, & if ~x\geqslant 0,\\
    -1,& otherwise.\\
\end{cases}}
\end{equation}

\indent After binarization, activations I and weights W will be:
\begin{equation}
\begin{array}{ll}
        { I \approx sign(I)=B_{I}} 
   
\end{array}
\end{equation}
\begin{equation}
\begin{array}{ll}

        { W \approx sign(W)=  B_{W}} \\ 
\end{array}
\end{equation}

where $B_{I}$ and $B_{W}$ are binary activations and binary weights, respectively.

\noindent

\begin{table}[!h]
 \caption{\label{tab:table-name}BNN XNOR Operations}
\begin{tabular}{ccc}
 %\hline
 %\multicolumn{4}{|c|}{Country List} \\
\hlineB{2.5} 
  \rowcolor{Gray}
 Binary Activations & Binary Weights &XNOR Result  \\
\hlineB{2.5}
 \rowcolor{LightCyan}
 -1(0)    & -1(0)  & +1(1)  \\
-1(0) &+1(1)  &-1(0)  \\
 \rowcolor{LightCyan}
+1(1) & -1(0) & -1(0) 
\\
+1(1)   &+1(1)&+1(1)\\

 \hline

\end{tabular}

\end{table}
\vspace{-0.6cm}

Because $B_{I}$ and $B_{W}$ values are $\{+1,-1\}$ which has the similar XNOR results compared to $\{0, +1\}$ as table 2 shown. Then, we can use bitwise XNOR and popcount to replace expansive matrix multiplication calculation. Figure. 6 presents the examples of convolution computation processes in native BNN and 32-bit CNN.\\

 \begin{figure}[!h]    %%%%%%%%%%%%%%%%%% FIGURE 4
   \centerline{\includegraphics[width=0.8\textwidth,clip=]{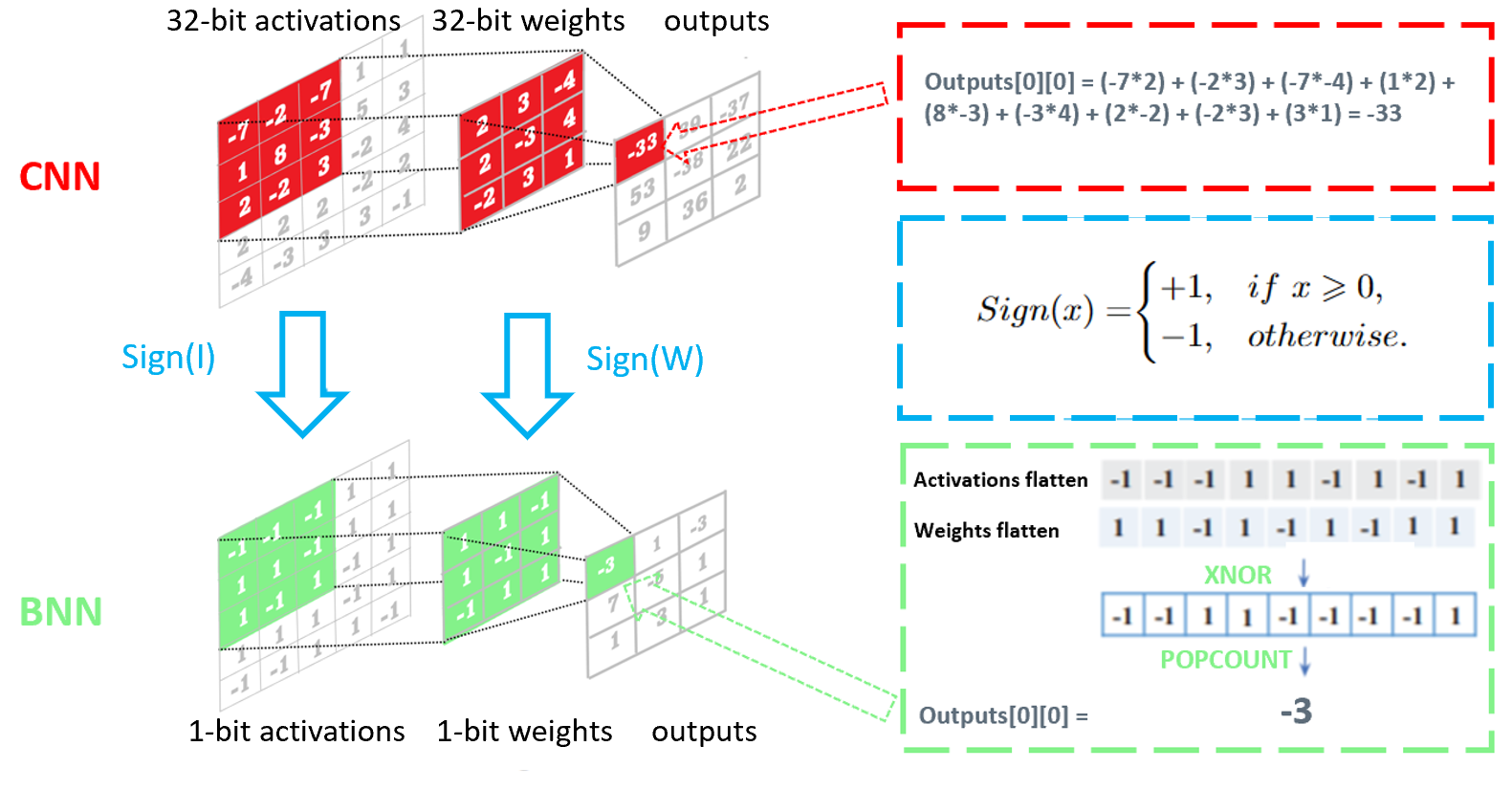}
              }
   \caption{     Naive BNN's forward propagation compared to CNN's}
    \label{F-appendix}
  \end{figure}

  \begin{figure}[h!]    %%%%%%%%%%%%%%%%%% FIGURE 5
   \centerline{\includegraphics[width=0.8\textwidth,clip=]{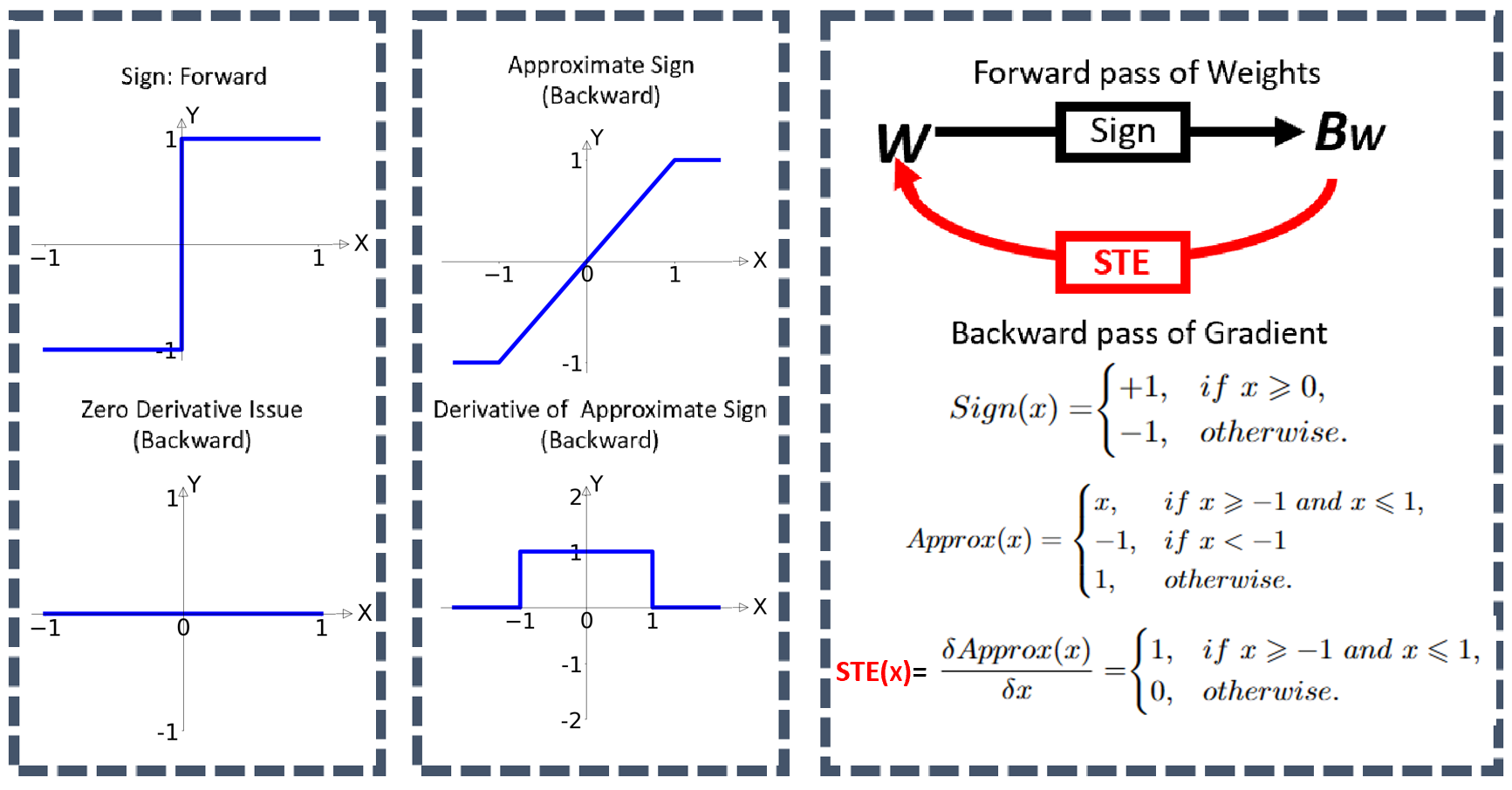}
              }
   \caption{   Binarized Neural Networks' Backward propagation}
    \label{F-appendix}
  
  \end{figure}
  
\subsection{Backward Propagation} 
Because the derivation result of binarization function (sign) is 0. Binary weights cannot be learned with the traditional gradient descent method based on a backward propagation algorithm. To resolve this issue,  Binarized-Neural-Networks \citep{courbariaux2016binarized} apply the technique called straight-through estimator (STE) \citep{tieleman2012lecture, bengio2013estimating} to learn binary weights in backward propagation.  Figure 7 explains the process of learning Binarization weights in Binarized-Neural-Networks. During the BNN training steps, each layer's real weights are kept and updated using STE. After training, binarized weights are saved and the real weights are discarded. Besides, Bitwise-Neural-Networks \citep{kim2016bitwise} contains two steps to train the BNN model. The first step is to train some compressed network parameters in real-value networks with weight compression. Then, the authors initialize the real-valued parameters for the target bitwise neural network, and adopt a training strategy that is similar to STE.

 \subsection{Summary} 
Although naive BNN has faster inference speeds and smaller weight sizes, the accuracy performance is much lower than that using full-precision CNN in the early stages. The reason is the severe information loss due to parameter binarization, including binary activations and binary weights. To address the above issue, a variety of novel optimization solutions have been proposed in recent years. In the next section, we regulate these methods into categories.

%%%%%%%%%%%%%%%%%%%%%%%%%%%%%%%%%%%%%%%%%%%%%%%%%%%%%%%%%%%%%%%%%%%%%%%%%%%%
\section{Binary Neural Network Optimization}
 To report as the latest solutions at the published time, each BNN model contained several optimization and improvement points/methods. We regulate these enhancement methods to 5 categories as the figure 8 shown: (1) quantization error minimization, (2) loss function improvement, (3) gradient approximation, (4) network topology structure and (5) training strategy and tricks. 
\begin{figure}[!h]
    \centering
  
     \includegraphics[scale=0.7]{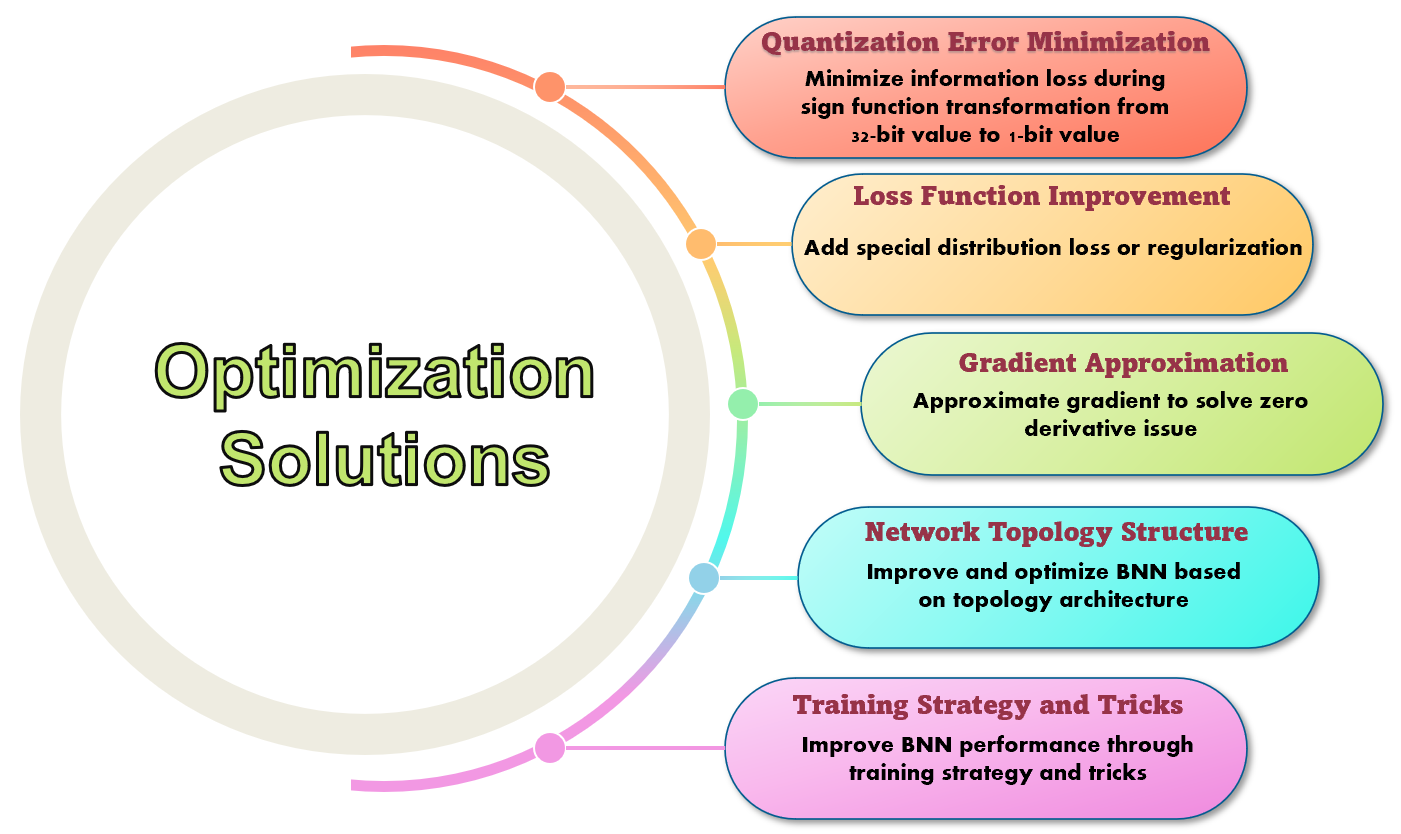}
    \caption{    BNN enhancement methods}
    \label{fig}
\end{figure}
\vspace{-0.5cm}
\subsection{Quantization Error Minimization}

\subsubsection{Scaling Factor}

To reduce the information loss during sign function transformation from 32-bit value to 1-bit value, XNOR-Net \citep{rastegari2016xnor} adds channel-wise scaling factors $\alpha$ and $\beta$ for activations and weights. Therefore, equations 3 and 4 can be changed to:
\begin{equation}
     I \approx \alpha * sign(I) = \alpha * B_{I}
\end{equation}
\vspace{-0.8cm}
\begin{equation}
     {W \approx  \beta * sign(W) = \beta * B_{W} }
\end{equation}
where $\alpha$ and $\beta$ are :
\begin{equation}
     {\alpha = \frac{1}{n} {\| I\|_{L_1}}}
\end{equation}
\vspace{-0.3cm}
\begin{equation}
     {\beta = \frac{1}{n} {\| W\|_{L_1}}}
\end{equation}
\noindent
Therefore, equation 1 can be changed:
\begin{equation}
     {Z = I * W \approx(\alpha * B_{I}) *(\beta * B_{W}) =(\alpha * \beta)*( B_{I} \circledast B_{W})}
\end{equation}

\noindent Hadamard matrices \citep{agaian1986hadamard} have the same properties as binarized matrices in which all the values are +1 or -1. Advancing on top of XNOR-Net, HadaNet \citep{akhauri2019hadanets} applies the concept of  hadamard transforms \citep{hadabook} to binary activations and weights without increasing filter map counts.
Also, XNOR-Net++ \citep{bulat2019xnor} proposes to merge the activation and weight scaling factors into a single one, and explore various ways to construct the shape of scaling factor based on input, output, channel and their combinations.  Later, \citet{zhao2021data} design DA-BNN which is a data-adaptive method that can generate an adaptive amplitude based on spatial and channel attention.

  \begin{figure}[h!]    %%%%%%%%%%%%%%%%%% FIGURE 5
   {
   \centerline{\includegraphics[width=1\textwidth,clip=]{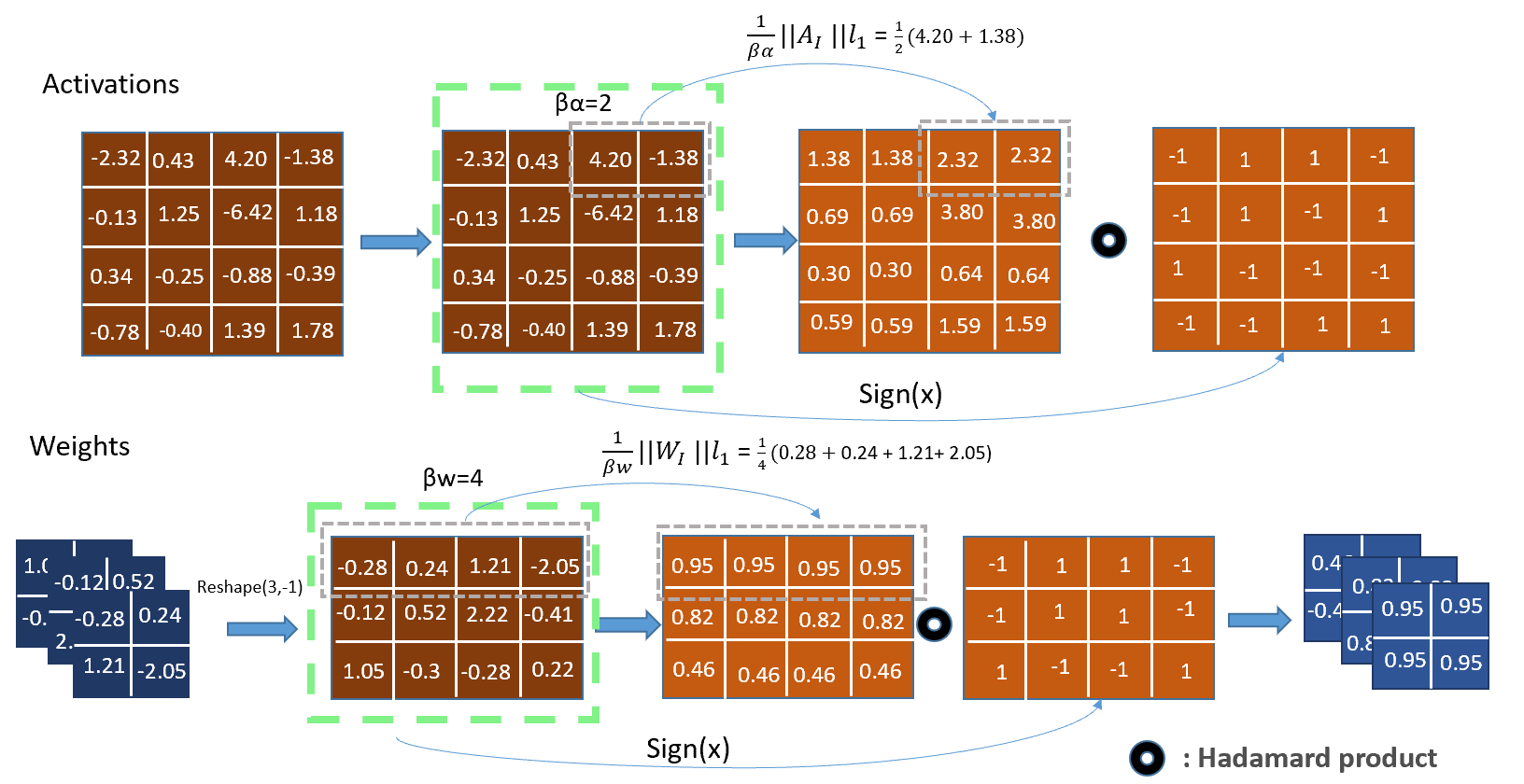}
              }
   \caption{   Binarization process in HadaNet}
    \label{F-appendix}
    }
  \end{figure}

%%%%%%%%%%%%%%%%%%%%%%%%%%%%%%%%%%%%%%%%%%%%%%%%%%%%%%%%%%%%%%%%%%%%%%%%%%%

\subsubsection{Quantization Function}

Different using sign functions to do activation and weight binarization, several prior works present new methods to binary the parameters to [-1, +1]. DoReFa-Net \citep{zhou2016dorefa}, UniQ \citep{pham2021training}, Quantization-Networks \citep{yang2019quantization} and DSQ \citep{gong2019differentiable} propose the k-bit method for parameter quantization including binarizaiton. Their 1-bit methods provide a different way to binary parameters compared to that used sign function. SI-BNN \citep{wang2020sparsity} proposes and finds that binary activations to [0, +1] and binary weights to [-1, +1] can alleviate information loss. Rectified clamp unit (ReCU) \citep{xu2021recu} is a weights standardization method to  reveal the inherent contradiction between minimizing the quantization error and maximizing the information entropy in BNN. SiMaN\citep{lin2022siman} proposed an angle alignment objective,sign-to-magnitude,to constrain the weight binarization to \{0, +1\}. AdaBin\citep{tu2022adabin} proposed an optimal binary sets of weights and activations for each layer.

\begin{table}[!h]
   \begin{threeparttable}
 \caption{\label{tab:table-name} Summary Table for Quantization Error Minimization }

\begin{tabular}{lllc}

\hlineB{2.5} 
  \rowcolor{Gray}
 BNN Name &Code Available&Key Idea&Idea Category \\
\hlineB{2.5}
\rowcolor{LightCyan}
 XNOR-NET\citeyearpar{rastegari2016xnor}&\cmark~(O,UO)&\vtop{\hbox{\strut Channel-wise scaling factor $\alpha$ and $\beta$ for }{\hbox{\strut ~activations and weights}}}&$\blacktriangle$ \\

 LAB2\citeyearpar{hou2016loss}&\cmark~(O)&\vtop{\hbox{\strut Minize the weight binarization loss through}{\hbox{\strut  ~the proximal Newton algorithm }{\hbox{\strut  ~with diagonal Hessian approximation}}}}&\circles\\
 
 \rowcolor{LightCyan}
 DoReFa-Net\citeyearpar{zhou2016dorefa}&\cmark~(O,UO)&\vtop{\hbox{\strut New quantization method to get binary or }{\hbox{\strut ~low bitwidth weights and activations }{\hbox{\strut ~using low bitwidth parameter gradients}}}}&$\blacksquare$ \\
 
 HORQ\citeyearpar{li2017performance}&\xmark~&\vtop{\hbox{\strut Order-Two Residual Quantization to alleviate}{\hbox{\strut ~information loss}}}&\circles\\
\rowcolor{LightCyan} 
HadaNet\citep{akhauri2019hadanets}&\cmark~(O)&Apply hadamard product to binarization &$\blacktriangle$ \\

XNOR-Net++\citeyearpar{bulat2019xnor} & \xmark~&Several ways to construct the scale factors &$\blacktriangle$ \\

\rowcolor{LightCyan}
\vtop{\hbox{\strut Quantization Networks}{\hbox{\strut\citeyearpar{yang2019quantization}}}}& \cmark~(O)&\vtop{\hbox{\strut Soft quantization function: formulate quantization}{\hbox{\strut ~as a differentiable non-linear mapping function}{\hbox{\strut ~based on sigmoid}}}}&$\blacksquare$\\

 \hline

\end{tabular}
   \begin{tablenotes}
          %\footnotesize   %% If you want them smaller like foot notes
   
            \item Note: O: Official implementation, UO: Un-official implementation, $\blacktriangle$: Scaling Factor,~$\blacksquare$: Quantization Function,  \circles:  Activations/Weights distribution and Others 
        \end{tablenotes}
    \end{threeparttable}
\vspace{-0.9cm}
\end{table}

\begin{table}[!h]
   \begin{threeparttable}
 \caption{\label{tab:table-name}     Summary Table for Quantization Error Minimization(Continue table 3) }

\begin{tabular}{lllc}

\hlineB{2.5} 

  \rowcolor{Gray}
 BNN Name &Code Available&Key Idea&Idea Category \\
\hlineB{2.5}

CI-BCNN\citeyearpar{wang2019learning}&\xmark~&\vtop{\hbox{\strut Interacted to alleviate xnor and popcount }\vtop{\hbox{\strut ~quantization error via channel-wise interaction}\vtop{\hbox{\strut ~by a reinforcement graph model}}}}&\circles \\

\rowcolor{LightCyan}
DSQ\citeyearpar{gong2019differentiable}&\cmark~(O)&\vtop{\hbox{\strut Soft quantization function: formulate quantization}{\hbox{\strut ~as a differentiable non-linear mapping function}{\hbox{\strut ~based on tanh}}}}&$\blacksquare$\\

IR-Net\citeyearpar{qin2020forward}&\cmark~(O)&\vtop{\hbox{\strut
Libra Parameter Binarization: a implicit rectifier} {\hbox{\strut ~that reshapes the data distribution} {\hbox{\strut ~before binarization}}}}&\circles\\

\rowcolor{LightCyan}
BBG-Net\citeyearpar{shen2020balanced}&\xmark~&Maximizing entropy with balanced binary weights&\circles\\

SI-BNN\citeyearpar{wang2020sparsity}&\xmark~&\vtop{\hbox{\strut Binary activations to 0 or +1 and binary weights}{\hbox{\strut ~to -1 or +1 to remains most information}}}&$\blacksquare$\\

\rowcolor{LightCyan}
LNS\citeyearpar{han2020training}&\xmark~& Binary weights mapping with noisy supervision&\circles\\

SLB\citeyearpar{NEURIPS2020_2a084e55}&\xmark~&\vtop{\hbox{\strut State batch normalization and low-bit}{\hbox{\strut ~search including binary}}} &\circles\\

\rowcolor{LightCyan}
ProxyBNN\citeyearpar{he2020proxybnn}&\xmark~ &\vtop{\hbox{\strut Learning orthogonal matrix basis coefficients to}{\hbox{\strut  ~and construct the pre-binarization weights}}}&\circles\\

RBNN\citeyearpar{lin2020rotated}& \cmark~(O)&\vtop{\hbox{\strut Rotate the full precision weight vector to its binary} {\hbox{\strut ~vector to reduce the angular bias}}}&\circles\\

\rowcolor{LightCyan}
UniQ\citeyearpar{pham2021training}&\cmark~(O)&Symmetric quantizer with
a trainable step size&$\blacksquare$\\

IA-BNN\citeyearpar{kim2020improving} &\xmark~&\vtop{\hbox{\strut Unbalanced Distribution of binary activations actually }{\hbox{\strut  ~improves the accuracy of BNN by shifting the }{\hbox{\strut  ~trainable thresholds of binary activations}}}}&\circles\\

\rowcolor{LightCyan}
DA-BNN \citeyearpar{zhao2021data}&\xmark~&Data-adaptive re-scaling&$\blacktriangle$\\

ReCU\citeyearpar{xu2021recu}&\cmark~(O)&Weights standardization&\circles\\

\rowcolor{LightCyan}
   Bi-half\citeyearpar{li2022equal}&\cmark~(O)&   \vtop{\hbox{\strut
Optimal Transport: optimizes the weight binarization}{\hbox{\strut ~by aligning a real-valued proxy weight distributions} {\hbox{\strut ~with an idealized distribution}}} }&\circles\\
   SiMaN\citeyearpar{lin2022siman}& \cmark~(O)&\vtop{\hbox{\strut     Sign-To-Magnitude:  constrain the weight
binarization }{\hbox{\strut ~   to \{0, +1\}}}}&$\blacksquare$\\

\rowcolor{LightCyan}
   AdaBin\citeyearpar{tu2022adabin}&\cmark~(O)&   \vtop{\hbox{\strut
 Optimal binary sets} }&$\blacksquare$\\

 \hline

\end{tabular}
   \begin{tablenotes}
          %\footnotesize   %% If you want them smaller like foot notes
   
            \item Note: O: Official implementation, UO: Un-official implementation, $\blacktriangle$: Scaling Factor,~$\blacksquare$: Quantization Function,  \circles:  Activations/Weights distribution and Others 
        \end{tablenotes}
    \end{threeparttable}
\vspace{-0.5cm}
\end{table}

\subsubsection{Activations/Weights distribution and Others}
Different to directly optimization the binarization process in the convolution layer, IR-Net \citep{qin2020forward}, BBG-Net \citep{shen2020balanced}, SLB \citep{NEURIPS2020_2a084e55}, RBNN \citep{lin2020rotated} , IA-BNN \citep{kim2020improving} and Bi-half \citep{li2022equal} optimize and reshape and activations and weights distribution before binarization in their models.  LAB2 \citep{hou2016loss} applies the proximal Newton algorithm to binary weights by directly considering the binarization loss. HORQ \citep{li2017performance} proposes to use recursive binary quantization to lighten information loss. CI-BCNN \citep{wang2019learning}, via learning reinforcement graph model, mines the channel-wise interactions to iterate popcount and reduce inconsistency of signs in binary feature maps and preserves the information of input samples. LNS \citep{han2020training} proposes to train binarization function to predict binarization weights via supervision noise learning. ProxyBNN \citep{he2020proxybnn} constructs the pre-binarization weights matrix using the basis and coordinates submatrix to reduce information loss after binarization.

\subsection{Loss Function Improvement}

To close the accuracy gap from real-valued networks, How-to-Train \citep{tang2017train}, Binarized- Convolutional \citep{bulat2017binarized}, BNN-RBNT \citep{darabi2018regularized}, PCNN \citep{gu2019projection}, BNN-DL \citep{ding2019regularizing}, CCNN \citep{xu2019accurate}, BONN \citep{gu2019bayesian}, RBNN \citep{ lin2020rotated} and LCR\citep{shang2022lipschitz} propose adding distribution loss or special regularization to the overall loss function. Their basic types can be expressed as : 
\begin{equation}
 {\L_{T} = \L_{S} + \lambda \L_{DR}}
\end{equation}
\noindent
where $L_{T}$ is total loss, $L_{S}$ is a cross-entropy loss, $L_{DR}$ is the added special distribution loss or regularization and $\lambda$ is a balancing hyper-parameter. LNS \citep{han2020training}, Real-to-Bin \citep{martinez2020training} and ReActNet \citep{liu2020reactnet} proposes special loss functions for servicing transfer learning strategy. Table 5 is a summary table for  BNN that proposed or used techniques for BNN loss function.

\vspace{0.2cm}
\begin{table}[h!]
  \begin{threeparttable}
 \caption{\label{tab:table-name}Summary table for BNN that proposed or used techniques for BNN loss function }

\begin{tabular}{lllc}

\hlineB{2.5} 
  \rowcolor{Gray}
 BNN Name &Code Available&Key Idea&Idea Category  \\

 \hlineB{2.5}
 
 \rowcolor{LightCyan}
How to Train\citeyearpar{tang2017train}& \xmark~&\vtop{\hbox{\strut New regularization to replace L2 regularization}{\hbox{\strut ~(Euclidean)}}}&$\blacktriangle$\\ 
\vtop{\hbox{\strut Binarized Convolutional}{\hbox{\strut\citeyearpar{bulat2017binarized}}}} &\cmark~(O)&Sigmoid cross-entropy pixel-wise loss function&$\blacktriangle$ \\

\rowcolor{LightCyan}
BNN-RBNT\citeyearpar{darabi2018regularized} &\xmark~&\vtop{\hbox{\strut Add regularization functions(Manhattan }{\hbox{\strut ~or Euclidean) to the overall loss function}}}&$\blacktriangle$
\\

PCNN\citeyearpar{gu2019projection} &\cmark~(O)&\vtop{\hbox{\strut Add projection loss to jointly learned with }{\hbox{\strut ~the conventional cross-entropy loss}}}&$\blacktriangle$ \\
 
\rowcolor{LightCyan}
BNN-DL\citeyearpar{ding2019regularizing}&\cmark~(O)&Add distribution loss to the overall loss function &$\blacktriangle$
\\

CCNN\citeyearpar{xu2019accurate} &\xmark~ &\vtop{\hbox{\strut L2 regularization term acting on the weight  }{\hbox{\strut ~scaling factors}}}&$\blacktriangle$\\

\rowcolor{LightCyan}
BONN\citeyearpar{gu2019bayesian} &\cmark~(O)&\vtop{\hbox{\strut Add Bayesian
kernel loss and Bayesian feature }{\hbox{\strut~loss to the overall loss function }}}&$\blacktriangle$
\\

RBNN\citeyearpar{ lin2020rotated}&\xmark~&\vtop{\hbox{\strut Add kernel approximation and adversarial }{\hbox{\strut ~learning loss to the overall loss function}}}&$\blacktriangle$\\

%\hline
%BATS\cite{bulat2020bats}&No & new regularization method that helps stabilizing the
%BNN architecture  search process \\

%\hline 
%BNAS\cite{kim2020learning} & Yes(O)&  diversity regularizer for BNN architecture search \\

\rowcolor{LightCyan}
LNS\citeyearpar{han2020training}&\xmark~ &Unbiased auxiliary loss for binary weights mapping&$\blacksquare$ \\

Real-to-Bin\citeyearpar{martinez2020training} &\cmark~(UO)&\vtop{\hbox{\strut Standard logit matching loss for attention}{\hbox{\strut ~transfer between BNN and real-valued networks}}}&$\blacksquare$\\

\rowcolor{LightCyan}
ReActNet\citeyearpar{liu2020reactnet} &\cmark~(O)&\vtop{\hbox{\strut Distributional Loss to learn similar between BNN}{\hbox{\strut ~and real-valued networks}}}&$\blacksquare$\\ 

   LCR\citeyearpar{shang2022lipschitz}&   \cmark~(O)&   \vtop{\hbox{\strut Retain the Lipschitz constant
serving }{\hbox{\strut ~as a regularization term}}}&   $\blacktriangle$\\

 \hline

\end{tabular}
   \begin{tablenotes}
          %\footnotesize   %% If you want them smaller like foot notes
   
            \item Note: O: official implementation, UO: Un-official implementation, $\blacktriangle$: Non-transferring Learning,~$\blacksquare$: Transferring Learning
        \end{tablenotes}
    \end{threeparttable}
    \vspace*{-15pt}
\end{table}
 \vspace{0.2cm}
\subsection{Gradient Approximation}
 \vspace{0.1cm}
As the derivative result of sign function equals to zero, it leads weights fail to get updated in the back-propagation. Straight-through estimator (STE) is one available method to approximate sign gradients. However, using STE fails to learn weights near the
borders of −1 and +1, that greatly harms the updating ability of back propagation. GB-Net \citep{sakr2018true} uses true gradient-based learning to train BNN with parametrized clipping functions (PCF) and replace PCF by scaled binary activation function (SBAF) to obtain final BNN interface. BNN-RBNT \citep{darabi2018regularized} proposes a backward approximation based on the sigmoid function. Bi-Real-Net \citep{liu2018bi} proposed a polynomial steps function to approximate forward sign function. CCNN \citep{xu2019accurate} introduces the derivation estimator to approximate their binarization funciton. CBCN \citep{liu2019circulant} designs the gradient approximation based on Gaussian functions. Although the authors of CBCN presented the function's characteristics and displayed the function graph, we don't know the detailed implementation for their function and in their open source code, we find they in effect use Bi-Real-Net's method for gradient approximation. IR-Net \citep{qin2020forward} and RBNN \citep{lin2020rotated} separately design a dynamic gradient estimator which can adjust the gradient approximation during the training process. SI-BNN \citep{wang2020sparsity} designs their gradient estimator with two trainable parameters on the top of STE. BinaryDuo \citep{kim2020binaryduo} Quantitatively analyzed the differentiable approximation function and proposed to use the gradient of smoothed loss function to estimate
the gradient. FDA\citep{xu2021learning} proposed a method that uses the combination of sine funcitons in the Fourier frequency domain to estimate the gradient of sign functions. Table 6 is a summary table for BNN that proposes techniques for gradient approximation. 
\clearpage
\begin{table}[!h]
  \begin{threeparttable}

 \caption{\label{tab:table-name}Summary Table for Gradient Approximation}

\begin{tabular}{lllc}

\hlineB{2.5} 
  \rowcolor{Gray}
 BNN Name &Code Available&Key Idea&Shape Number \\
 \hlineB{2.5}
 \rowcolor{LightCyan}
 GB-Net\citeyearpar{sakr2018true} &\xmark~&\vtop{\hbox{\strut parametrized clipping functions(PCF),}{\hbox{\strut ~scaled binary activation function(SBAF)}}}&3
%$SS_{\beta}(x)=2\sigma(\beta x)[1+\beta x{1-\sigma(\beta x)}]-1$, where $\sigma(z)$: sigmoid function, $\beta$ controls speed of activation asymptote 
\\
BNN-RBNT\citeyearpar{darabi2018regularized} &\xmark~&\vtop{\hbox{\strut SignSwish:a Gradient Approximation based on}{\hbox{\strut ~sigmoid function}}}&4
%$SS_{\beta}(x)=2\sigma(\beta x)[1+\beta x{1-\sigma(\beta x)}]-1$, where $\sigma(z)$: sigmoid function, $\beta$ controls speed of activation asymptote 
\\

\rowcolor{LightCyan}
Bi-Real-Net\citeyearpar{liu2018bi} & \cmark~(O, UO)&\vtop{\hbox{\strut Tight approximation to the derivative of sign}{\hbox{\strut ~function with respect to activations,
magnitude-aware}{\hbox{\strut ~ gradient with respect to weights}}}}&5\\
 
CCNN\citeyearpar{xu2019accurate} &\xmark~&Long-tailed higher-order estimator&6\\

\rowcolor{LightCyan}
CBCN\citeyearpar{liu2019circulant}& \cmark~(O)&Gaussian
function as the approximation of the gradient&-\\
 
IR-Net\citeyearpar{qin2020forward}&\cmark~(O)&\vtop{\hbox{\strut
Error Decay Estimator: a training-aware Gradient}{\hbox{\strut ~Approximation function based on tanh function}}}&7\\

\rowcolor{LightCyan}
SI-BNN\citeyearpar{wang2020sparsity}&\xmark~&Trainable thresholds into  backward propagation&8\\

RBNN\citeyearpar{lin2020rotated}& \cmark~(O)&\vtop{\hbox{\strut Training-aware Gradient Approximation function based}{\hbox{\strut ~on sign function}}}&9\\

\rowcolor{LightCyan}
BinaryDuo\citeyearpar{kim2020binaryduo}&\cmark~(O)&\vtop{\hbox{\strut Quantitatively estimate the gradient mismatch}{\hbox{\strut ~using cosine similarity applying CDG}}}&-\\

   FDA\citeyearpar{xu2021learning}&    \cmark~(O)&   \vtop{\hbox{\strut Decomposing sign with Fourier Series}}&   -\\

 \hline

\end{tabular}

   \begin{tablenotes}
          %\footnotesize   %% If you want them smaller like foot notes
   
               \item Note: O: Official implementation, UO: Un-official implementation, Number: shape plots in Figures 10-12, - means none
        \end{tablenotes}
    \end{threeparttable}
 \vspace*{-15pt}
\end{table}
\begin{figure}[!h]
\vspace{-0.1cm}
    \centering
    \includegraphics[width=0.95\columnwidth]{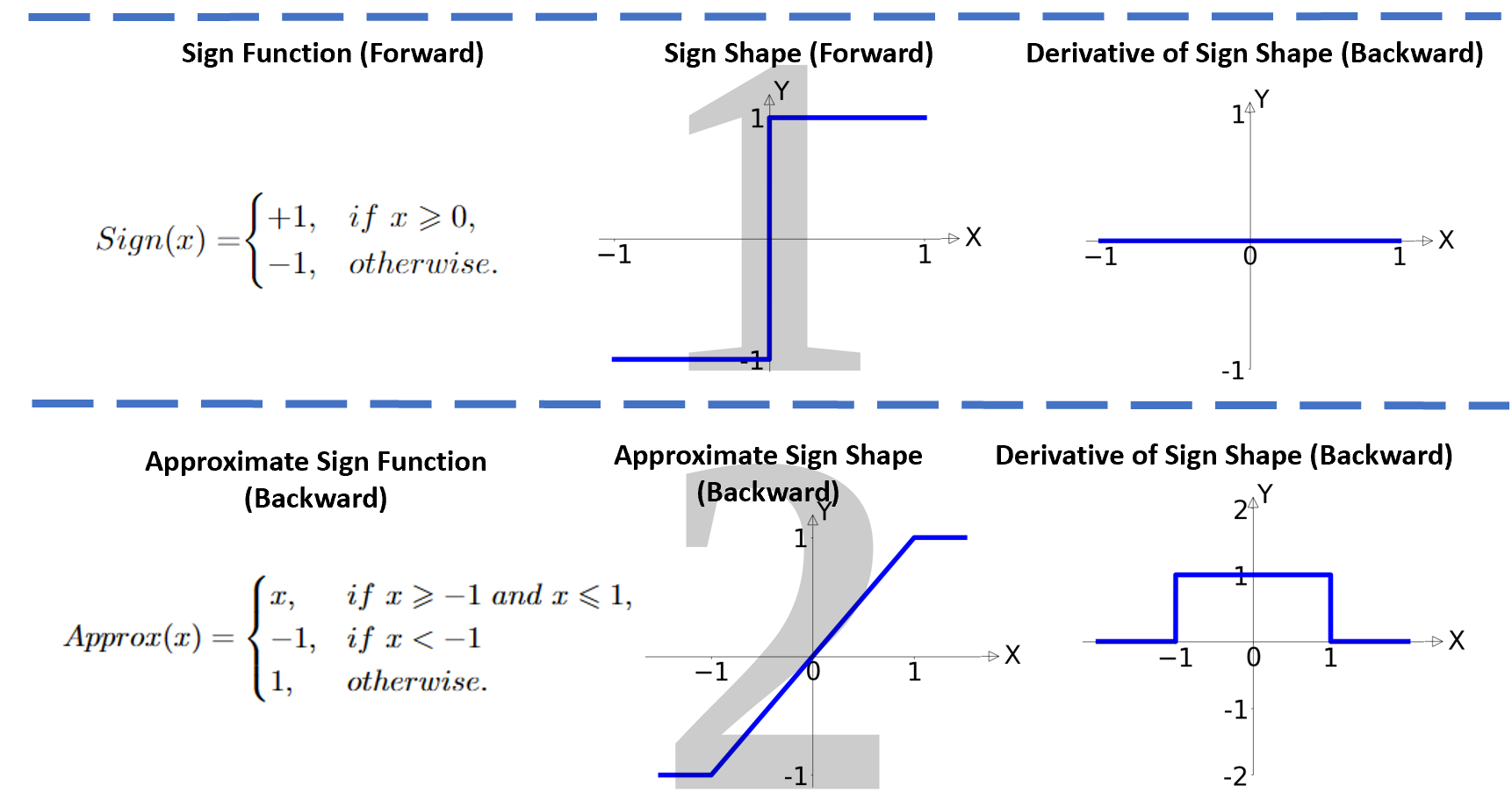}

     \includegraphics[width=0.95\columnwidth]{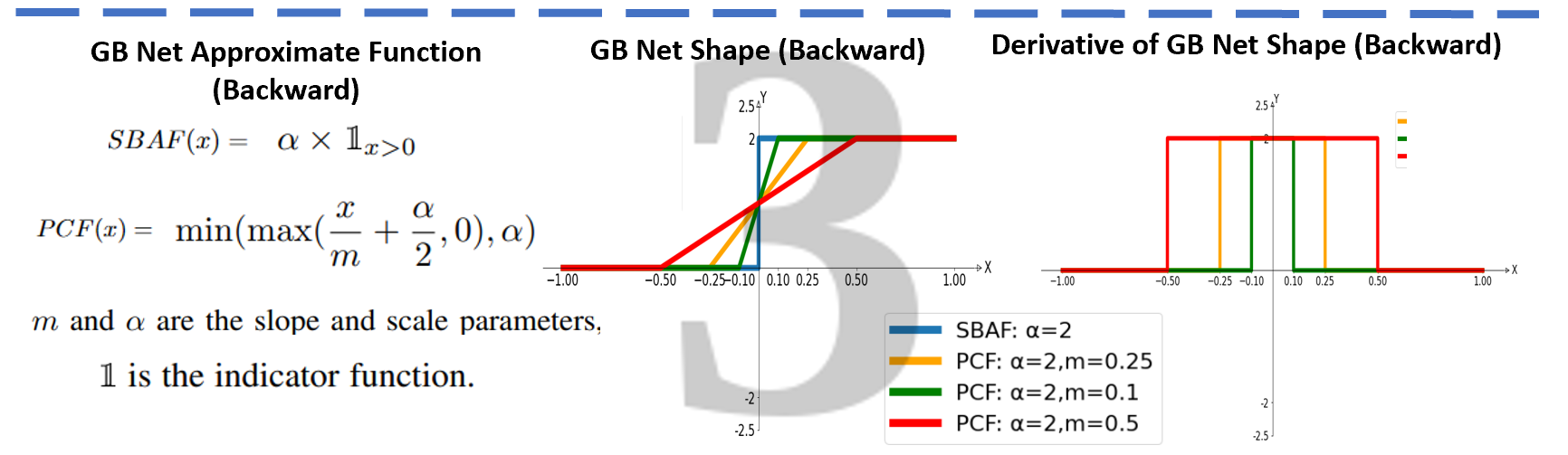}

    \caption{     Shapes of sign or approx-sign functions and their derivatives }
    \label{fig}
    
\end{figure}

\begin{figure}[!h]
\vspace{-0.2cm}
    \centering

     \includegraphics[width=0.95\columnwidth]{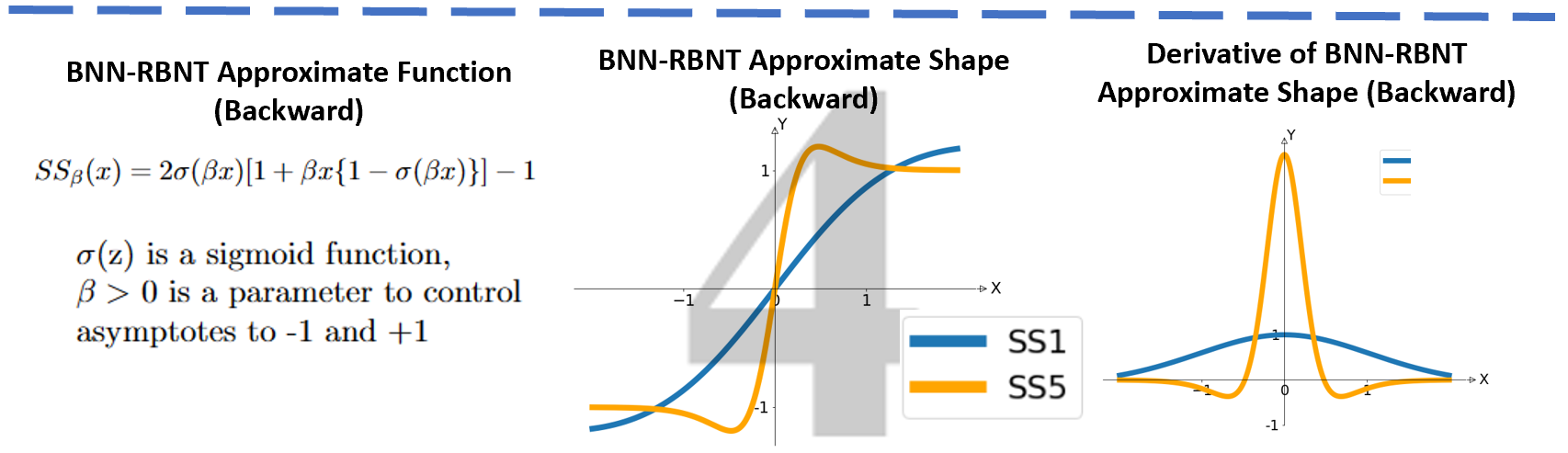}

     \includegraphics[width=0.95\columnwidth]{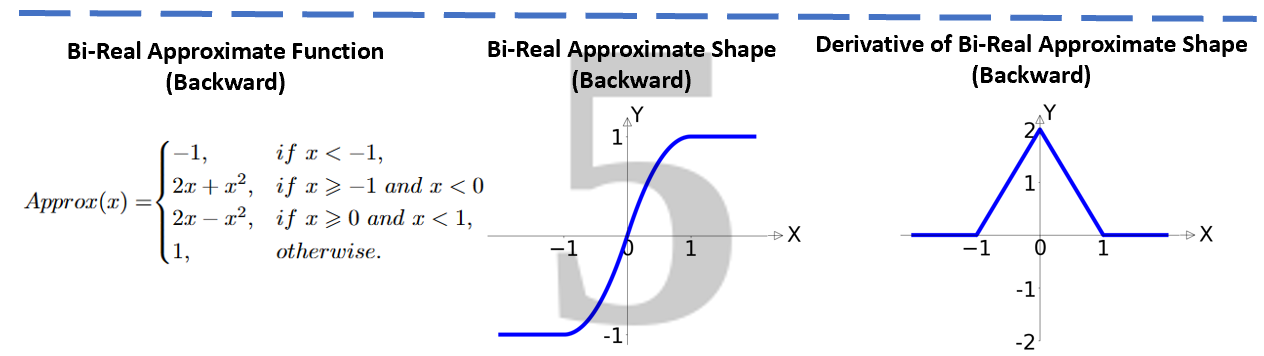}
   
     \includegraphics[width=0.95\columnwidth]{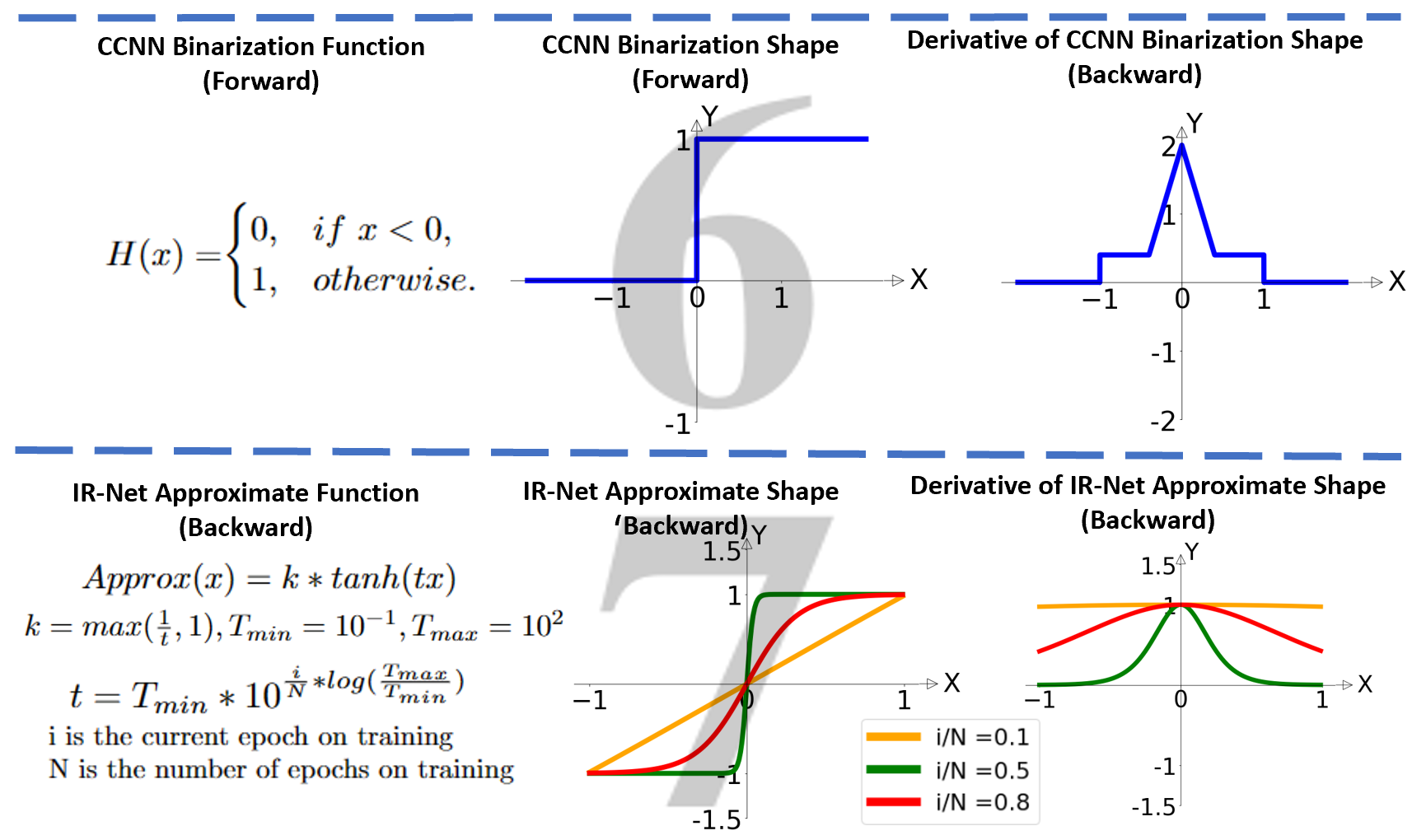}
     
     \includegraphics[width=0.95\columnwidth]{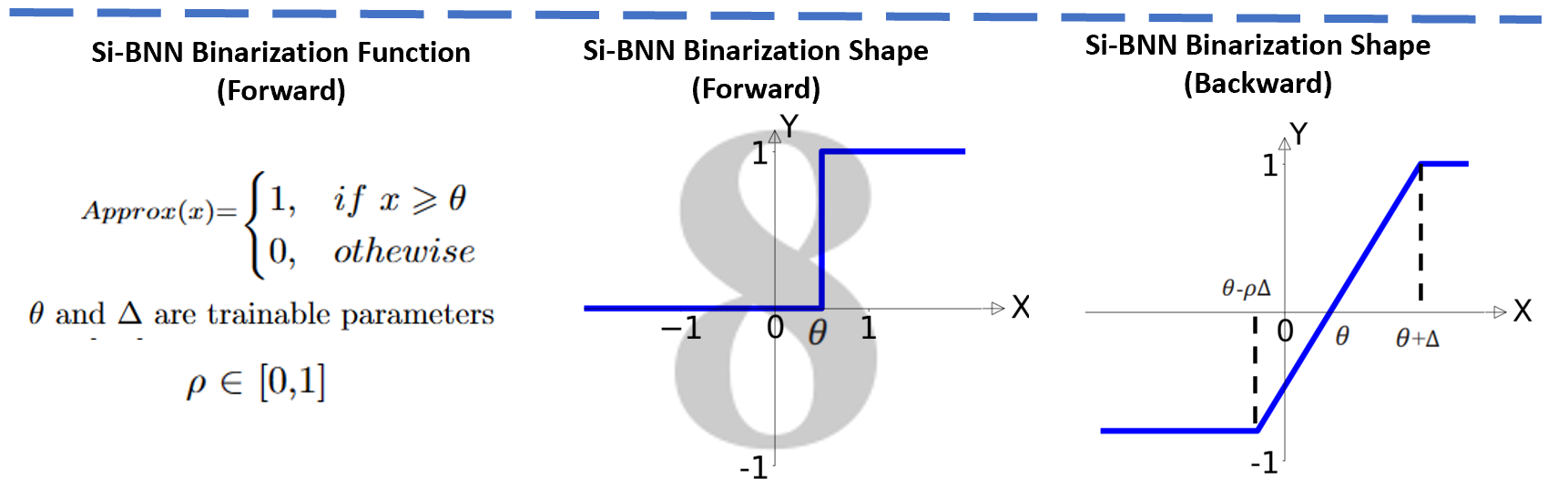}

    \caption{   Shapes of sign or approx-sign functions and their derivatives continue figure 10.}
    \label{fig}
    
\end{figure}

\begin{figure}[!h]

    \centering

     \includegraphics[width=1\columnwidth]{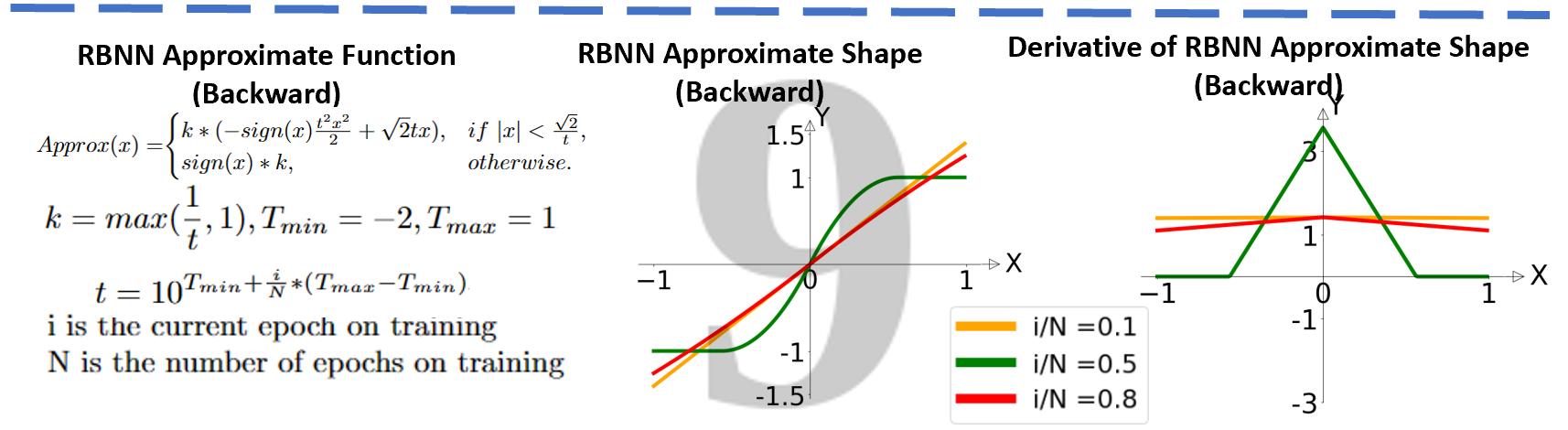}

    \caption{     Shapes of sign or approx-sign functions and their derivatives continue Figure 10. }
    \label{fig}
    
\end{figure}

\subsection{Network Topology Structure}

\noindent Network architectures can affect BNN performance. ABC-Net \citep{lin2017towards}, CBCN \citep{liu2019circulant}, Bi-Real-Net \citep{liu2018bi}, WPRN \citep{mishra2017wrpn}, Group-Net \citep{zhuang2019structured}, BBG-Net \citep{shen2020balanced} and Real-to-Bin \citep{martinez2020training} propose to modified classical network (e.g. ResNet) to improve accuracy performance. BENN \citep{zhu2019binary} proposes to leverage ensemble BNN to improve prediction performance. BinaryDenseNet \citep{Bethge_2019_ICCV} proposes methods and constructs customized BNN special dense network to improve accuracy performance against similar model size BNN. Search Accurate \citep{shen2019searching} and DMS \citep{Li_2020_ICLR_NAS}, via designing search algorithm, adjust the number of channel to close the accuracy gap compared to full-precision network. BATS \citep{bulat2020bats} , BNAS \citep{kim2020learning}, NASB \citep{zhu2020nasb} and High-Capacity-Expert \citep{bulat2020high}, through designed NAS methods, search architectures for BNN to compare accuracy performance with other BNNs which have similar model sizes and binarized from classic network (e.g. ResNet). Especially, High-Capacity-Expert \citep{bulat2020high} first applied condition computing within BNN called expert convolution and combined it with grouped convolution. Inspired by MobileNet-v1, MoBiNet-Mid \citep{phan2020mobinet} and Binarized MobileNet \citep{phan2020binarizing} propose new BNN architecture with high accuracy performance, fewer ops and lighter model size. MeliusNet \citep{bethge2020meliusnet} and ReActNet \citep{liu2020reactnet} designs new BNN architectures which can beat the accuracy rate of full-precision light-weight MobileNet with fewer OPs computation cost. Inspired by BN-free \citep{brock2021high}, BNN-BN-free \citep{chen2021bnn} replaces the batch normalization (BatchNorm) with scaling factor and the ReActNet without BatchNorm still has competitive classification top-1 accuracy on the ImageNet dataset. FracBNN \citep{zhang2021fracbnn} extends ReActNet's topology, re-balances the blocks of networks and designs a two 1-bit activation scheme to improve feature learning. FracBNN has a competitive top-1 prediction result on the ImageNet dataset compared to full precision MobileNet-v2. BCNN \citep{redfern2021bcnn} designs a customized structure for ImageNet classification with lower model size compared to MeliusNet and ReActNet. BiMLP \citep{xu2022bimlp} proposes a  binary architecture of vision Multi-Layer Perceptrons.
    \begin{figure}[!h]    %%%%%%%%%%%%%%%%%% FIGURE 6
   \centerline{\includegraphics[width=0.85\textwidth,clip=]{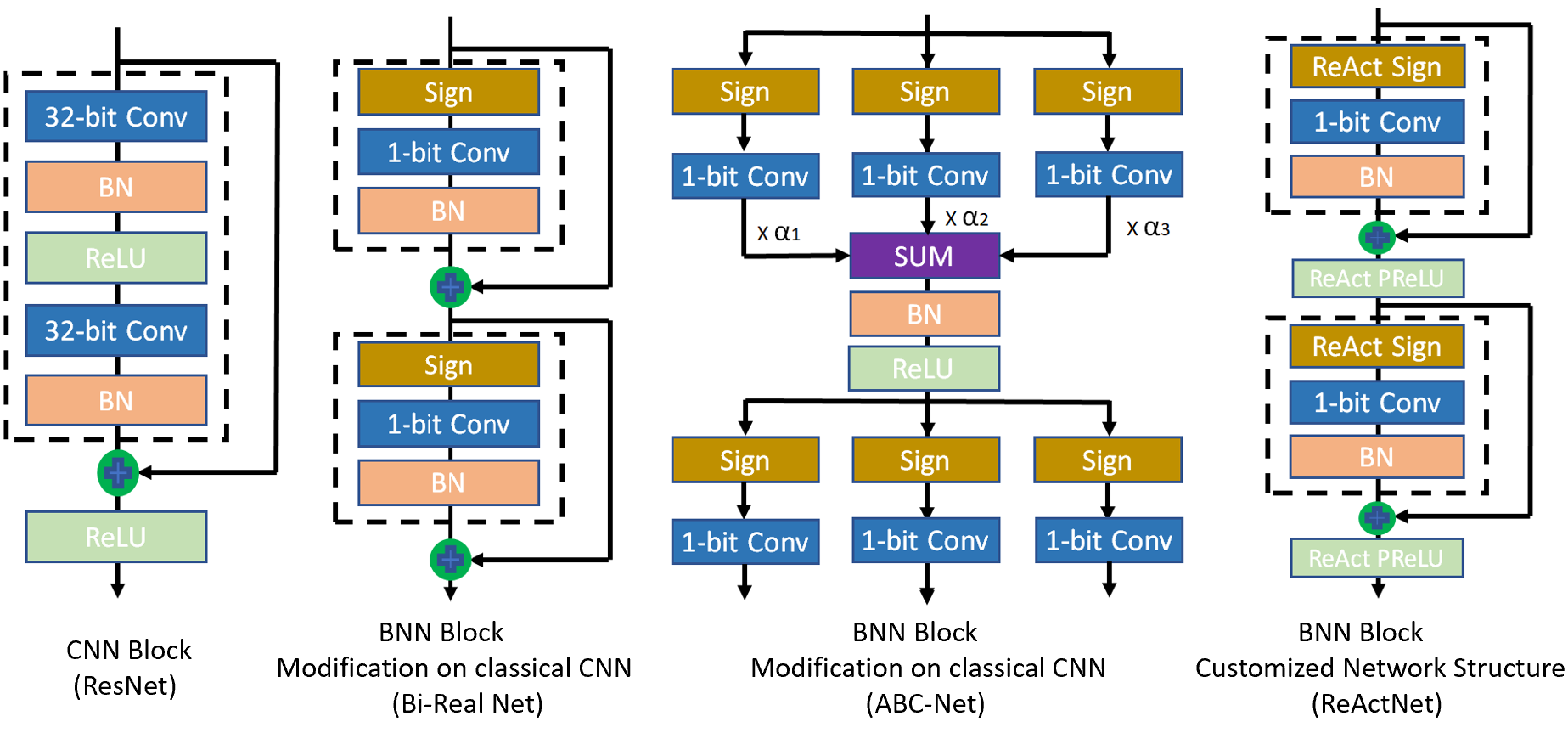}
              }
   \caption{     Representative CNN and BNN block structure}
    \label{F-appendix}
  \end{figure}

\begin{table}[!h]
 \begin{threeparttable}
\centering

\begin{tabular}{lllc}

\hlineB{2.5} 
  \rowcolor{Gray}
 BNN Name &Code Available&Key Idea&Idea Category  \\
 \hlineB{2.5}
 \rowcolor{LightCyan}
ABC-Net\citeyearpar{lin2017towards}&\cmark~(UO)&\vtop{\hbox{\strut Employ multiple binary activations and linear }{\hbox{\strut~combination of
multiple binary weight to alleviate}{\hbox{\strut~single channel information loss}}}}&$\bigstar$,~$\blacktriangleright$\\

WPRN\citeyearpar{mishra2017wrpn}&\xmark&\vtop{\hbox{\strut Increase the number of filters to compensate for}{\hbox{\strut~ information loss}}}&$\bigstar,\blacktriangleleft$\\

\rowcolor{LightCyan}
Bi-Real-Net\citeyearpar{liu2018bi} & \cmark~(O, UO)&Shortcut for one layer per block &$\bigstar,\blacktriangle$\\

CBCN\citeyearpar{liu2019circulant}& \cmark~(O)&\vtop{\hbox{\strut Circulant
filters(CiFs) and circulant binary }{\hbox{\strut~ convolution(CBConv)
to enhance the capacity of  }{\hbox{\strut~binarized convolutional features}}}}&$\bigstar,\blacktriangleleft$\\
\rowcolor{LightCyan}
Group-Net\citeyearpar{zhuang2019structured} &\cmark~(O)&\vtop{\hbox{\strut Group binarization base block to approximate full }{\hbox{\strut~precision network}}}&$\bigstar,\blacktriangleright$ \\

BENN\citeyearpar{zhu2019binary}&\cmark~(O)&\vtop{\hbox{\strut Use ensemble variant bagging to aggregate multiple  }{\hbox{\strut~BNN results}}}&$\bigstar,\blacktriangledown$\\

\rowcolor{LightCyan}
BinaryDenseNet\citeyearpar{Bethge_2019_ICCV} &\cmark~(O)&New BNN efficiency architecture: BinaryDenseNet&$\bigstar,\blacktriangledown$\\

Search Accurate\citeyearpar{shen2019searching}&\xmark &\vtop{\hbox{\strut Evolutionary algorithm to search and adjust the }{\hbox{\strut~number  of channels in each convolutional layer }{\hbox{\strut~after binarization}}}}&\circles,~$\blacksquare$\\

\rowcolor{LightCyan}
DMS\citeyearpar{Li_2020_ICLR_NAS}&\cmark~(O)&\vtop{\hbox{\strut Differentiable dimension search algorithm to search  }{\hbox{\strut~and adjust the number of channels in each  }{\hbox{\strut~ convolutional layer after binarization}}}}&\circles,~$\blacksquare$\\

BATS\citeyearpar{bulat2020bats}&\xmark &space specially BNN architectures search&\circles,~$\blacksquare$ \\

\rowcolor{LightCyan}
  BNAS\citeyearpar{kim2020learning}  &\cmark~(O)&\vtop{\hbox{\strut BNN architectures searches based on the cell based  }{\hbox{\strut~on search methods}}}&\circles,~$\blacksquare$\\ 

NASB\citeyearpar{zhu2020nasb}  &\xmark&\vtop{\hbox{\strut Neural Architecture Search algorithm for optimal}{\hbox{\strut~ BNN architecture}}}&\circles,~$\blacksquare$ \\ 

\rowcolor{LightCyan}
BBG-Net\citeyearpar{shen2020balanced}&\xmark&Reconstructing information flow with gated residual&$\bigstar,~\blacktriangle$\\
 Real-to-Bin\citeyearpar{martinez2020training} &\cmark~(UO)&Data-driven channel re-scaling gated residual&$\bigstar,~\blacktriangle$\\
 \rowcolor{LightCyan}
MoBiNet-Mid\citeyearpar{phan2020mobinet}&\xmark&\vtop{\hbox{\strut MoBiNet: a lightweight module binarization with}{\hbox{\strut~ the support of skip connection, the three block }{\hbox{\strut~designs and the K-dependency}}}}&\circles,~$\blacklozenge$ \\

MeliusNet\citeyearpar{bethge2020meliusnet} &\cmark~(O)&\vtop{\hbox{\strut DenseBlock: increases the feature capacity, }{\hbox{\strut~ Improvement Block: increases the feature quality}}}&\circles,~$\blacklozenge$ \\
\rowcolor{LightCyan}
 \vtop{\hbox{\strut Binarized MobileNet}{\hbox{\strut\citeyearpar{phan2020binarizing}}}} & \xmark &\vtop{\hbox{\strut Evolutionary search to explore group structures }{\hbox{\strut~when either using depth-wise or fully convolutional }{\hbox{\strut~layers in MobileNet}}}}&\circles,~$\blacklozenge$ \\
 
 High-Capacity-Expert\citeyearpar{bulat2020high}& \cmark~(O) &\vtop{\hbox{\strut Condition computing(experts convolution) }{\hbox{\strut~grouped convolution and NAS strategy}}}&\circles,~$\blacklozenge$  \\ 
 
 \rowcolor{LightCyan}
ReActNet\citeyearpar{liu2020reactnet} & \cmark~(O) &\vtop{\hbox{\strut ReAct-Sign and ReAct-PReLU to reshape and }{\hbox{\strut~shift the activation distributions}}}&\circles,~$\blacklozenge$  \\ 

FracBNN\citeyearpar{zhang2021fracbnn} & \cmark~(O)&\vtop{\hbox{\strut Design a dual-precision activation scheme to }{\hbox{\strut ~compute features }}}&\circles,~$\blacklozenge $\\ 

\rowcolor{LightCyan}
BCNN\citeyearpar{redfern2021bcnn}&\xmark &\vtop{\hbox{\strut Design a network for ImageNet classficiation}{\hbox{\strut ~with lower model size}}}&\circles,~$\blacklozenge$ \\ 
BNN-BN-free\citeyearpar{chen2021bnn}& \cmark~(O)&Replace batch normlization with scaling factors&\circles, $\bigstar, \blacktriangledown$\\

 \rowcolor{LightCyan}
DyBNN\citeyearpar{zhang2022dynamic} & \cmark~(O) &\vtop{\hbox{\strut DySign and DyPReLU to reshape and }{\hbox{\strut~shift the  channel-wise activation distributions}}}&\circles,~$\blacklozenge$  \\

 DyBNN\citeyearpar{zhang2022dynamic} &      \cmark~(O) &     \vtop{\hbox{\strut DySign and  DyPReLU  to reshape and }{\hbox{\strut~shift the activation distributions}}}&     \circles,~$\blacklozenge$  \\ 
 \rowcolor{LightCyan}     
 BiMLP\citeyearpar{xu2022bimlp}&   \cmark~(O)&   \vtop{\hbox{\strut Binary vision Multi-Layer Perceptrons}}&   \circles, $\blacktriangledown$\\

 \hline

\end{tabular}

 \caption{\label{tab:table-name}Summary Table for Network Topology Structure }

  \begin{tablenotes}
          %\footnotesize   %% If you want them smaller like foot notes
   
            \item Note: O: Official implementation, UO: Un-official implementation, $\bigstar$: Modification on classical CNN model,  \circles: Customized Network Structure, $\blacktriangle$:  Increasing shortcut/residual gate, $\blacktriangleleft$: Increasing filters, $\blacktriangleright$: Increasing number of channels, $\blacktriangledown$: other, $\blacksquare$: Search algorithms to optimize BNN
, $\blacklozenge$: aims to beat real value light-weight network 
        \end{tablenotes}
    \end{threeparttable}
    \vspace{-0.6cm}
\end{table}

\subsection{Training Strategy and Tricks}

Different training scheme and tricks can also affect final accuracy of BNN. SQ-BWN \citep{dong2017learning} applied Stochastic Quantization (SQ) algorithm to gradually train and quantize BNN to compensate the quantization error. Bi-Real-Net \citep{liu2018bi} initiated trainable parameters based on pre-train real value network and replaced RELU activation function with Htanh function. How to Train \citep{tang2017train} replaced the RELU activation function with PReLU function and explored that learning rate can have an influence on the accuracy of the final trained BNN. Empirical Study \citep{alizadeh2018empirical} explored the impact of pooling, optimizer and learning rate initialization for training BNN. Bop \citep{helwegen2019latent} and UniQ \citep{pham2021training} separately proposed new optimizer for their BNN training. Main/Subsidiary \citep{xu2019main} proposed filter pruning for BNN. Also, inspired by The Lottery Ticket Hypothesis \citep{frankle2018lottery}, MPT \citep{diffenderfer2021multi} designed a scheme to learn  highly accurate BNN simply by pruning and quantizing randomly weighted full precision CNN. CI-BCNN \citep{wang2019learning} simultaneously trained reinforcement graph model and BNN to alleviate binarization inconsistency, their training scheme is similar to RBNN \citep{ lin2020rotated} which applied generative adversarial network (GAN) to train BNN. Real-to-Bin \citep{martinez2020training} designed a two-step training strategy that applied transfer teaching method to train BNN through learning real value pre-train network. Utilizing Real-to-Bin's training strategy, some BNN works finally trained a high accuracy model such as ReActNet \citep{liu2020reactnet}, High-Capacity-Expert \citep{bulat2020high} and BCNN \citep{redfern2021bcnn}.
Extend based on Real-to-Bin's training strategy, BNN-Adam \citep{liu2021how} investigates and designs a new training scheme based on Adam optimizer and can successfully improve Real-to-Bin and ReActNet's trained performance. BinaryDuo \citep{kim2020binaryduo} proposed a two-stage training scheme to decouple a ternary-activation network into a two-binary-activation BNN network. MD-tanh-s \citep{ajanthan2021mirror} applied mirror descent to optimize BNN' optimizer. Instread of training BNN on conventional hardware such as GPU and TPU, BNN-EP \citep{laydevant2021training} and BNN-Edge \citep{wang2021enabling} explores to directly train BNN on the chip and Edge. BNN-EP \citep{laydevant2021training} proposes to use Equilibrium Propagation (EP) to train BNN and finds its possibility of training on-chip BNNs with compact circuitry. 
BNN-Edge \citeyearpar{wang2021enabling} designs a low-memory and low-energy training scheme by modifying the forward propagation and back propagation including binarying weight gradients, changing batch normalization layer and using low-precision floating-point data.
BNN-stochastic \citeyearpar{Livochka_2021_CVPR} proposes a transfer training and initialization scheme for BNN using the stochastic relaxation approach and improves the accuracy on the small-scale
CIFAR-10 dataset.  Sub-bit Neural Networks (SNNs) \citeyearpar{wang2021sub} proposed a new method to further compress and accelerate BNN in FPGA based on the observation of the binary kernels in BNN. Tables 8-9 are a summary table for BNN that proposed techniques for training strategy and tricks.  \\

\begin{table}[h!]
   \begin{threeparttable}
\centering

\begin{tabular}{lllc}

\hlineB{2.5} 
  \rowcolor{Gray}
 BNN Name &Code Available&Key Idea&Idea Category  \\
 
  \hlineB{2.5}
  \rowcolor{LightCyan}
SQ-BWN\citeyearpar{dong2017learning}&\cmark~(O)&Stochastic quantization(SQ) algorithm for training&\circles \\

Bi-Real-Net\citeyearpar{liu2018bi} &\cmark~(O, UO)&\vtop{\hbox{\strut Initialization: replace ReLU with clip(−1, x, 1) to }{\hbox{\strut ~pre-train the real-valued CNN model}}}  &\circles,~$\blacktriangle$\\

\rowcolor{LightCyan} 
How to Train\citeyearpar{tang2017train}& \xmark&\vtop{\hbox{\strut Low learning rate better, use PReLU, scale }{\hbox{\strut ~ layer, multiple activation}}}&\circles\\

Empirical Study\citeyearpar{alizadeh2018empirical} & \cmark~(O)&\vtop{\hbox{\strut Identify the essential techniques required for   }{\hbox{\strut ~optimisation of BNN}}}&$\blacktriangle$ \\

\rowcolor{LightCyan}
Bop\citeyearpar{helwegen2019latent} & \cmark~(O)&\vtop{\hbox{\strut  Bop: a Latent-Free Optimizer designed specifically }{\hbox{\strut ~ for BNN}}}&$\blacktriangle $\\

Main/Subsidiary\citeyearpar{xu2019main} &\cmark~(O)&BNN filter-level pruning&$\blacktriangle$ \\

\rowcolor{LightCyan} 
CI-BCNN\citeyearpar{wang2019learning}&\xmark&\vtop{\hbox{\strut Train BNN and reinforcement graph model }{\hbox{\strut ~ simultaneously to alleviate binarization}{\hbox{\strut ~ inconsistency }}}}&\circles\\

RBNN\citeyearpar{ lin2020rotated} & \xmark &Use generative adversarial network to train&\circles\\

\rowcolor{LightCyan} 
Real-to-Bin\citeyearpar{martinez2020training} &\cmark~(UO)&\vtop{\hbox{\strut Two-step training strategy: spatial attention transfer  }{\hbox{\strut ~computed from a teacher real-valued network to }{\hbox{\strut ~the binary
network.}}}} &\circles\\

BinaryDuo\citeyearpar{kim2020binaryduo}&\cmark~(O)&Decouple  ternary activation to two binary activations&\circles \\

\rowcolor{LightCyan} 
ReActNet\citeyearpar{liu2020reactnet} & \cmark~(O) &Adopt two-step training strategy from Real-to-Bin  &\circles  \\ 

High-Capacity-Expert\citeyearpar{bulat2020high}&\cmark~(O) &\vtop{\hbox{\strut Adopt and improve two-step training strategy from}{\hbox{\strut ~ Real-to-Bin}}}  &\circles \\ 

\rowcolor{LightCyan} 
MD-tanh-s\citeyearpar{ajanthan2021mirror}&\cmark~(O)&Apply mirror descent to BNN&$\blacktriangle$\\

UniQ\citeyearpar{pham2021training}&\cmark~(O)&\vtop{\hbox{\strut  Special optimizer and warm-up strategy for binary }{\hbox{\strut ~ training with symmetric quantizer}}}&\circles,$\blacktriangle$ \\

\rowcolor{LightCyan} 
BCNN\citeyearpar{redfern2021bcnn}&\xmark&\vtop{\hbox{\strut Adopt two-step training strategy from Real-to-Bin}} &\circles
\\
MPT\citeyearpar{diffenderfer2021multi}&\cmark~(O)&Multi-Prize Lottery Ticket Hypothesis &$\blacktriangle$\\

 \hline

\end{tabular}
 \caption{\label{tab:table-name}Summary Table for Training Strategy and Tricks}

  \begin{tablenotes}
          %\footnotesize   %% If you want them smaller like foot notes
   
            \item Note:  O: Official implementation, UO: Un-official implementation, \circles: Train strategy, $\blacktriangle$: tricks/activations/optimizer/learning rate/pruning  
        \end{tablenotes}
    \end{threeparttable}
  \vspace{-0.8cm}
\end{table}

\begin{table}[h!]
   \begin{threeparttable}
\centering

\begin{tabular}{lllc}

\hlineB{2.5} 
  \rowcolor{Gray}
 BNN Name &Code Available&Key Idea&Idea Category  \\
 
  \hlineB{2.5}

 \rowcolor{LightCyan} 
BNN-stochastic\citeyearpar{livochka2021initialization}&\xmark&Initialization and Transfer Learning stochastic BNN&\circles\\
BNN-Edge\citeyearpar{wang2021enabling}&\cmark~(O)&Low-memory and low-energy training &\circles\\

\rowcolor{LightCyan} 
 BNN-EP\citeyearpar{laydevant2021training}&\cmark~(O)&Equilibrium Propagation for training BNN &\circles\\
  BNN-Adam\citeyearpar{liu2021how}&\cmark~(O)&Adam-based optimizers investigation &\circles,$ \blacktriangle$\\
  \rowcolor{LightCyan} 
  SNNs\citeyearpar{wang2021sub}&\cmark~(O)&Further compress BNN &\circles,$ \blacktriangle$\\
 \hline

\end{tabular}
 \caption{\label{tab:table-name}Summary Table for Training Strategy and Tricks (Continue to Table 8)}

  \begin{tablenotes}
          %\footnotesize   %% If you want them smaller like foot notes
   
            \item Note:  O: Official implementation, UO: Un-official implementation, \circles: Train strategy, $\blacktriangle$: tricks/activations/optimizer/learning rate/pruning  
        \end{tablenotes}
    \end{threeparttable}
  \vspace{-0.8cm}
\end{table}
 \subsection{Summary}

In this section, we put BNN enhancement methods into five categories: (1) quantization error minimization, (2) loss function improvement, (3) gradient approximation, (4) network topology structure, and (5) training strategy and tricks. With the development of BNN optimization, we notice that just unitizing one enhancement method is hard to improve BNN’s accuracy performance. We understand some problems are still unsolved.

How does each binarization layer affect the entire BNN performance? Understanding the degree of information loss from each binarization layer can promote the production of layer-wise optimization methods. Besides, OPs are becoming an equally important performance indicator as well as accuracy rate. To design BNN with high accuracy and lower operations per second (OPs) simultaneously becomes critical.
How to effectively speed up BNN training time? Although BNN has faster inference speed and lighter weight size in resource-limited devices, training BNN still has to be done on conventional devices such as GPU and takes expansive computing cost and a long time. There are a few published works that can reduce memory and energy usage. But we still cannot find any breakthrough to significantly reduce BNN training time. How to effectively speed up BNN training time is still an open problem.

For a given BNN application, which training strategy and tricks should be used? There are many different published works that report their proposed training strategies and tricks that can improve trained models’ accuracy performance. However, all the benchmark results were tested based on the designed BNN structure and specific datasets such as CIFAR-10 or ImageNet. We are not sure if a similar improvement effect could be achieved in different datasets and BNN variants. It is necessary to do a survey research study to compare the difference among proposed training strategies and tricks.

%%%%%%%%%%%%%%%%%%%%%%%%%%%%%%%%%%%%%%%%%%%%%%%%%%%%%%%%%%%%%%%%%%%%%%%%%%%
\section{Open Source Frameworks of Binary Neural Network}
 \subsection{Open Source Frameworks of BNN Introductions} 
BNN has the ability to decrease the memory consumption and computational complexity. However, most published implementations of BNN do not really store their weight parameters in the binary format and cannot use XNOR and popcount to perform binary matrix multiplications in convolutions and fully connected layers. The reason is that deep learning models are directly implemented by python frameworks such as TensorFlow \citep{abadi2016tensorflow} and PyTorch \citep{paszke2019PyTorch}, but python cannot store the data in binary form and does bit type data operations like C/C++ language. In the literature, there are several published available  open-source BNN inference framework that can make the BNN' models achieve the actual BNN performance. This section introduces and reviews the published BNN inference frameworks BMXNet \citep{yang2017bmxnet}, BMXNet2 \citep{bethge2020bmxnet}, daBNN \citep{zhang2019dabnn}, Riptide \citep{fromm2020riptide}, FINN \citep{blott2018finn}  and Larq \citep{bannink2021larq}.\\
\indent BMXNet is an Apache-licensed open-source BNN library framework. It is written based on MXNet \citep{chen2015mxnet} which is a high-performance and modular deep learning library. Depends on custom MXNet operators such as QActivation, QConvolution and QFullyConnected, BMXNet is able to support quantization and binarization of input data and weights. BMXNet can store the weights of convolutional and fully connected layers in their binarized format and perform matrix multiplication using bit-wise operations (XNOR and popcount). The BMXNet library, several sample code and a collection of pre-trained binary deep models are available  at https://github.com/hpi-xnor\\
\indent daBNN is a BSD-licensed open-source BNN inference framework highly optimized for ARM-based devices. daBNN designs an upgraded bit-packing scheme to pack multiple elements simultaneously, which reports that the speed of naive sequential method by about 4 times. In additions, daBNN proposes a new binary direct convolution to squeeze the cost of extra instructions in binary convolution, and creates a new novel memory layout to reduce memory access. daBNN is implemented in C++ and ARM assembly. Also, daBNN provides Java and Android package. Compared to BMXNet, daBNN is constructed based on standard ONNX \cite{onnx} operators (Sign and Convolution) to  ensure interoperability. daBNN can convert PyTorch float-points models to BNN models. daBNN reports that their converted BNN model's performance is 7-23 times faster on a single binary convolution than BMXNet. daBNN's source code, sample projects and pre-trained models are available on-line: https://github.com/JDAI-CV/dabnn\\
\indent BMXNet2 is an Apache-licensed open-source BNN framework, implemented based on BMXNet framework. Compared to the original BMXNet framework, BMXNet2 reuses more of the original MXNet operators and only adds three new functions in the C++ backend: the sign, round with STE and gradcancel operator. BMXNet2 can easily have minimal changes to C++ code to get better maintainability with future versions of MXNet. Besides, in the original BMXNet, the code for optimized inference was mixed with the training code. In BMXNet2, the two parts' code is separately implemented, which can further simplify debugging and unit testing. The BMXNet2 source code and demos are available at https://github.com/hpi-xnor/BMXNet-v2

\indent
Riptide is a private licensed open-source BNN framework. It is built on the top of TensorFlow and TVM \citep{chen2018tvm}, to service for BNN training and deployment. TVM is an open source deep learning compiler framework for diverse hardware environments including CPUs, GPUs, and deep learning accelerators. TVM can automatically deliver the optimized kernels for deep learning models on a special hardware platform. Depended on TVM, Riptide designs and develops new customs functions to support the optimized kernels  generation for BNN. And Riptide proposes a new solution to completely remove floating-point arithmetic in the intermediate ‘glue’ layers(weight scaling, batch normalisation, and binary re-quantization) between pairs of binarized convolutions. Riptide reports their BNN models performance can achieve 4-12 times speed-up compared to a floating-point implementation. The Riptide source code and library are available at https://github.com/jwfromm/Riptide\\
\indent
FINN is a BSD-3-Clause License framework developed and maintained by Xilinx Research Labs. It services Xilinx's series of FPGA boards. The framework can support model's development, training, format conversion and embed in boards for both low-bit networks and 1-bit BNN. FINN has three components where are (1) brevitas: a PyTorch library for model develop and training; (2) FINN compiler: model format transformation and compiled; and (3) PYNQ: a python package to connect transformed model with Xilinx board. The FINN source code, library and demos are available at https://github.com/Xilinx/FINN\\
\indent
Larq is an Apache-licensed open-source BNN framework. It is built on the top of TensorFlow and TensorFlow Lite, and servicing for BNN model deployment, training and conversion. Larq contains two parts: Larq library and Larq Compute Engine (LCE). Larq library includes BNN quantization functions which extends TensorFlow. LCE contains a TensorFlow model graph converter and highly optimized implementations of binary operations and accelerations. Larq reports that it is the art-of-the-state fastest BNN framework over the existing inference frameworks. The Larq source code, library and demos are available at https://github.com/larq/larq

\vspace{0.5cm}
\begin{table}[h!]

\begin{tabular}{ llll }
 %\hline
 %\multicolumn{4}{|c|}{Country List} \\
\hlineB{2.5} 
  \rowcolor{Gray}
 Framework & Model Format Base &Model Training &Maintenance\\
 \hlineB{2.5}
  \rowcolor{LightCyan}
 BMXNet    & MXNet  & MXNet &  Until Nov 18, 2019\\
 
BMXNet2 &MXNet and BMXNet  &MXNet &Until Jul 2, 2020\\
 \rowcolor{LightCyan}
daBNN &ONNX & PyTorch &  Until Nov 11, 2019
\\
 Riptide   &TensorFlow and TVM&TensorFlow&  Until May 13, 2020\\
  \rowcolor{LightCyan}
 FINN &ONNX&Brevitas(PyTorch Modification)&Present - \\

 Larq&   TensorFlow and TensorFlow Lite  &TensorFlow&Present - \\

 \hline

\end{tabular}
 \caption{\label{tab:table-name}BNN Frameworks Comparisons and Characterises}

\end{table}
{\subsection{   BNN Library for CPUs and GPUs} 
Exclude the discussed open source frameworks for BNN, there are a few of published research works that focus on BNN Library on CPUs and GPUs. Espresso \citep{pedersoli2017espresso} is a library written in C/CUDA that does  bit type data operations and bitwise operations required for the forward propagation of BNN on CPUs and GPUs.  BitFlow \citep{hu2018bitflow} is the first implemented framework for exploiting computing
power of BNNs on CPU. BSTC \citep{li2019bstc} and O3BNN \citep{geng2020o3bnn} are architecture design strategies for BNN on FPGAs, CPUs, and GPUs. Phonebit \citep{chen2020phonebit} is a GPU-accelerated
BNN inference engine for Android-based mobile devices.  
 \subsection{Summary} 
When native BNN was published and taking full advantage of bits, users had to re-implement the model in a low-level programming language such as C/C++ to embed the BNN model on tiny devices such as FPGA or mobile phones. Such an engineering task is not convenient for un-professional software developers. With the rapid development of BNN open-source platforms, we can easily transform the trained BNN model to the related inference format on tiny devices without worrying about professional engineering works.

However, as table 10 shows, only two BNN frameworks, FINN and Larq, are actively maintained. BMXNet, daBNN, BMXNet2, and Riptide have stopped version updating. Unfortunately, Larq only supports TensorFlow based models, and FINN exclusively services Xilinx’s FPGA boards. There are no other options for developers who prefer to use different libraries such as PyTorch to design and develop BNN systems. 
{   Besides, to fully unlock BNN model benefits on GPUs, researchers and developers have to re-implement the BNN models with C/C++ and CUDA. }How to create and maintain the cross-platform open-source framework that can efficiently import any library-based BNN models, like NCNN (Tencent, 2017), is a new exciting   research problem and opportunity.

\section{   Limit-resource Hardware Architecture} 

\begin{figure}[h!]    %%%%%%%%%%%%%%%%%% FIGURE 4
   \centerline{\includegraphics[width=1\textwidth,clip=]{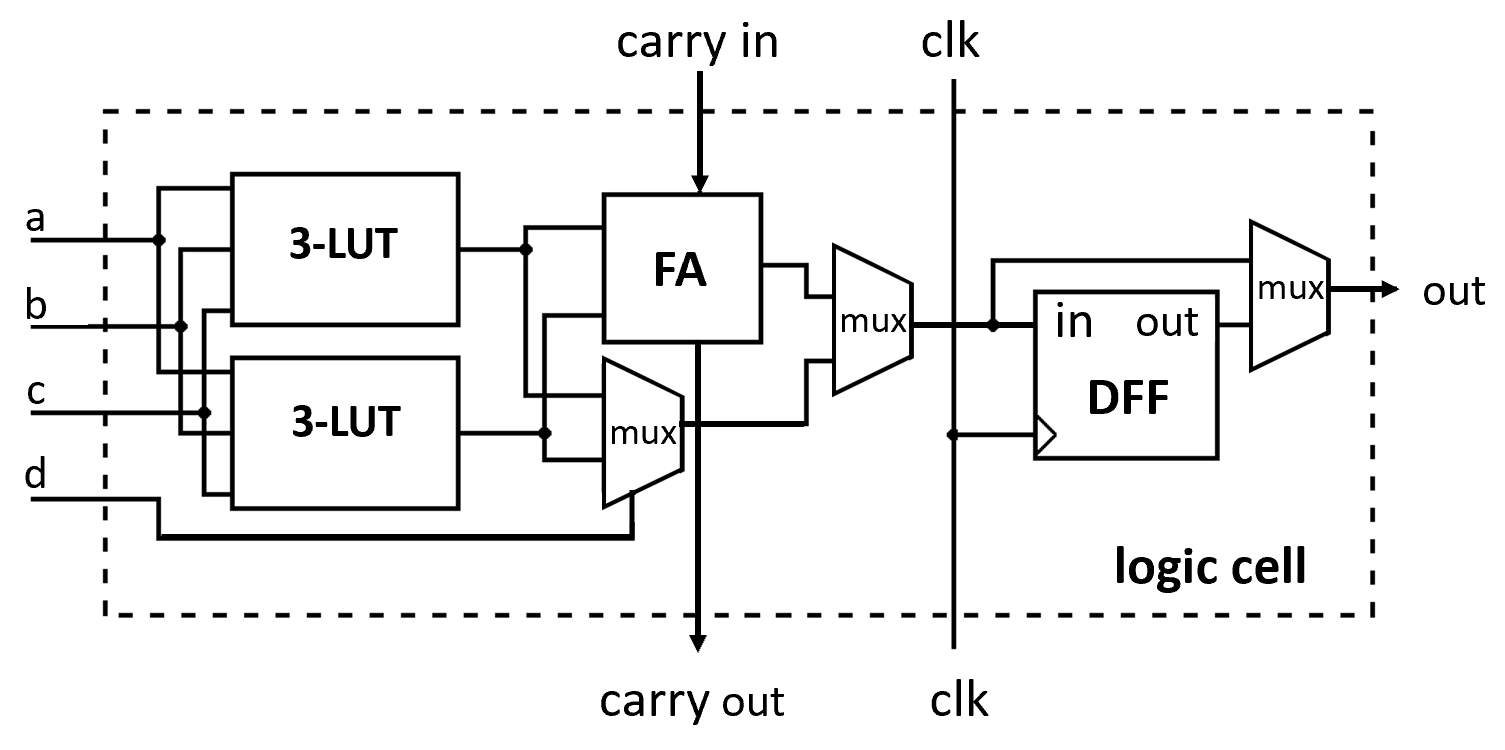}
              }
   \caption{     An example of FPGA logic cell (Drawed by Petter Kallstrom,  public domain license)  }
    \label{F-appendix}
  \end{figure}
  \noindent Currently, in many real-world applications such as robotics,
wearable devices and even self-driving car, recognition vision tasks
need to be carried out in a timely fashion on a computationally limited platform. Instead of applying complex and heavy-weight neural network in expensive hardware such as graphics processing unit (GPU), it is a trend to use the resource constraint hardware to embed with efficiency trained models. In this section, limit-resource hardware is briefly introduced and discussed. Among the various kinds of efficiency devices, field programmable gate array (FPGA) is one of the most popular circuit architectures.  Because the dominant computations on FPGA are bitwise logic operations and FPGA's memory requirements are greatly reduced, FPGA is well suited for BNN. Figure 14 is an example of an FPGA logic cell.To make this figure and the rest of this section easy to understand, we first define the terms and units used in this section.

\begin{quote}
{\textbf{FPGA}: an abbreviation for {field programmable gate array}. FPGA is an integrated circuit that enable users to program for different algorithms after manufacturing.}
\end{quote}
\begin{quote}
\indent \textbf{LUT}: an abbreviation for {Look-Up Table}. FPGA uses it to do boolean algebra such as XNOR, AND, OR, etc. The LUT can be programmed by the designer to execute a boolean algebra equation. 
\end{quote}
\begin{quote}
\indent \textbf{FA}: an abbreviation for {Full Adder}. FA is central to most digital circuits that perform addition or subtraction.  
\end{quote}
\begin{quote}
\indent \textbf{DFF}: an abbreviation for {D flip-flop}. DFF is also known as a "data" or "delay" flip-flop, which is used for the storage of state. One DFF keeps a single bit (binary digit) of data; one of its two states represents a "one" and the other represents a "zero". 
\end{quote}
\begin{quote}
\indent \textbf{MUX}: an abbreviation for {multiplexer}, that selects a single input among input set to output. 
\end{quote}
\begin{quote}
\indent \textbf{BRAM}: an abbreviation for {Block Random Access Memory}. BRAM is also known as "Block RAMs" Block RAMs are used for storing large amounts of data in FPGA.
\end{quote}
\begin{quote}
\indent \textbf{clk}: an abbreviation for {clock}. clk is a signal inside any digital circuit which presents how the performance a flip flop (or a group of flip flops) works. On the FPGA platform, the faster the clock, the faster the designed function will run.
\end{quote}
\begin{quote}
\indent \textbf{DSP slices}: (digital signal processing)DSP slices form the basis of a versatile, coarse grain DSP architecture in Xilinx FPGA, that can enable efficiently add powerful FPGA-based DSP functionality. 
\end{quote}
\begin{quote}
\indent \textbf{FPS}: an abbreviation for {Frames Per Second}. FPS measures the frame rate that evaluates one trained model's inference speed. 
\end{quote}
\begin{quote}
\indent \textbf{ASIC}: an abbreviation for {Application Specific Integrated Circuit}. Different from FPGA, it does not allow users to reprogram or modify after fabrication. 
\end{quote}
\begin{quote}
\indent \textbf{SBC}: an abbreviation for {Single-Board Computer}. SBC is a complete computer built on a single circuit board such as Raspberry Pi. It contains microprocessor(s), memory, input/output(I/O) and other features required of a functional computer. 
\end{quote}
\begin{quote}
\indent \textbf{ARM-processor}: stand for one kinds of CPU based on reduced instruction set computing(RISC) architectures. It is widely used in mobile machine and efficiency platform. 
\end{quote}
\begin{quote}
\indent \textbf{SoC}: an abbreviation for {System-on-a-Chip}. SoC comprises of various functional units on a single silicon chip. 
\end{quote}
\begin{quote}
\indent \textbf{SoC FPGA}: SoC FPGA devices integrate both processor and FPGA architectures into a single device such as Xilinx family's FPGA development board. 
\end{quote}

\begin{figure}[h]   %%%%%%%%%%%%%%%%%% FIGURE 4
   \centerline{\includegraphics[width=0.9\textwidth,clip=]{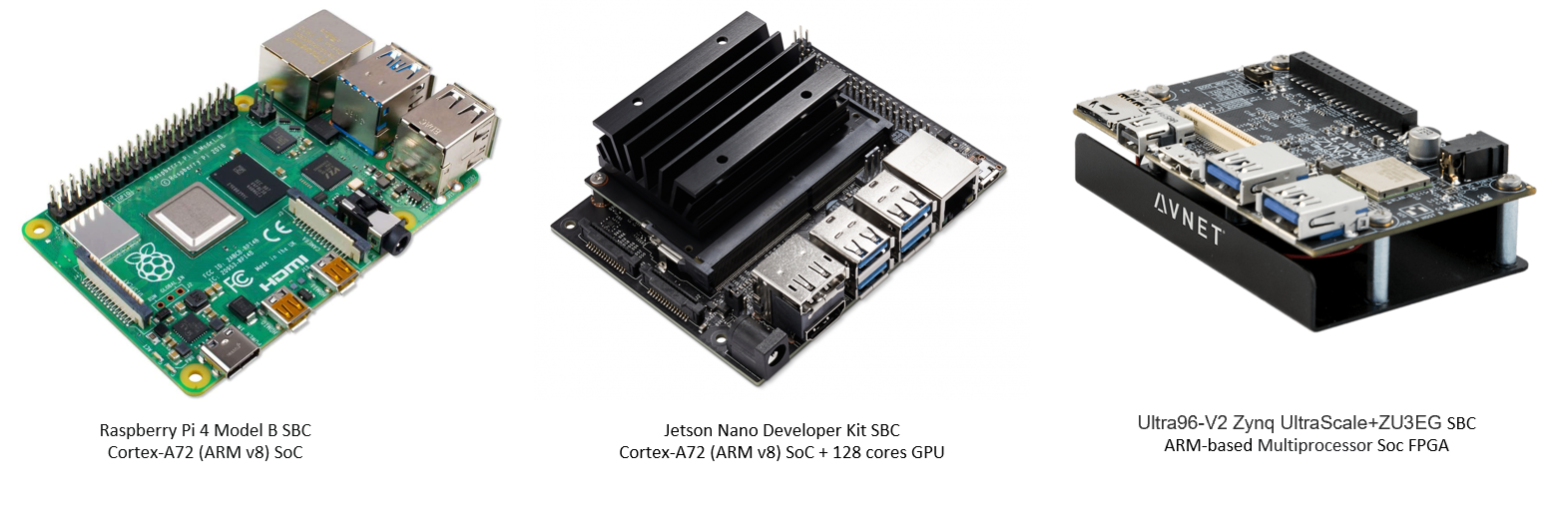}
              }
   \caption{Representative SBC platforms }
    \label{F-appendix}
  \end{figure}
\noindent Figure 15 shows some representative resource-constrained SBC platforms based on FPGA and ARM processors. To testing if one neural network has a good efficiency performance on the resource-limited platform, there are common efficiency indicators including inference memory usage and inference speed such as FPS or latency time. In a FPGA platform, the number of measured DSP, LUT, BRAM and powers can also explain one trained model's efficiency performance. In recent years, there are some works that report efficiency performance testing for BNN in above platforms. The performance results based on the special dataset are listed in the tables for reference in the next section.

\clearpage
%%%%%%%%%%%%%%%%%%%%%%%%%%%%%%%%%%%%%%%%%%%%%%%%%%%%%%%%%%%%%%%%%%%%%%%%%%
\section{Binary Neural Network Applications} 

\subsection{Image Classification }
\vspace{-0.1cm}
Image classification is the basic benchmark testing task to evaluate BNN' performance. The following tables show the summary of a variety of BNN' performance on common benchmark image datasets including CIFAR-10 \citep{CIFAR-10data} and ILSVRC12 ImageNet \citep{russakovsky2015imagenet}. For each BNN model result, we contain the published years, BNN name, topology and accuracy. In particular for ImageNet results, we also contain the BOPs, FLOPs and OPs results (OPs = FLOPs + 1/64*BOPs \citep{Bethge_2019_ICCV,liu2018bi}) which are equally important to represent the BNN efficiency performance compared to accuracy results. For BNN, FLOPs is for the floating-point operations in the networks, excepting the operation calculated in BOPs which is bitwise. For full precision networks, OPs equals to FLOPs. To the best of our ability, we collect all the published reliable BNN' performance results at the time of this paper submission. 
\vspace{-0.1cm}
\subsubsection{CIFAR-10 Dataset }
\vspace{-0.1cm}
CIFAR-10 dataset consists of 60000 color images in 10 classes. Each class has 6000 images and each image is 32x32 pixels. There are 50000 training images and 10000 test images.

%\vtop{\hbox{\strut Customized DenseNet37 }{\hbox{\strut~(dilated)}}}
\begin{table}[!h]

\begin{threeparttable}

\begin{tabular}{llllllllll}

\hlineB{2.5} 
  \rowcolor{Gray}
BNN&Device & FPS & Acc(\%)  & Bits(W/A) & $F_{max}$(MHz)&Power(W)     & DSP& BRAM & LUT                 \\
\hlineB{2.5} 
\rowcolor{LightCyan} 
Acce-BNN\citeyearpar{zhao-bnn-fpga2017} &\vtop{\hbox{\strut Zynq}{\hbox{\strut 7Z020}}}&168.4&\vtop{\hbox{\strut 87.73 }{\hbox{\strut 88.8\footnotemark[1]}}}&1/1&143&4.7&3 &94&46900\\

 FC-BNN\citeyearpar{Nakahara2017AFC} &\vtop{\hbox{\strut Zynq}{\hbox{\strut 7ZC020}}} & 420& 81.8 & 1/1 &143&2.3    &1& 32& 14509                  \\

\rowcolor{LightCyan} 
 FO-BNN\citeyearpar{10.1145/3218603.3218615} &\vtop{\hbox{\strut Zynq}{\hbox{\strut 7ZC020}}} & 930& 86 & 1/1 &143&2.4   &53& 135& 23436                  \\
 FP-BNN \citeyearpar{liang2018fp} &Stratix-V&7692.3&86.31&1/1&150&26.2&20 &2210&219010\\
 
 \rowcolor{LightCyan} 
 ReBNet\citeyearpar{ghasemzadeh2018rebnet} &\vtop{\hbox{\strut Zynq}{\hbox{\strut ZC702}}} & 2000 & 86.98  & 1/1 &200&-    &-& -&-                 \\

 FBNA\citeyearpar{8532584} &\vtop{\hbox{\strut Zynq}{\hbox{\strut ZC702}}} & 520.8 & 88.61 & 1/1 &-&3.3    &-& 103& 29600                  \\

\rowcolor{LightCyan} 
FracBNN\citeyearpar{zhang2021fracbnn} &\vtop{\hbox{\strut Zynq}{\hbox{\strut ZU3EG}}} & 2806.9 & 89.1 & 1/1.4\footnotemark[2] &250&4.1    &126& 212& 51444                 \\

    \hline

\end{tabular}
 \caption{\label{tab:table-name} BNN efficiency comparisons on CIFAR-10 using FPGA }
 
  \begin{tablenotes}
          %\footnotesize   %% If you want them smaller like foot notes
         \item Note:  \textbf{${}^1$}: support materials from its code GitHub page, \textbf{${}^2$}:   1.4bit  based  on  the  analysis  of  quantized  activations, convolution still executes 1W/1A operations
        \end{tablenotes}
 
 \end{threeparttable}
 \vspace{-0.2cm}
\end{table}

\begin{table}[!h]
 {
\begin{threeparttable}

 \caption{\label{tab:table-name}BNN Accuracy comparisons on CIFAR-10}
\begin{tabular}{lll}
\hlineB{2.5} 
  \rowcolor{Gray}
 \multicolumn{3}{c}{Full Precision Base CNN} \\
 \rowcolor{Gray}
\multicolumn{2}{l}{CNN Name}
                    & \multicolumn{1}{l}{Acc(\%)}\\
 \hlineB{2.5}
  \rowcolor{LightCyan}
\multicolumn{2}{l}{VGG-Small\citeyearpar{zhang2018lq}}
                    & \multicolumn{1}{l}{93.8}\\

\multicolumn{2}{l}{VGG-11\citeyearpar{xu2019main}}
                    & \multicolumn{1}{l}{83.8}\\
  \rowcolor{LightCyan}
\multicolumn{2}{l}{NIN\citeyearpar{xu2019main}}
                    & \multicolumn{1}{l}{84.2}\\

\multicolumn{2}{l}{ResNet-18\citeyearpar{Qin:cvpr20}}
                    & \multicolumn{1}{l}{ 93.0}\\
  \rowcolor{LightCyan}
\multicolumn{2}{l}{ResNet-20\citeyearpar{Qin:cvpr20}}
                    & \multicolumn{1}{l}{ 91.7}\\

\multicolumn{2}{l}{WRN-22\citeyearpar{zagoruyko2016wide}}
                    & \multicolumn{1}{l}{ 92.62}\\
  \rowcolor{LightCyan}
\multicolumn{2}{l}{WRN-22(4 x Kernel Stage)\footnotemark[1]\citeyearpar{zagoruyko2016wide}}
                    & \multicolumn{1}{l}{ 95.75}\\

 \hlineB{2.5}
 \rowcolor{Gray}
\multicolumn{3}{c}{BNN Accuracy Performance}
                 \\

 \rowcolor{Gray}
 BNN Name & Topology & Acc(\%)                \\
 \hlineB{2.5}
   \rowcolor{LightCyan}
 BNN\citeyearpar{courbariaux2016binarized} & VGG-Small & 87.13               \\

 {XNOR-Net\citeyearpar{rastegari2016xnor} }& VGG-Small & 87.38             \\
& WRN-22 & 81.90\citeyearpar{gu2019bayesian}           \\
& WRN-22(4 x Kernel Stage)\footnotemark[1] \enspace \enspace \enspace \enspace \enspace \enspace& 88.52\citeyearpar{gu2019bayesian}     \enspace \enspace       \\
&ResNet-18&90.21\citeyearpar{chen2021bnn}\\

  \rowcolor{LightCyan}

    \hline

\end{tabular}
 \begin{tablenotes}
          %\footnotesize   %% If you want them smaller like foot notes
         \item Note:  \textbf{${}^1$}: WRN-22 \citep{zagoruyko2016wide} original kernelstage is 16-16-32-64,  %\textbf{${}^5$}: OPs similar .
        \end{tablenotes}
    \end{threeparttable}
}
\end{table}

\begin{table}[!h]
 {
\begin{threeparttable}

 \caption{\label{tab:table-name}Continue Table 12 BNN Accuracy comparisons on CIFAR-10}
\begin{tabular}{lll}
\hlineB{2.5}

 \rowcolor{Gray}
\multicolumn{3}{c}{BNN Accuracy Performance}
                 \\

 \rowcolor{Gray}
 BNN Name & Topology & Acc(\%)                \\
 \hlineB{2.5}

  \rowcolor{LightCyan}
 LAB2\citeyearpar{hou2016loss} & VGG-Small & 87.72            \\

 DoReFa-Net\citeyearpar{zhou2016dorefa} & ResNet-20 & 79.3\citeyearpar{rastegari2016xnor}           \\
  \rowcolor{LightCyan}
 HORQ\citeyearpar{li2017performance} & Customized & 82.0\citeyearpar{rastegari2016xnor} \\  
GB-Net\citeyearpar{sakr2018true} & Customized & 89.59 \\  
  \rowcolor{LightCyan}
Bi-Real-Net\citeyearpar{liu2018bi}&ResNet-18$\ast \ast$\footnotemark[2]&89.12\citeyearpar{chen2021bnn}\\
HadaNet\citeyearpar{akhauri2019hadanets} &Customized($\beta$w=4;$\beta$a=4) & 88.64 \\ 
& NIN($\beta$w=4; $\beta$a=4) & 87.33 \\ 
&Customized($\beta$w=16; $\beta$a=2) & 89.02 \\
& NIN($\beta$w=16; $\beta$a=2) & 88.74 \\ 
  \rowcolor{LightCyan}
{PCNN\citeyearpar{gu2019projection}} & WRN-22 & 89.17(J=1)\footnotemark[3]  \\
  \rowcolor{LightCyan}
&  WRN-22 & 91.27(J=2)\footnotemark[3] \\
  \rowcolor{LightCyan}
&  WRN-22 & 92.79(J=4)\footnotemark[3] \\
  \rowcolor{LightCyan}
& WRN-22(4 x Kernel Stage)\footnotemark[1] & 94.31(J=1)\footnotemark[3]\\
  \rowcolor{LightCyan}
& WRN-22(4 x Kernel Stage)\footnotemark[1] & 95.39(J=4)\footnotemark[3]\\

{BONN\citeyearpar{gu2019bayesian}} & WRN-22 & 87.34  \\
&  WRN-22(4 x Kernel Stage)\footnotemark[1] & 92.36\\
  \rowcolor{LightCyan}
 CBCN\citeyearpar{liu2019circulant}&Customized ResNet-18(4W, 4A)\footnotemark[4]& 90.22\\

RBNN\citeyearpar{ lin2020rotated}&WRN-22(4 x Kernel Stage)\footnotemark[1]&93.28\\
  \rowcolor{LightCyan}
{BNN-DL\citeyearpar{ding2019regularizing}} & VGG-Small & 89.90 \\
  \rowcolor{LightCyan}
&ResNet-18&90.47\\

CCNN\citeyearpar{xu2019accurate}  & VGG-Small & 92.3  \\
  \rowcolor{LightCyan}
{CI-BCNN\citeyearpar{wang2019learning}} & VGG-Small & 92.47  \\
  \rowcolor{LightCyan}
& ResNet-20 & 91.10 \\
{Main/Subsidiary\citeyearpar{xu2019main}} & NIN & 83.11 \\
& VGG-11 & 81.97 \\
& ResNet-18 & 86.39 \\
  \rowcolor{LightCyan}
{DSQ\citeyearpar{gong2019differentiable}} & VGG-Small & 91.72  \\
  \rowcolor{LightCyan}
& ResNet-20 & 84.11 \\

{Search Accurate\citeyearpar{shen2019searching}} & Customized VGG & 92.17  \\
& Customized VGG & 93.06 \\
  \rowcolor{LightCyan}
{BBG-Net\citeyearpar{shen2020balanced}} & ResNet-20  & 85.34  \\
  \rowcolor{LightCyan}
& ResNet-20(2W,2A)\footnotemark[1] & 90.71  \\
  \rowcolor{LightCyan}
& ResNet-20(4W,4A)\footnotemark[1]   & 92.46  \\

 SI-BNN\citeyearpar{wang2020sparsity}  & VGG-Small & 90.2  \\
  \rowcolor{LightCyan}
{IR-Net\citeyearpar{ qin2020forward}} & VGG-Small & 90.4  \\
  \rowcolor{LightCyan}
& ResNet-18 & 91.5 \\
  \rowcolor{LightCyan}
& ResNet-20 & 85.4 \\
  \rowcolor{LightCyan}
& ResNet-20$\ast \ast$ \footnotemark[2] & 86.5 \\

{SLB\citeyearpar{NEURIPS2020_2a084e55}} & VGG-Small & 92.0  \\
& ResNet-20 & 85.5 \\

\rowcolor{LightCyan}
{RBNN\citeyearpar{ lin2020rotated}} & VGG-Small & 91.3  \\
\rowcolor{LightCyan}
& ResNet-18 & 92.2 \\
\rowcolor{LightCyan}
& ResNet-20 & 86.5 \\
\rowcolor{LightCyan}
& ResNet-20$\ast \ast$ \footnotemark[2] & 87.8 \\

DMS\citeyearpar{Li_2020_ICLR_NAS}&Customized VGG-11(DMS-A)&84.16\\
&Customized VGG-11(DMS-B)&89.10\\
&Customized ResNet-18(DMS-A)&89.32\\
&Customized ResNet-18(DMS-B)&92.70\\

\rowcolor{LightCyan}
{BNAS\citeyearpar{kim2020learning} }& Customized\footnotemark[5] ResNet-18 &92.70\\
\rowcolor{LightCyan}
& Customized\footnotemark[5] ResNet-34 & 93.76 \\

BATS\citeyearpar{bulat2020bats} &Customized &96.1\\

    \hline

\end{tabular}
 \begin{tablenotes}
          %\footnotesize   %% If you want them smaller like foot notes
         \item Note:  \textbf{${}^1$}: WRN-22 \citep{zagoruyko2016wide} original kernelstage is 16-16-32-64, \textbf{${}^2$}: ResNet{$\ast \ast$}: Variant ResNet\citep{liu2018bi}, \textbf{${}^3$}: J is  total projection number, \textbf{${}^4$}: number of Activation and Weights, %\textbf{${}^5$}: OPs similar .
        \end{tablenotes}
    \end{threeparttable}
}
\end{table}
\mbox{}
\clearpage
 \vspace{0.5cm}
 \begin{table}[!h]

\begin{threeparttable}

\begin{tabular}{lll}

 \hlineB{2.5}
 \rowcolor{Gray}
 BNN Name & Topology & Acc(\%)                \\

 \hlineB{2.5}

\rowcolor{LightCyan}
{ReActNet\citeyearpar{liu2020reactnet} }& \vtop{\hbox{\strut ReActNet-A }{\hbox{\strut(Customized\footnotemark[5] MobileNet-v1) }}} & 82.95\citeyearpar{chen2021bnn}\\
\rowcolor{LightCyan}
&Customized ResNet-20&85.8\citeyearpar{zhang2021fracbnn}\\
\rowcolor{LightCyan}
& \vtop{\hbox{\strut ReActNet-18 }{\hbox{\strut(Customized ResNet-18)}}} & 92.31\citeyearpar{chen2021bnn}\\

FracBNN\citeyearpar{zhang2021fracbnn} &Customized ResNet-20 &87.2\\

\rowcolor{LightCyan}
MPT\citeyearpar{diffenderfer2021multi}& VGG-Small(75\% weights pruned)&88.52\\ \rowcolor{LightCyan}
& VGG-Small+BN\footnotemark[6](75\% weights pruned)&91.9\\

BNN-BN-free\citeyearpar{chen2021bnn}&XNOR(Based on ResNet-18)&79.67\\

&Bi-Real-Net(Based on ResNet-18$\ast \ast$\footnotemark[2])&79.59\\

&\vtop{\hbox{\strut ReActNet-18 }{\hbox{\strut(Customized ResNet-18)}}}&92.08\\

&\vtop{\hbox{\strut ReActNet-A }{\hbox{\strut(Customized MobileNet-v1)}}}&83.91\\

\rowcolor{LightCyan}   LCR\citeyearpar{shang2022lipschitz} &   ResNet-18 &   91.8\\ 
\rowcolor{LightCyan}
&    ResNet-20&   86.0\\ 
\rowcolor{LightCyan}
&    Bi-Real-Net(Based on ResNet-18$\ast \ast$\footnotemark[2])&   87.2\\

ReCU\citeyearpar{xu2021recu}& VGG-Small&92.2\\
& ResNet-18&92.8\\ 
& ResNet-20&87.4\\

\rowcolor{LightCyan}   FDA \citeyearpar{xu2021learning}&    VGG-Small&   92.54\\
&    ResNet-20&   86.2\\

   SiMaN\footnotemark[7]\citeyearpar{lin2022siman}&    VGG-Small&   92.5\\

&    ResNet-18&   92.5\\ 

&    ResNet-20&   87.4\\ 
\rowcolor{LightCyan}
   DIR-Net\footnotemark[8]\citeyearpar{qin2022distribution}&   
VGG-Small&   91.1\\

&    ResNet-18&   92.8\\ 

&    ResNet-20&   89.0\\

   AdaBin\citeyearpar{tu2022adabin}&    VGG-Small&   92.3\\
\rowcolor{LightCyan}
&    ResNet-18&   93.1\\ 
\rowcolor{LightCyan}
&    ResNet-20&   88.2\\ 

    \hline

\end{tabular}
 \caption{\label{tab:table-name}Continue Table 12 BNN Accuracy comparisons on CIFAR-10 }
 
  \begin{tablenotes}
          %\footnotesize   %% If you want them smaller like foot notes
         \item Note: % \textbf{${}^1$}: WRN-22 \citep{zagoruyko2016wide} original kernelstage is 16-16-32-64,
         \textbf{${}^2$}: ResNet{$\ast \ast$}: Variant ResNet(\citep{liu2018bi}, %\textbf{${}^3$}: J is  total projection number, \textbf{${}^4$}: number of Activation and Weights, 
         \textbf{${}^5$}: OPs similar,\textbf{${}^6$}: BN: BatchNorm,   \textbf{${}^7$}:SiMaN: activations \{-1,+1\}, weights \{0,+1\},   \textbf{${}^8$}:DIR-Net: journal version of IR-Net{\citep{ qin2020forward}}
        \end{tablenotes}
 
 \end{threeparttable}
\vspace{-1cm}
\end{table}

%%%%%%%%%%%%%%%%%%%%%%%%%%%%%%%%%%%%%%%%%%%%%%%%%%%%%%%%%%%%%%%%%%%%%%%%%%

\subsubsection{ImageNet Dataset}
\noindent ImageNet is a large images dataset that is usually used to test the trained model's performance. There are different versions of ImageNet dataset. The common version used for BNN is ILSVRC2012 ImageNet which was used for the competition dataset of "ImageNet Large Scale Visual Recognition Challenge 2012". The ILSVRC2012 ImageNet consists of three subset parts; training, validation and test dataset. Training dataset has more than 1.2 million color images in about 1000 classes. Validation dataset has 50000 color images and test dataset contains 100000 color images. 

\begin{table}[!h]
 
\begin{threeparttable}

\begin{tabular}{lllllllllll}
\hlineB{2.5}
\rowcolor{Gray}
\multicolumn{11}{c}{FPGA Platform}\\
\rowcolor{Gray}

BNN&Device & FPS & \vtop{\hbox{\strut Top-1}{\hbox{\strut Acc(\%)}}}& \vtop{\hbox{\strut Top-5}{\hbox{\strut Acc(\%)}}}  & \vtop{\hbox{\strut  Bits}{\hbox{\strut(W/A)}}} &  \vtop{\hbox{\strut  $F_{max}$ }{\hbox{\strut(MHz)}}}&Power(W)     & DSP& BRAM & LUT                 \\
\hlineB{2.5}
\rowcolor{LightCyan}
 Oc-BNN\citeyearpar{7965031} &\vtop{\hbox{\strut Zynq}{\hbox{\strut ZU9EG}}}&31.48&-&-&1/1&150&22&4&1367&-\\
 
 ReBNet\citeyearpar{ghasemzadeh2018rebnet} &\vtop{\hbox{\strut Virtex}{\hbox{\strut VCU108}}}&170&41.43&-&1/1&200&-&- &-&-\\
 \rowcolor{LightCyan}
 FP-BNN\citeyearpar{liang2018fp} &Stratix-V&862.1&42.9&66.8&1/1&150&26.2&384 &2210&230918\\

FracBNN\citeyearpar{zhang2021fracbnn} &\vtop{\hbox{\strut Zynq}{\hbox{\strut ZU3EG}}} & 48.1&71.8 & 90.1 & 1/1.4\footnotemark[2] &250&6.1    &224& 201& 50656                \\

    \hline

\end{tabular}

 \caption{\label{tab:table-name} BNN efficiency comparisons on ImageNet using FPGA}

  \begin{tablenotes}
          %\footnotesize   %% If you want them smaller like foot notes
         \item Note:  \textbf{${}^1$}: support materials from its code GitHub page, \textbf{${}^2$}:   1.4bit  based  on  the  analysis  of  quantized  activations, convolution still executes 1W/1A operations,\textbf{${}^3$}:  without bit-shift scales
        \end{tablenotes}
 
 \end{threeparttable}
\end{table}

\begin{table}[h]

    \begin{threeparttable}
    
 \caption{\label{tab:table-name}BNN performance comparisons on ImageNet}

\begin{tabular}{lllllll}
\hlineB{2.5}
\rowcolor{Gray}
 \multicolumn{7}{c}{Full Precision Base CNN} \\
\rowcolor{Gray}
\multicolumn{2}{l}{CNN Name}&\multicolumn{1}{l}{\vtop{\hbox{\strut Top-1 Acc}{\hbox{\strut(\%)}}}} &\multicolumn{2}{l}{\vtop{\hbox{\strut Top-5 Acc}{\hbox{\strut(\%)}}}}&\multicolumn{2}{l}{\vtop{\hbox{\strut FLOPs/OPs}{\hbox{\strut(x$10^{8}$)}}}}\\
\hlineB{2.5}
\rowcolor{LightCyan}
\multicolumn{2}{l}{AlexNet\citeyearpar{zhang2018lq}}&\multicolumn{1}{l}{57.1 }
                    & \multicolumn{2}{l}{80.2}&\multicolumn{2}{l}{-}\\

\multicolumn{2}{l}{ResNet-18\citeyearpar{zhang2018lq}}&\multicolumn{1}{l}{69.6 }
                    & \multicolumn{2}{l}{89.2}&\multicolumn{2}{l}{18.1\citeyearpar{liu2018bi,bethge2020meliusnet} }\\
\rowcolor{LightCyan}
\multicolumn{2}{l}{ResNet-34\citeyearpar{zhang2018lq}}&\multicolumn{1}{l}{73.3 }
                    & \multicolumn{2}{l}{91.3}&\multicolumn{2}{l}{36.6\citeyearpar{liu2018bi,bethge2020meliusnet} }\\

\multicolumn{2}{l}{ResNet-50\citeyearpar{zhang2018lq}}&\multicolumn{1}{l}{76.0 }
                    & \multicolumn{2}{l}{93.0}&\multicolumn{2}{l}{38.6\citeyearpar{zhu2020nasb}  }\\
\rowcolor{LightCyan} 
\multicolumn{2}{l}{ BN-Inception\citeyearpar{mishra2017wrpn}}&\multicolumn{1}{l}{71.64 }
                    & \multicolumn{2}{l}{-}&
                    \multicolumn{2}{l}{-}\\

\multicolumn{2}{l}{ MobileNet-v1 0.5\citeyearpar{howard2017mobilenets}}&\multicolumn{1}{l}{63.7 }
                    & \multicolumn{2}{l}{-}&
                    \multicolumn{2}{l}{1.49\citeyearpar{bethge2020meliusnet} }\\
\rowcolor{LightCyan}
\multicolumn{2}{l}{ MobileNet-v1 0.75\citeyearpar{howard2017mobilenets}}&\multicolumn{1}{l}{68.4 }
                    & \multicolumn{2}{l}{-}&
                    \multicolumn{2}{l}{3.25\citeyearpar{bethge2020meliusnet} }\\
 
\multicolumn{2}{l}{ MobileNet-v1 1.0\citeyearpar{howard2017mobilenets}}&\multicolumn{1}{l}{70.6 }
                    & \multicolumn{2}{l}{-}&
                    \multicolumn{2}{l}{5.69\citeyearpar{bethge2020meliusnet} }\\

\rowcolor{LightCyan}
\multicolumn{2}{l}{ MobileNet-v2\citeyearpar{sandler2018mobilenetv2}}&\multicolumn{1}{l}{71.53 }
                    & \multicolumn{2}{l}{-}&
                    \multicolumn{2}{l}{-}\\

\hlineB{2.5}
\rowcolor{Gray}
 \multicolumn{7}{c}{BNN Accuracy Performance} \\
 
 \rowcolor{Gray}

 BNN Name & Topology & \vtop{\hbox{\strut Top-1 Acc}{\hbox{\strut(\%)} }} & \vtop{\hbox{\strut Top-5 Acc}{\hbox{\strut(\%) }}} & \vtop{\hbox{\strut BOPs }{\hbox{\strut(x$10^{9}$)}}}&\vtop{\hbox{\strut FLOPs}{\hbox{\strut(x$10^{8}$) }}}&    \vtop{\hbox{\strut OPs}{\hbox{\strut(x$10^{8}$)  }}}       \\ 
\hlineB{2.5}

 \rowcolor{LightCyan}
{BNN\citeyearpar{courbariaux2016binarized}} & AlexNet & 27.9& 50.42  &1.70&1.20&1.47\\
 \rowcolor{LightCyan}
&ResNet-18 &{42.2
\citeyearpar{kim2020learning} } & {69.2
\citeyearpar{kim2020learning} } &1.70&1.31&1.67\\

{XNOR-Net\citeyearpar{rastegari2016xnor}} & AlexNet &44.2& 69.2  &-&-&-\\
&ResNet-18 &51.2 & 73.2&1.70&1.33&1.60\\
&ResNet-34 &56.49
\citeyearpar{martinez2020training}  &79.13
\citeyearpar{martinez2020training}&-&-&1.78\\

 \rowcolor{LightCyan}
{DoReFa-Net\citeyearpar{zhou2016dorefa} }   & {AlexNet} &40.1&-&-&-&-\\
 \rowcolor{LightCyan}
&{AlexNet}& 43.6(initialized) &-&-&-&- \\

{ABC-Net\citeyearpar{lin2017towards}}  & ResNet-18 &42.7&67.6&-&-&1.48\\&ResNet-18(3W, 1A)\footnotemark[1]{} &49.1&73.8&-&-&- \\
&ResNet-18(3W, 3A)\footnotemark[1]{} &61.0&83.2&-&-&- \\
&ResNet-18(3W, 5A)\footnotemark[1]{} &63.1&84.8 &-&-&-\\
&ResNet-18(5W, 1A)\footnotemark[1]{} &54.1&78.1 &-&-&- \\
&ResNet-18(5W, 3A)\footnotemark[1]{} &62.5&84.2 &-&-&5.20 \\
&ResNet-18(5W, 5A)\footnotemark[1]{} &65.0&85.9 &-&-&7.85 \\
&ResNet-34 &52.4&76.5 &-&-&- \\
&ResNet-34(3W, 3A)\footnotemark[1]{} &66.7&87.4 &-&-&- \\
&ResNet-34(5W, 5A)\footnotemark[1]{} &68.4&88.2 &-&-&- \\
&ResNet-50(5W, 5A)\footnotemark[1]{} &70.1&89.7 &-&-&- \\

 \rowcolor{LightCyan}
{WRPN\citeyearpar{mishra2017wrpn}}  & AlexNet(2x wide)\footnotemark[2]{} & 48.3
&-&&&\\
 \rowcolor{LightCyan}
 &ResNet-34 & 60.54&-&-&-&- \\
  \rowcolor{LightCyan}
&ResNet-34(2x wide)\footnotemark[2]{} &69.85&- &-&-&-\\
 \rowcolor{LightCyan}
&ResNet-34(3x wide)\footnotemark[2]{} &72.38&- &-&-&-\\
 \rowcolor{LightCyan}
&BN-Inception(2x wide)\footnotemark[2]{} &65.02&- &-&-&-\\

 SQ-BWN\citeyearpar{dong2017learning}  & AlexNet & 45.5        \citeyearpar{dong2019stochastic} & 70.6\citeyearpar{dong2019stochastic}   &-&-&-\\

 \rowcolor{LightCyan}
{BNN-RBNT\citeyearpar{darabi2018regularized}}  & AlexNet & 46.1& 75.7&-&-&-  \\
 \rowcolor{LightCyan}
& ResNet-18&53.01& 72.98&-&-&- \\

 { Bi-Real-Net\citeyearpar{liu2018bi}}  &  ResNet-18$\ast \ast$\footnotemark[9] & 56.4& 79.5 &1.68& 1.39& 1.63 \\
&ResNet-18$\ast \ast$\footnotemark[9]\citeyearpar{bethge2020meliusnet}&60.6& -&& & 1.14  \\

&ResNet-34$\ast \ast$\footnotemark[9]&62.2& 83.9&3.53& 1.39& 1.93  \\
  &ResNet-34$\ast \ast$\footnotemark[9]\citeyearpar{bethge2020meliusnet}&63.7& -&& & 1.43 \\
&{ResNet-50$\ast \ast$\footnotemark[9]\citeyearpar{zhuang2019structured}}&62.6& 83.9&& & \\
  &{ResNet-152$\ast \ast$\footnotemark[9]\citeyearpar{liu2020bi}}&64.5& -&10.7&4.48 & 6.15 \\
&{MobileNet-v1 1.0\citeyearpar{zhuang2019structured}}&58.2& -&-&-& - \\

 \rowcolor{LightCyan}
 PCNN\citeyearpar{gu2019projection}  & ResNet-18$\ast \ast$\footnotemark[9] & 57.3& 80.0  &-&-&1.63\\
 
 HadaNet\citeyearpar{akhauri2019hadanets} & AlexNet($\beta$w=4;$\beta$a=4)&  46.3& 71.2  &-&-&-\\
 & AlexNet($\beta$w=16;$\beta$a=2)&  47.3& 73.3  &-&-&-\\
  & ResNet-18($\beta$w=4;$\beta$a=4)&  53.3& 77.3 &-&-&-\\
 & ResNet-18($\beta$w=16;$\beta$a=2)& 53.8& 77.2  &-&-&-\\
\rowcolor{LightCyan}
  XNOR-Net++\citeyearpar{bulat2019xnor}   & AlexNet($\alpha,\beta, \gamma$)\footnotemark{} & 46.9& 71.0  &-&-&-\\
 \rowcolor{LightCyan}
&ResNet-18&55.5& 78.5 &-&-&-\\
\rowcolor{LightCyan}
&ResNet-18($\alpha_{1}$)\footnotemark[4]&56.1& 79.0&-&-&- \\
\rowcolor{LightCyan}
&ResNet-18($\alpha_{2},\beta_{1}$)\footnotemark[5]&56.7& 79.5&-&-&- \\
\rowcolor{LightCyan}
&ResNet-18($\alpha,\beta, \gamma$)\footnotemark[3]&57.1& 79.9&1.695&1.333&1.60 \\

    \hline
\end{tabular}

  \begin{tablenotes}
          %\footnotesize   %% If you want them smaller like foot notes
   
            \item Note: \textbf{${}^1$}: Nums W: number of Weights parallel, Nums A: number of Activations parallel, \textbf{${}^2$}: Nums x wide: number of filters, %3: $\alpha,\beta,\gamma$ statistically learned via channels, heights, weights, 4: $\alpha_{1}$  a dense scaling,one value for each output pixel, 5: $\alpha_{2}$ learns the statistics over the output channel dimension, $\beta_{1}$ learns it over the spatial dimensions, 6: Compact-Net\citep{tang2017train} uses 2 bits for activations while BNN-DL \citep{ding2019regularizing} only uses 1 bit, 7: SBN: State Batch Normalization\citep{NEURIPS2020_2a084e55}, 8: with only layer-wise scale factor to compare with XNOR-Net,
\textbf{${}^9$}: ResNet$\ast \ast$: Variant ResNet\citep{liu2018bi}
%, 10: K-layer dependency to improve single-layer dependency in depth-wise convolution, 11: OPs and Size similar, 12: BNN networks bagging, 13:OPs similar , 14: DMS-A,DMS-B: customized ResNet-18 with different channel number
        \end{tablenotes}
    \end{threeparttable}

\end{table}

%%%%%%%%%%%%%%%%%%%%%%%%%%%%%%%%%%%%%%%%%%%%%%%%%%%%%%%%%%%%%%%%%%%%%%%%%%
\clearpage

\begin{table}[!h]
\begin{threeparttable}
\caption{\label{tab:table-name}Continue Table 16.    }
\centering

\begin{tabular}{lllllll}
\hlineB{2.5}
\rowcolor{Gray}
 BNN Name & Topology & \vtop{\hbox{\strut Top-1 Acc}{\hbox{\strut(\%)} }} & \vtop{\hbox{\strut Top-5 Acc}{\hbox{\strut(\%) }}} & \vtop{\hbox{\strut BOPs }{\hbox{\strut(x$10^{9}$)}}}&\vtop{\hbox{\strut FLOPs}{\hbox{\strut(x$10^{8}$) }}}&    \vtop{\hbox{\strut OPs}{\hbox{\strut(x$10^{8}$)  }}}       \\ 

\hlineB{2.5}

{ Bop\citeyearpar{helwegen2019latent}}  & AlexNet(BNN\citeyearpar{courbariaux2016binarized})& 41.1& 65.4 &-&-&- \\
&AlexNet(XNOR-Net\citeyearpar{rastegari2016xnor})&45.9& 70.0&-&-&- \\
& ResNet-18$\ast \ast$\footnotemark[9]&56.6& 79.4&-&-&1.63 \\

  \rowcolor{LightCyan}
{ BNN-DL\citeyearpar{ding2019regularizing}}  & AlexNet(BNN\citeyearpar{courbariaux2016binarized})& 41.3& 65.8&-&-&-  \\
\rowcolor{LightCyan}
&AlexNet(XNOR-Net\citeyearpar{rastegari2016xnor})&47.8& 71.5&-&-&- \\
\rowcolor{LightCyan}
&{AlexNet(DoReFa-Net\citeyearpar{zhou2016dorefa})}&47.8& 71.5&-&-&- \\
\rowcolor{LightCyan}
&{AlexNet(Compact-Net\citeyearpar{tang2017train}\footnotemark[6])}&47.6& 71.9&-&-&- \\
\rowcolor{LightCyan}
&AlexNet(WRPN\citeyearpar{mishra2017wrpn})&53.8& 77.0&-&-&- \\

 Main/Subsidiary \citeyearpar{xu2019main}  
  & ResNet-18(78.6\% filters) & 50.13& - &-&-&-\\

\rowcolor{LightCyan}
{CCNN\citeyearpar{xu2019accurate}}    & AlexNet & 46.13& 70.9 &-&-&-\\ 
\rowcolor{LightCyan}
& ResNet-18 & 54.2& 77.9 &-&-&-\\

\vtop{\hbox{\strut Quantization Networks }{\hbox{\strut~\citeyearpar{yang2019quantization}}}}  & AlexNet & 47.9& 72.5 &-&-&-\\
& ResNet-18 & 53.6& 75.3&-&-&1.63 \\

\rowcolor{LightCyan}
{CI-BCNN\citeyearpar{wang2019learning}}  &  ResNet-18& 56.73& 80.12  &-&-&1.54\\
\rowcolor{LightCyan}
&ResNet-18$\ast \ast$\footnotemark[9]&59.90&84.18&-&-&$>$1.54 \\
\rowcolor{LightCyan}
& ResNet-34&62.41&84.35 &-&-&1.82\\
\rowcolor{LightCyan}
&ResNet-34$\ast \ast$\footnotemark[9]&64.93& 86.61&-&-&$>$1.82 \\

BONN\citeyearpar{gu2019bayesian}   & ResNet-18 & 59.3& 81.6 &-&-&-\\

\rowcolor{LightCyan}
CBCN\citeyearpar{liu2019circulant}   & ResNet-18(4W,4A)\footnotemark[1]{} & 61.4& 82.8 &-&-&6.56\\

RBCN\citeyearpar{liu2019rbcn} &ResNet-18 & 59.5& 81.6&-&-&- \\

\rowcolor{LightCyan}
{\vtop{\hbox{\strut Search Accurate }{\hbox{\strut\citeyearpar{shen2019searching}}}}}&Customized ResNet-18&68.64&88.46&-&-&4.95\\
\rowcolor{LightCyan}
&Customized ResNet-18&69.65&89.08&-&-&6.60\\

{Group-Net\citeyearpar{zhuang2019structured}}  & ResNet-18(4W,4A)\footnotemark[1]{} & 64.2& 85.6 &-&-&-\\
& ResNet-18(5W,5A)\footnotemark[1]{} & 64.8& 85.7 &-&-&-\\
& ResNet-18(8W,8A)\footnotemark[1]{} & 67.5& 88.0 &-&-&-\\
& ResNet-18$\ast \ast$\footnotemark[9](4W,4A)\footnotemark[1]{} & 66.3& 86.6&-&-&- \\
& ResNet-18$\ast \ast$\footnotemark[9](5W,5A)\footnotemark[1]{} & {67.0\citeyearpar{liu2018bi}}&{87.5\citeyearpar{liu2018bi}} &-&-&2.68\\
& ResNet-34(5W,5A)\footnotemark[1]{} & 68.5& 88.0 &-&-&-\\
& ResNet-34(8W,8A)\footnotemark[1]{} & 71.8& 90.4 &-&-&-\\
 & ResNet-34$\ast \ast$\footnotemark[9](5W,5A)\footnotemark[1]{} &
{70.5\citeyearpar{liu2018bi}}&{89.3(\citeyearpar{liu2018bi}}  &-&-&4.13\\
& ResNet-50(5W,5A)\footnotemark[1]{} & 69.5& 89.2&-&-&- \\
& ResNet-50(8W,8A)\footnotemark[1]{} & 72.8& 90.5 &-&-&-\\

\rowcolor{LightCyan}
{\vtop{\hbox{\strut BinaryDenseNet}{\hbox{\strut\citeyearpar{Bethge_2019_ICCV}} }} }& Customized DenseNet28 & 60.7& 82.4 &-&-&2.58 \\
\rowcolor{LightCyan}
&\vtop{\hbox{\strut Customized DenseNet28}{\hbox{\strut~\citeyearpar{bethge2020meliusnet} }}} & 62.6& - &-&-&2.09\\
\rowcolor{LightCyan}
& Customized DenseNet37  & 62.5& 83.9 &-&-&2.71\\
\rowcolor{LightCyan}
& \vtop{\hbox{\strut Customized DenseNet37 }{\hbox{\strut~(dilated)}}}  & 63.7& 84.7 &-&-&-\\
\rowcolor{LightCyan}
& \vtop{\hbox{\strut Customized DenseNet37}{\hbox{\strut~\citeyearpar{bethge2020meliusnet}}} } &64.2 & - &-&-&2.20\\
{BENN\citeyearpar{zhu2019binary}}  & AlexNet(3W,3A)\footnotemark[12]{} & \vtop{\hbox{\strut 48.8}{\hbox{\strut(bagging\citeyearpar{breiman1996bagging})}}}& -&-&-&-\\
& AlexNet(3W,3A)\footnotemark[12]{} & \vtop{\hbox{\strut 50.2 }{\hbox{\strut(boosting\citeyearpar{freund1997decision})}}}& - &-&-&-\\
& AlexNet(6W,6A)\footnotemark[12]{} & \vtop{\hbox{\strut52.0}{\hbox{\strut(bagging\citeyearpar{breiman1996bagging})}}}& - &-&-&-\\
& AlexNet(6W,6A)\footnotemark[12]{} & \vtop{\hbox{\strut54.3}{\hbox{\strut(boosting\citeyearpar{freund1997decision})}}}& - &-&-&-\\
& ResNet-18(3W,3A)\footnotemark[12]{} & \vtop{\hbox{\strut53.3}{\hbox{\strut(bagging\citeyearpar{breiman1996bagging})}}}& -&-&-&- \\

  \hline

\end{tabular}
 \begin{tablenotes}
          %\footnotesize   %% If you want them smaller like foot notes
   
            \item Note:  \textbf{${}^1$}: Nums W: number of Weights parallel, Nums A: number of Activations parallel,
            %2: Nums x wide: number of filters,
            \textbf{${}^3$}: $\alpha,\beta,\gamma$ statistically learned via channels, heights, weights, \textbf{${}^4$}: $\alpha_{1}$  a dense scaling,
one value for each output pixel, \textbf{${}^5$}: $\alpha_{2}$ learns the statistics over the output channel dimension, $\beta_{1}$ learns it over the spatial dimensions, \textbf{${}^6$}: Compact-Net\citep{tang2017train} uses 2 bits for activations while BNN-DL \citep{ding2019regularizing} only uses 1 bit, %7: SBN: State Batch Normalization\citep{NEURIPS2020_2a084e55}, 8: with only layer-wise scale factor to compare with XNOR-Net, 
\textbf{${}^9$}: ResNet$\ast \ast$: Variant ResNet\citep{liu2018bi},%, 10: K-layer dependency to improve single-layer dependency in depth-wise convolution, 11: OPs and Size similar, 
\textbf{${}^{12}$}: BNN networks bagging, %13:OPs similar %, 14: DMS-A,DMS-B: customized ResNet-18 with different channel number
        \end{tablenotes}
    \end{threeparttable}
\end{table}

   \clearpage
\begin{table}[t]
 \begin{threeparttable}
 \caption{\label{tab:table-name}Continue Table 16.    }
\centering

\begin{tabular}{lllllll}

 \hlineB{2.5}
 \rowcolor{Gray}
 BNN Name & Topology & \vtop{\hbox{\strut Top-1 Acc}{\hbox{\strut(\%)} }} & \vtop{\hbox{\strut Top-5 Acc}{\hbox{\strut(\%) }}} & \vtop{\hbox{\strut BOPs }{\hbox{\strut(x$10^{9}$)}}}&\vtop{\hbox{\strut FLOPs}{\hbox{\strut(x$10^{8}$) }}}&    \vtop{\hbox{\strut OPs}{\hbox{\strut(x$10^{8}$)  }}}       \\ 
 
 \hlineB{2.5}

 \rowcolor{LightCyan}
{BENN\citeyearpar{zhu2019binary}}
& ResNet-18(3W,3A)\footnotemark[12]{} & \vtop{\hbox{\strut53.6}{\hbox{\strut(boosting\citeyearpar{freund1997decision})}}}& -&-&-&- \\
\rowcolor{LightCyan}
& ResNet-18(6W,6A)\footnotemark[12]{} &\vtop{\hbox{\strut57.9}{\hbox{\strut(bagging\citeyearpar{breiman1996bagging})}}}& - &-&-&-\\
\rowcolor{LightCyan}
& ResNet-18(6W,6A)\footnotemark[12]{} & \vtop{\hbox{\strut61.0}{\hbox{\strut(boosting\citeyearpar{freund1997decision})}}}& -&-&-&- \\
%&& ResNet-34(6W,6A)\footnotemark[12]{} & \shortstack[c]{64.7\\(\cite{zhuang2019structured}reported)}& \shortstack[c]{84.4\\(\cite{zhuang2019structured}reported)}&&& \\
%&& ResNet-50(6W,6A)\footnotemark[12]{} & \shortstack[c]{66.2\\(\cite{zhuang2019structured}reported)}& \shortstack[c]{65.8\\(\cite{zhuang2019structured}reported)}&&& \\
%&& MobileNet-v1 1.0(6W,6A)\footnotemark[12]{} & \shortstack[c]{63.0\\(\cite{zhuang2019structured}reported)}& -&-&-&- \\

{IR-Net\citeyearpar{qin2020forward}}  & ResNet-18$\ast \ast$\footnotemark[9] & 58.1& 80.0 &-&-&1.63 \\
& ResNet-34$\ast \ast$\footnotemark[9] & 62.9& 84.1 &-&-&-\\
\rowcolor{LightCyan}
{BATS\citeyearpar{bulat2020bats} } &Customized  & 60.4 & 83.0&1.149 & 0.805&0.985\\
\rowcolor{LightCyan}
&Customized(2x-wider) &66.1&87.0&2.157&1.210&1.547\\

{BNAS\citeyearpar{kim2020learning} } &\vtop{\hbox{\strut Customized\footnotemark[13] ResNet-18}{\hbox{\strut ~(XNOR-Net)}}} & 57.69 & 79.89&- & -&1.48\\

&  Customized\footnotemark[13]  ResNet-18$\ast \ast$\footnotemark[9] & 58.76& 80.61&- & - &1.63\\
&\vtop{\hbox{\strut Customized\footnotemark[13] ResNet-34  }{\hbox{\strut(XNOR-Net)}}} & 58.99& 80.85&- & - &1.78\\
&Customized\footnotemark[13]   ResNet-34$\ast \ast$\footnotemark[9] & 59.81& 81.61&- &-  &1.93\\
& \vtop{\hbox{\strut  Customized\footnotemark[13] ResNet-18 }{\hbox{\strut(4W,4A)\footnotemark[1]{}(CBCN) }}} & 63.51& 83.91& -& - &6.56\\

\rowcolor{LightCyan}
{NASB\citeyearpar{zhu2020nasb}}  & Customized\footnotemark[13] ResNet-18& 60.5& 82.2&-&-&1.71\\
\rowcolor{LightCyan}
& Customized\footnotemark[13] ResNet-34 & 64.0& 84.7&-&-&2.01 \\
\rowcolor{LightCyan}
& Customized\footnotemark[13] ResNet-50 & 65.7& 85.8&-&-&6.18 \\

{Si-BNN\citeyearpar{wang2020sparsity}}  & AlexNet& 50.5& 74.6&-&-&- \\
& ResNet-18 & 58.9& 81.3&-&-&- \\
& ResNet-18$\ast \ast$\footnotemark[9] & 59.7& 81.8&-&-&- \\
& ResNet-34 & 63.3& 84.4&-&-& -\\
 
\rowcolor{LightCyan}
{LNS\citeyearpar{han2020training}}  & AlexNet\footnotemark[8] & 44.4& - &-&-&-\\
\rowcolor{LightCyan}
& ResNet-18 & 59.4& 81.7 &-&-&1.63\\
 
{SLB\citeyearpar{NEURIPS2020_2a084e55}}  & ResNet-18(w/o SBN\footnotemark[7]) & 61.0& 82.9 &-&-&-\\
& ResNet-18 & 61.3& 83.1&-&-&- \\

\rowcolor{LightCyan}
{Real-to-Bin\citeyearpar{martinez2020training}}  & ResNet-18$\ast \ast$\footnotemark[9](baseline)  & 60.9 & 83.0&1.68 &1.54 &1.63\\
\rowcolor{LightCyan}
 & \vtop{\hbox{\strut ResNet-18$\ast \ast$\footnotemark[9]}{\hbox{\strut(BNN-Adam\citeyearpar{liu2021how})}}}& 63.2  & 84.0&1.68 &1.54 &1.80\\
 \rowcolor{LightCyan}
& ResNet-18$\ast \ast$\footnotemark[9] & 65.4&86.2& 1.68& 1.56 &1.83\\

{ProxyBNN\citeyearpar{he2020proxybnn}}  & AlexNet& 51.4& 75.5&-&-&- \\
& ResNet-18 & 58.7& 81.2&-&-&- \\
& ResNet-18$\ast \ast$\footnotemark[9] & 63.7& 84.8&-&-&- \\
& ResNet-34$\ast \ast$\footnotemark[9]  & {66.3}&86.5&-&-&- \\

\rowcolor{LightCyan}
{RBNN\citeyearpar{lin2020rotated}}  & ResNet-18 &59.9  & 81.9&- & -&-\\
\rowcolor{LightCyan}
& ResNet-34& 63.1&84.4&- &- &-\\

{BinaryDuo\citeyearpar{kim2020binaryduo} }  & AlexNet & 52.7 & 76.0&- &- &1.19\\
& ResNet-18  & 60.4& 82.3&- &-  &1.64\\
& ResNet-18$\ast \ast$\footnotemark[9]   & 60.9& 82.6& -& - &1.64\\

\rowcolor{LightCyan}
MoBiNet-Mid\citeyearpar{phan2020mobinet}   & Customized\footnotemark[10](K=3) & 53.47 & 76.46&- &- &0.49\\
\rowcolor{LightCyan}
& Customized\footnotemark[10](K=4)  & 54.4& 77.5&- &-  &0.52\\

{MeliusNet\citeyearpar{bethge2020meliusnet}}  & \vtop{\hbox{\strut 
MeliusNetA }{\hbox{\strut(Customized\footnotemark[11] ResNet-18$\ast \ast$\footnotemark[9])}} }& 63.4 & 84.2&4.85 &0.86 &1.62\\
& \vtop{\hbox{\strut 
MeliusNetB }{\hbox{\strut(Customized\footnotemark[11] ResNet-34$\ast \ast$\footnotemark[9]) }}} & 65.7& 85.9&5.72 &1.06  &1.96\\
&\vtop{\hbox{\strut 
MeliusNetC }{\hbox{\strut(Customized\footnotemark[11] MobileNetv1 0.5)} }} & 64.1& 85.0&4.35 & 0.82 &1.50\\

&\vtop{\hbox{\strut
MeliusNet42 }{\hbox{\strut(Customized\footnotemark[11] MobileNetv1 0.75)}} }  & 69.2& 88.3&9.69 & 1.74 &3.25\\
 & \vtop{\hbox{\strut MeliusNet59}{\hbox{\strut(Customized\footnotemark[11] MobileNet-v1 1.0)} }} & 71.0& 89.7&18.3 &2.45  &5.25\\
\hline
 
\end{tabular}

 \begin{tablenotes}
          %\footnotesize   %% If you want them smaller like foot notes
   
            \item Note:  \textbf{${}^1$}: Nums W: number of Weights parallel, Nums A: number of Activations parallel, 
            %2: Nums x wide: number of filters, 3: $\alpha,\beta,\gamma$ statistically learned via channels, heights, weights, 4: $\alpha_{1}$  a dense scaling, one value for each output pixel, 5: $\alpha_{2}$ learns the statistics over the output channel dimension, $\beta_{1}$ learns it over the spatial dimensions, 6: Compact-Net\citep{tang2017train} uses 2 bits for activations while BNN-DL \citep{ding2019regularizing} only uses 1 bit, 
          \textbf{${}^7$}: SBN: State Batch Normalization\citep{NEURIPS2020_2a084e55}, \textbf{${}^8$}: with only layer-wise scale factor to compare with XNOR-Net, \textbf{${}^9$}: ResNet$\ast \ast$: Variant ResNet\citep{liu2018bi}, 
            \textbf{${}^{10}$}: K-layer dependency to improve single-layer dependency in depth-wise convolution, \textbf{${}^{11}$}: OPs and Size similar, 
           \textbf{${}^{12}$}: BNN networks bagging, \textbf{${}^{13}$}:OPs similar %, 14: DMS-A,DMS-B: customized ResNet-18 with different channel number
        \end{tablenotes}
    \end{threeparttable}
\end{table}
          
     \clearpage

\begin{table}[!h]
\caption{\label{tab:table-name}Continue Table 16.    }
    \begin{threeparttable}
\centering

\begin{tabular}{lllllll}

 \hlineB{2.5}
 \rowcolor{Gray}
 BNN Name & Topology & \vtop{\hbox{\strut Top-1 Acc}{\hbox{\strut(\%)} }} & \vtop{\hbox{\strut Top-5 Acc}{\hbox{\strut(\%) }}} & \vtop{\hbox{\strut BOPs }{\hbox{\strut(x$10^{9}$)}}}&\vtop{\hbox{\strut FLOPs}{\hbox{\strut(x$10^{8}$) }}}&    \vtop{\hbox{\strut OPs}{\hbox{\strut(x$10^{8}$)  }}}       \\ 
 
 \hlineB{2.5}
\rowcolor{LightCyan}
{MeliusNet\citeyearpar{bethge2020meliusnet}}
& \vtop{\hbox{\strut 
MeliusNet22 }{\hbox{\strut(Customized\footnotemark[11] BDenseNet28\citeyearpar{Bethge_2019_ICCV}) }}} & 63.6&84.7 &4.62 &1.35  &2.08\\
\rowcolor{LightCyan}
&\vtop{\hbox{\strut 
MeliusNet29}{\hbox{\strut(Customized\footnotemark[11] BDenseNet37\citeyearpar{Bethge_2019_ICCV})}}}   & 65.8& 86.2& 5.47&1.29  &2.14\\

\vtop{\hbox{\strut Binarized MobileNet}{\hbox{\strut\citeyearpar{phan2020binarizing}}} }  & Customized MobileNet-v3 & 51.06 & 74.18& -& -&0.33\\
& Customized MobileNet-v2  & 59.30& 81.00& -& - &0.62\\
& Customized MobileNet-v1  & 60.90& 82.60&- &  -&1.54\\

\rowcolor{LightCyan}
{DMS\citeyearpar{Li_2020_ICLR_NAS}}&Customized(DMS-A)\footnotemark[15]&60.20& 82.94&-&-&-\\
\rowcolor{LightCyan}
&Customized(DMS-B)\footnotemark[15]&67.93&87.84&-&-&-\\

\vtop{\hbox{\strut High-Capacity-Expert}{\hbox{\strut \citeyearpar{bulat2020high}}}}&Customized(4 experts)&67.5& 87.5&1.7&1.1&1.37\\
&Customized(4 experts)\footnotemark[18]&70.0& 89.2&1.7&1.1&1.37\\
&Customized(4 experts)\footnotemark[19]&71.2& 90.1&1.7&1.1&1.37\\

\rowcolor{LightCyan}
{ReActNet\citeyearpar{liu2020reactnet} }& \vtop{\hbox{\strut ReActNet-18 }{\hbox{\strut(Customized ResNet-18)}}} & 65.5\citeyearpar{chen2021bnn}&- & -& -& -\\
\rowcolor{LightCyan}
&\vtop{\hbox{\strut ReActNet-A }{\hbox{\strut(Customized MobileNet-v1)}}} & 69.4&88.6\citeyearpar{liu2021how}& 4.82&0.12 &0.87 \\
\rowcolor{LightCyan}
&\vtop{\hbox{\strut ReActNet-A(BNN-Adam\citeyearpar{liu2021how}) }{\hbox{\strut(Customized MobileNet-v1)}}} & 70.5&89.1& 4.82&0.12 &0.87 \\

\rowcolor{LightCyan}
&\vtop{\hbox{\strut ReActNet-B }{\hbox{\strut(Customized MobileNet-v1)}}}& 70.1&-&4.69& 0.44& 1.63\\
\rowcolor{LightCyan}
& \vtop{\hbox{\strut ReActNet-C }{\hbox{\strut(Customized MobileNet-v1)}}} & 71.4&- & 4.69& 1.40& 2.14\\

{MD-tanh-s\citeyearpar{ajanthan2021mirror} }  & ResNet-18$\ast \ast$\footnotemark[9] & 60.3 & 82.3&- &- &-\\
& ResNet-18$\ast \ast$\footnotemark[9](dilated)  & 62.8&84.3& &-  &-\\

\rowcolor{LightCyan}
{UniQ\citeyearpar{pham2021training} }  & ResNet-18 & 60.5 & -&- &- &-\\
\rowcolor{LightCyan}
& ResNet-34  & 65.8&-& & - &-\\
\rowcolor{LightCyan}
& MobileNet-v2  & 23.2&- &- & - &-\\

{IA-BNN\citeyearpar{kim2020improving} } & AlexNet(BNN\citeyearpar{courbariaux2016binarized}) & 42.1&66.6 & -& -& -\\
& AlexNet(XNOR-Net\citeyearpar{rastegari2016xnor})& 45.6&69.6& -& - &-\\
& ResNet-18(XNOR-Net\citeyearpar{rastegari2016xnor}) & 54.2&77.6 & -& - &-\\
& ResNet-18$\ast \ast$\footnotemark[9]  &57.2&80.2&- & - &-\\
& ResNet-34$\ast \ast$\footnotemark[9] & 62.8&84.5 &- &-  &\\

\rowcolor{LightCyan}
FracBNN\citeyearpar{zhang2021fracbnn}&Customized(1W/1.4A)\footnotemark[15]&71.8&90.1&7.30&-&-\\
\vtop{\hbox{\strut Group-Net Extend}{\hbox{\strut\citeyearpar{zhuang2019structured,zhuang2021structured}}}}  & ResNet-18$\ast \ast$\footnotemark[9](4W,4A)\footnotemark[1]{} & 68.2&88.3 &-&-&- \\
& ResNet-34$\ast \ast$\footnotemark[9](4W,4A)\footnotemark[1]{}& 72.2&90.5&-&-&-\\
& ResNet-50$\ast \ast$\footnotemark[9](4W,4A)\footnotemark[1]{} & 73.4&91.0&-&-&-\\
& MobileNet-v1 1.0  &70.8&-&-&-&-\\
\rowcolor{LightCyan}
BCNN   \citeyearpar{redfern2021bcnn}&Customized(P=1)\footnotemark[16]&69.0&-&2.41&-&1.31\\
\rowcolor{LightCyan}
&Customized(P=2)\footnotemark[16]&71.2&-&4.83&-&2.08\\

MPT\citeyearpar{diffenderfer2021multi}& WRN-34(60\% weights pruned)&45.06&-&-&-&-\\ 
& WRN-34+BN\footnotemark[17](60\% weights pruned)&52.07&-&-&-&-\\ 

\rowcolor{LightCyan}
BNN-BN-free\citeyearpar{chen2021bnn}&\vtop{\hbox{\strut ReActNet-18 }{\hbox{\strut(Customized ResNet-18)}}}&61.1&-&-&-&-\\
\rowcolor{LightCyan}
&\vtop{\hbox{\strut ReActNet-A }{\hbox{\strut(Customized MobileNet-v1)}}}&68.0&-&-&-&-\\

ReCU \citeyearpar{xu2021recu}&  ResNet-18 &61.0&82.6&-&-&-\\ 
& ReActNet-18&66.4&86.5&-&-&-\\ 
& ResNet-34&65.1&85.8&-&-&-\\ 

\rowcolor{LightCyan}
DA-BNN\citeyearpar{zhao2021data}&  ResNet-18$\ast \ast$\footnotemark[9] &63.1&84.3&-&1.69&-\\ 
\rowcolor{LightCyan}
& ReActNet-18&66.3&86.7&-&1.69&-\\

\hline

\end{tabular}
 \begin{tablenotes}
          %\footnotesize   %% If you want them smaller like foot notes
   
            \item Note:  \textbf{${}^1$}: Nums W: number of Weights parallel, Nums A: number of Activations parallel,
            %2: Nums x wide: number of filters, 3: $\alpha,\beta,\gamma$ statistically learned via channels, heights, weights, 4: $\alpha_{1}$  a dense scaling,one value for each output pixel, 5: $\alpha_{2}$ learns the statistics over the output channel dimension, $\beta_{1}$ learns it over the spatial dimensions, 6: Compact-Net\citep{tang2017train} uses 2 bits for activations while BNN-DL \citep{ding2019regularizing} only uses 1 bit, 7: SBN: State Batch Normalization\citep{NEURIPS2020_2a084e55}, 8: with only layer-wise scale factor to compare with XNOR-Net, 
         
            \textbf{${}^9$}: ResNet$\ast \ast$: Variant ResNet\citep{liu2018bi}, \textbf{${}^{10}$}: K-layer dependency to improve single-layer dependency in depth-wise convolution, \textbf{${}^{11}$}: OPs and Size similar, 
            %12: BNN networks bagging, 
            %13:OPs similar , 14: DMS-A,DMS-B: customized ResNet-18 with different channel number, 
           \textbf{${}^{15}$}: 1.4bit based on the analysis of quantized activations, \textbf{${}^{16}$}: number of parallel branches, \textbf{${}^{17}$}: BN:BatchNorm
           \textbf{${}^{18}$}: training strategy from Real-to-Bin\citep{martinez2020training}, \textbf{${}^{19}$}: Improved training strategy from Real-to-Bin\citep{martinez2020training}
        \end{tablenotes}
    \end{threeparttable}

\end{table}
                             
%%%%%%%%%%%%%%%%%%%%%%%%%%%%%%%%%%%%%%%%%%%%%%%%%%
\clearpage
\begin{table}[!h]
\caption{\label{tab:table-name}Continue Table 16.    }
    \begin{threeparttable}
\centering

\begin{tabular}{lllllll}

 \hlineB{2.5}
 \rowcolor{Gray}
 BNN Name & Topology & \vtop{\hbox{\strut Top-1 Acc}{\hbox{\strut(\%)} }} & \vtop{\hbox{\strut Top-5 Acc}{\hbox{\strut(\%) }}} & \vtop{\hbox{\strut BOPs }{\hbox{\strut(x$10^{9}$)}}}&\vtop{\hbox{\strut FLOPs}{\hbox{\strut(x$10^{8}$) }}}&    \vtop{\hbox{\strut OPs}{\hbox{\strut(x$10^{8}$)  }}}       \\ 
 
 \hlineB{2.5}

   FDA \citeyearpar{xu2021learning}&    AlexNet&   46.2&   69.7&-&-&-\\ 
&     ResNet-18 &   62.2&   82.3&-&-&-\\ 
&    ReActNet-18&   66.0&   86.4&-&-&-\\

\rowcolor{LightCyan}   LCR\citeyearpar{shang2022lipschitz} &   ResNet-18&   59.6&   81.6&-&-&-\\ 
\rowcolor{LightCyan}
&    ReActNet-18&   69.8&   85.7&-&-&-\\ 
\rowcolor{LightCyan}
&    ReActNet-34&   63.5&   84.6&-&-&-\\

   Bi-half\citeyearpar{li2022equal}&     ResNet-18 &   60.4&   82.86&-&-&-\\ 

&    ResNet-34&   64.17&   85.36&-&-&-\\ 

\rowcolor{LightCyan}
   SiMaN\footnotemark[7]\citeyearpar{lin2022siman}&     ResNet-18 &   60.1&   82.3&-&-&-\\ 
\rowcolor{LightCyan}
&    ReActNet-18&   66.1&   85.9&-&-&-\\ 
\rowcolor{LightCyan}
&   ResNet-34&   63.9&   84.8&-&-&-\\

   AdaBin\citeyearpar{tu2022adabin}&     AlexNet &   53.9&   77.6&-&-&-\\ 
&    ResNet-18&   63.1&   84.3&-&-&-\\ 
&    ReActNet-18&   66.4&   86.5&   1.69&   1.41&   1.67\\ 

&   ResNet-34&   66.4&   86.6&-&-&-\\ 

\rowcolor{LightCyan}
   DyBNN\citeyearpar{zhang2022dynamic}&     ReActNet-18&   67.4&   87.5&   -&   -&   -\\ 
\rowcolor{LightCyan}
&    \vtop{\hbox{\strut ReActNet-A }{\hbox{\strut(Customized MobileNet-v1)}}}&   71.2&   89.8&   -&   -&   -\\

   DIR-Net\footnotemark[8]\citeyearpar{qin2022distribution}&     ResNet-18&   60.4&   81.9&-&-&-\\ 
&   ReActNet-18&   66.5&   87.1&   -&   -&   -\\ 
&   ResNet-34&   64.1&   85.3&   -&   -&   -\\ 
&   ReActNet-34&   67.9&   88.2&   -&   -&   -\\

   BiMLP-S\citeyearpar{xu2022bimlp}&     Customized MLP &   70.0&   89.6&   2.25&   1.21&   1.56\\ 

&     Customized MLP &   72.7&   91.1&   4.32&   1.21&   1.88\\

\hline

\end{tabular}
 \begin{tablenotes}
          %\footnotesize   %% If you want them smaller like foot notes
   
            \item Note:  \textbf{${}^1$}: Nums W: number of Weights parallel, Nums A: number of Activations parallel,
            %2: Nums x wide: number of filters, 3: $\alpha,\beta,\gamma$ statistically learned via channels, heights, weights, 4: $\alpha_{1}$  a dense scaling,one value for each output pixel, 5: $\alpha_{2}$ learns the statistics over the output channel dimension, $\beta_{1}$ learns it over the spatial dimensions, 6: Compact-Net\citep{tang2017train} uses 2 bits for activations while BNN-DL \citep{ding2019regularizing} only uses 1 bit, 7: SBN: State Batch Normalization\citep{NEURIPS2020_2a084e55}, 8: with only layer-wise scale factor to compare with XNOR-Net, 
          \textbf{${}^7$}:SiMaN: activations \{-1,+1\}, weights \{0,+1\}, \textbf{${}^8$}:DIR-Net: journal version of IR-Net{\citep{ qin2020forward}},
            \textbf{${}^9$}: ResNet$\ast \ast$: Variant ResNet\citep{liu2018bi}, \textbf{${}^{10}$}: K-layer dependency to improve single-layer dependency in depth-wise convolution, \textbf{${}^{11}$}: OPs and Size similar, 
            %12: BNN networks bagging, 
            %13:OPs similar , 14: DMS-A,DMS-B: customized ResNet-18 with different channel number, 
           \textbf{${}^{15}$}: 1.4bit based on the analysis of quantized activations, \textbf{${}^{16}$}: number of parallel branches, \textbf{${}^{17}$}: BN:BatchNorm
           \textbf{${}^{18}$}: training strategy from Real-to-Bin\citep{martinez2020training}, \textbf{${}^{19}$}: Improved training strategy from Real-to-Bin\citep{martinez2020training}
        \end{tablenotes}
    \end{threeparttable}

\end{table}
%%%%%%%%%%%%%%%%%%%%%%%%%%%%%%%%%%%%%%%%%%%%%%%%%%%%%%%%%%%%%%%%%%%%%%%%%%%

\subsection{Point Cloud Classification}
 \vspace{-0.2cm}
Different from 2D image classification tasks, 3D tasks using BNN are much more challenging as binarization will amplify information loss during aggregating point-wise features in pooling layers and lead to huge distortion at the point-wise feature extraction stage. BiPointNet \citep{qin2020bipointnet} proposed first binarization method for learning on 3D point cloud. Tables 21-22 list reported comparison results from \citep{qin2020bipointnet}. The benchmark dataset for testing accuracy used ModelNet40 \citep{wu20153d}, which contains 12311 pre-aligned shapes from 40 categories.
 \vspace{-0.1cm}

\begin{table}[!h]

\begin{threeparttable}

\begin{tabular}{llllllll}
\hlineB{2.5} 
  \rowcolor{Gray}
\multicolumn{8}{c}{Time Cost}\\
  \rowcolor{Gray}
\multicolumn{2}{l}{Method\;\;\;\;\;\;\;\;\;}&\multicolumn{2}{l}{Device}&\multicolumn{2}{l}{Bits(W/A)}&\multicolumn{2}{l}{Time(ms)}\\
\hlineB{2.5} 
 \rowcolor{LightCyan}
\multicolumn{2}{l}{{PointNet\citep{qi2017pointnet}}}&\multicolumn{2}{l}{Raspberry Pi 3B(ARM Cortex-A53)}&\multicolumn{2}{l}{32/32}&\multicolumn{2}{l}{131.8}\\
\multicolumn{2}{l}{BiPointNet\citep{qin2020bipointnet}}&\multicolumn{2}{l}{Raspberry Pi 3B(ARM Cortex-A53)}&\multicolumn{2}{l}{1/1}&\multicolumn{2}{l}{9}\\
 \rowcolor{LightCyan}
\multicolumn{2}{l}{{PointNet\citep{qi2017pointnet}}}&\multicolumn{2}{l}{Raspberry Pi 4B(ARM Cortex-A72)}&\multicolumn{2}{l}{32/32}&\multicolumn{2}{l}{67.3}\\
\multicolumn{2}{l}{BiPointNet\citep{qin2020bipointnet}}&\multicolumn{2}{l}{Raspberry Pi 4B(ARM Cortex-A72)}&\multicolumn{2}{l}{1/1}&\multicolumn{2}{l}{5.5}\\

\hlineB{2.5} 
  \rowcolor{Gray}
\multicolumn{8}{c}{Storage Usage}\\
 \rowcolor{Gray}
 
\multicolumn{2}{l}{Method}&\multicolumn{2}{l}{Device}&\multicolumn{2}{l}{Bits(W/A)}&\multicolumn{2}{l}{Storage(MB)}\\
\hlineB{2.5} 
 \rowcolor{LightCyan}
\multicolumn{2}{l}{{PointNet\citep{qi2017pointnet}}}&\multicolumn{2}{l}{Raspberry Pi 4B(ARM Cortex-A72)}&\multicolumn{2}{l}{32/32}&\multicolumn{2}{l}{3.16}\\
\multicolumn{2}{l}{BiPointNet\citep{qin2020bipointnet}}&\multicolumn{2}{l}{Raspberry Pi 4B(ARM Cortex-A72)}&\multicolumn{2}{l}{1/1}&\multicolumn{2}{l}{0.17}\\

    \hline

\end{tabular}
 \caption{\label{tab:table-name} 3D BNN efficiency comparisons  using Arm-based platform}
 
  \begin{tablenotes}
          %\footnotesize   %% If you want them smaller like foot notes
         \item Note:  \textbf{${}^1$}: support materials from its code GitHub page, \textbf{${}^2$}:   1.4bit  based  on  the  analysis  of  quantized  activations, convolution still executes 1W/1A operations,\textbf{${}^3$}:  without bit-shift scales
        \end{tablenotes}
 
 \end{threeparttable}
 \vspace{-0.5cm}
\end{table}
\vspace{-0.4cm}
\begin{table}[!h]
   \begin{threeparttable}
\centering

\begin{tabular}{llll}

\hlineB{2.5} 
  \rowcolor{Gray}
Method/(W/A)&Based Model  & Aggregation & OA(\%)  \\
\hlineB{2.5} 
\rowcolor{LightCyan}
{Full precision(32/32)}&PointNet(Vanilla)\citep{qi2017pointnet}&Max&86.8\\
\rowcolor{LightCyan}
&{PointNet\citep{qi2017pointnet}}&Avg&86.5\\
\rowcolor{LightCyan}
&&Max&88.2\\
\rowcolor{LightCyan}
&PointNet++\citep{qi2017pointnet++}&Max&90.0\\
\rowcolor{LightCyan}
&PointCNN\citep{li2018pointcnn}&Avg&90.0\\
\rowcolor{LightCyan}
&DGCNN\citep{wang2019dynamic}&Max&89.2\\
\rowcolor{LightCyan}
&PointConv\citep{wu2019pointconv}&-&90.8\\

{BNN\citep{courbariaux2016binarized}(1/1)}&{PointNet\citep{qi2017pointnet}}&Max&7.1\\
&&EMA-max&16.2\\
\rowcolor{LightCyan}
{IR-Net\citep{qin2020forward}(1/1)}&{PointNet\citep{qi2017pointnet}}&Max&7.3\\
\rowcolor{LightCyan}
&&EMA-max&63.5\\

{Bi-Real-Net\citep{liu2018bi}(1/1)}&{PointNet\citep{qi2017pointnet}}&Max&4.0\\
&&EMA-max&77.5\\

\rowcolor{LightCyan}
{ABC-Net\citep{lin2017towards}(1/1)}&{PointNet\citep{qi2017pointnet}}&Max&4.1\\
\rowcolor{LightCyan}
&&EMA-max&77.8\\

{XNOR-Net++\citep{bulat2019xnor}(1/1)}&{PointNet\citep{qi2017pointnet}}&Max&4.1\\
&&EMA-max&78.4\\
\rowcolor{LightCyan}
{XNOR-Net\citep{rastegari2016xnor}(1/1)}&PointNet(Vanilla)\citep{qi2017pointnet}&Max&61.0\\
\rowcolor{LightCyan}
&{PointNet\citep{qi2017pointnet}}&Max&64.9\\
\rowcolor{LightCyan}
&&EMA-max&81.9\\
\rowcolor{LightCyan}
&PointNet++\citep{qi2017pointnet++}&Max&63.1\\
\rowcolor{LightCyan}
&PointCNN\citep{li2018pointcnn}&Avg&83.0\\
\rowcolor{LightCyan}
&DGCNN\citep{wang2019dynamic}&Max&51.5\\
\rowcolor{LightCyan}
&PointConv\citep{wu2019pointconv}&-&83.1\\

{BiPointNet\citep{qin2020bipointnet}(1/1)}&PointNet(Vanilla)\citep{qi2017pointnet}&EMA-max&85.6\\
&{PointNet\citep{qi2017pointnet}}&Max&4.1\\
&&EMA-max&86.4\\
&PointNet++\citep{qi2017pointnet++}&EMA-max&87.8\\
&PointCNN\citep{li2018pointcnn}&EMA-avg&83.8\\
&DGCNN\citep{wang2019dynamic}&EMA-max&83.4\\
&PointConv\citep{wu2019pointconv}&-&87.9\\

\hline
\end{tabular}

 \caption{\label{tab:table-name}Accuracy Comparison for Points Cloud Classification }
 
  \begin{tablenotes}
          %\footnotesize   %% If you want them smaller like foot notes
   
            \item Note:  EMA-avg, EMA-max was proposed in BiPointNet\citep{qin2020bipointnet}, OA: overall accuracy 
        \end{tablenotes}
    \end{threeparttable}
 
\end{table}

\newpage
\subsection{Object Detection}
Object detection is a more complex and difficult task than 2D image classification. Recently, there are a few published BNN works on objects detection. \citet{sun2018fast} propose a fast object detection based on BNN \citeyearpar{courbariaux2016binarized}. \citet{Bethge_2019_ICCV} apply their proposed BinaryDenseNet to object detection, which has a comparable accuracy performance compared to full-precision Yolo.
ASDA-FRCNN\citep{10.1145/3410338.3412340} applies the experience from Modulated-Convolutional-Networks\citep{Wang_2018_CVPR} to represent full-precision kernel with BNN-based amplitude and direction, and designs a new loss function to reconstruct the full-precision kernels. \citet{wang2020bidet} propose BiDet which employs the information bottleneck (IB) principle to remove redundancy information for taking full advantage of BNN and concentrate posteriors on informative detection prediction via learning the sparse object before.  \citet{zhao2021data} apply their proposed re-scaling BNN method DA-BNN to Object Detection.  \citet{Xu_2021_CVPR} propose LWS-Det which is a training scheme under a student-teacher framework including (1) layer-wise minimizing angular loss by a differentiable binarization search method and (2) layer-wise minimizing amplitude error by learning scale factors. PASCAL-VOC (PASCAL Visual Object Classes)  \citep{everingham2010pascal} and MS-COCO (Microsoft Common Objects in Context) \citep{lin2014microsoft} are popular benchmark datasets to evaluate trained models' performance for object detection. \citet{9494360} develop  the block scaling factor XNOR (BSF-XNOR) convolutional layer and enhanced the accuracy on VisDrone2019 dataset \citep{zhu2018vision} compared to XNOR-Net \citep{rastegari2016xnor}  

PASCAL-VOC is a collection of datasets for object detection, which was used as the competition dataset in the PASCAL Visual Object Classes Challenge Competition. There are two commonly versions: VOC2007 and VOC2012. VOC2007 \citep{pascal-voc-2007} dataset has 5011 training/validation (trainval) images and 4952 test images from 20 categories. VOC2012 \citep{pascal-voc-2012} dataset has 11540 trainval  images including 20 categories. VOC2012 is usually used as an additional data resource for model training. 

MS-COCO is a large-scale object detection, segmentation, key-point detection, and captioning dataset which consists of 80 categories of images. The commonly used version of MS-COCO is MS-COCO2014. Table 20 is the collection of benchmark results in BNN and full-precision CNN, in where  models were trained on 80000 images from the training set and
35000 images sampled from validation set (MS-COCO trainval35k) ASDA-FRCNN \citep{10.1145/3410338.3412340} applies the experience from Modulated-Convolutional-Networks \citep{Wang_2018_CVPR} to represent full-precision kernel with BNN-based amplitude and direction, and designs a new loss function to reconstruct the full-precision kernels. \citet{wang2020bidet} propose BiDet which employs the information bottleneck (IB) principle to remove redundancy information

Tables 23-24 are the latest BNN benchmark results on the VOC2007 test dataset. When we summarize this table, we notice \citep{qin2020binary} wrongly cited FQN \citep{li2019fully} as a BNN variant and collected its results on MS-COCO2017\citep{caesar2018cvpr}. FQN \citep{li2019fully} results were actually collected from a 4-bit quantization neural network. Tables 25-26 are the latest BNN benchmark results on MS-COCO minival test dataset.

%%%%%%%%%%%%%%%%%%%%%%%%%%%%%voc
\begin{table}[!h]
 \begin{threeparttable}
\centering

\begin{tabular}{llllll}

\hlineB{2.5} 
 \rowcolor{Gray}
Framework & BackBone&Method&(W/A) &Trained data& mAP(\%) \\

\hlineB{2.5}

{Faster-RCNN\citeyearpar{long2015fully}}& \cellcolor{LightCyan}VGG-16&\cellcolor{LightCyan}Full Precision&\cellcolor{LightCyan}(32/32)&\cellcolor{LightCyan}VOC2007&\cellcolor{LightCyan}68.9\citeyearpar{sun2018fast}\\

&VGG-16&BNN\citeyearpar{courbariaux2016binarized}&(1/1)&VOC2007&47.3\citeyearpar{sun2018fast}\\

\cline{2-6}
\\[-1em]
\cline{2-6}
\\[-8pt]
&\cellcolor{LightCyan}AlexNet&\cellcolor{LightCyan}Full Precision&\cellcolor{LightCyan}(32/32)&\cellcolor{LightCyan}VOC2007&\cellcolor{LightCyan}66.0\citeyearpar{sun2018fast}\\
&AlexNet&BNN\citeyearpar{courbariaux2016binarized}&(1/1)&VOC2007&46.4\citeyearpar{sun2018fast}\\

\cline{2-6}
\\[-1em]
\cline{2-6}
\\[-8pt]
&\cellcolor{LightCyan}ResNet-18&\cellcolor{LightCyan}Full Precision&\cellcolor{LightCyan}(32/32)&\cellcolor{LightCyan}VOC2007&\cellcolor{LightCyan}67.8\citeyearpar{10.1145/3410338.3412340}\\

&ResNet-18&Full Precision&(32/32)&VOC2007+2012&73.2\citeyearpar{10.1145/3410338.3412340}\\

&\cellcolor{LightCyan}ResNet-18&\cellcolor{LightCyan}Full Precision&\cellcolor{LightCyan}(32/32)&\cellcolor{LightCyan}VOC2007+2012&\cellcolor{LightCyan}74.5\citeyearpar{wang2020bidet}\\

&ResNet-18&Full Precision&(32/32)&VOC2007+2012&76.4\citeyearpar{Xu_2021_CVPR}\\
&\cellcolor{LightCyan}ResNet-18&\cellcolor{LightCyan}BNN\citeyearpar{courbariaux2016binarized}&\cellcolor{LightCyan}(1/1)&\cellcolor{LightCyan}VOC2007+2012&\cellcolor{LightCyan}35.6\citeyearpar{wang2020bidet}\\

&ResNet-18&XNOR-Net\citeyearpar{rastegari2016xnor}&(1/1)&VOC2007+2012&48.4\citeyearpar{wang2020bidet}\\

&\cellcolor{LightCyan}ResNet-18&\cellcolor{LightCyan}Bi-Real\citeyearpar{liu2018bi}&\cellcolor{LightCyan}(1/1)&\cellcolor{LightCyan}VOC2007&\cellcolor{LightCyan}51.0\citeyearpar{10.1145/3410338.3412340}\\

&ResNet-18&Bi-Real\citeyearpar{liu2018bi}&(1/1)&VOC2007+2012&58.2\citeyearpar{wang2020bidet}\\

&\cellcolor{LightCyan}ResNet-18&\cellcolor{LightCyan}Bi-Real\citeyearpar{liu2018bi}&\cellcolor{LightCyan}(1/1)&\cellcolor{LightCyan}VOC2007+2012&\cellcolor{LightCyan}60.6\citeyearpar{10.1145/3410338.3412340}\\

&ResNet-18&Bi-Real\citeyearpar{liu2018bi}&(1/1)&VOC2007+2012&60.9\citeyearpar{Xu_2021_CVPR}\\

&\cellcolor{LightCyan}ResNet-18&\cellcolor{LightCyan}ASDA-FRCNN\citeyearpar{10.1145/3410338.3412340}&\cellcolor{LightCyan}(1/1)&\cellcolor{LightCyan}VOC2007&\cellcolor{LightCyan}54.6\citeyearpar{10.1145/3410338.3412340}\\

&ResNet-18&ASDA-FRCNN\citeyearpar{10.1145/3410338.3412340}&(1/1)&VOC2007+2012&63.4\citeyearpar{10.1145/3410338.3412340}\\

&\cellcolor{LightCyan}ResNet-18&\cellcolor{LightCyan}BiDet\citeyearpar{wang2020bidet}&\cellcolor{LightCyan}(1/1)&\cellcolor{LightCyan}\cellcolor{LightCyan}VOC2007+2012&\cellcolor{LightCyan}50.0\citeyearpar{wang2020bidet}\\

&ResNet-18&BiDet(SC)\footnotemark[1]\citeyearpar{wang2020bidet}&(1/1)&VOC2007+2012&59.5\citeyearpar{wang2020bidet}\\

&\cellcolor{LightCyan}ResNet-18&\cellcolor{LightCyan}BiDet\citeyearpar{wang2020bidet}&\cellcolor{LightCyan}(1/1)&\cellcolor{LightCyan}VOC2007+2012&\cellcolor{LightCyan}62.7\citeyearpar{Xu_2021_CVPR}\\

&   ResNet-18&   DIR-Net\footnotemark[2]\citeyearpar{qin2022distribution}&   (1/1)&   VOC2007+2012&   60.4\citeyearpar{qin2022distribution}\\
&\cellcolor{LightCyan}ResNet-18&\cellcolor{LightCyan}DA-BNN\citeyearpar{zhao2021data}&\cellcolor{LightCyan}(1/1)&\cellcolor{LightCyan}VOC2007+2012&\cellcolor{LightCyan}63.5\citeyearpar{zhao2021data}\\

&ResNet-18&ReActNet\citeyearpar{liu2020reactnet}&(1/1)&VOC2007+2012&69.6\citeyearpar{Xu_2021_CVPR}\\

&\cellcolor{LightCyan}ResNet-18&\cellcolor{LightCyan}LWS-Det\citeyearpar{Xu_2021_CVPR}&\cellcolor{LightCyan}(1/1)&\cellcolor{LightCyan}VOC2007+2012&\cellcolor{LightCyan}73.2\citeyearpar{Xu_2021_CVPR}\\

\cline{2-6}
\\[-1em]
\cline{2-6}
\\[-8pt]
&ResNet-34&Full Precision&(32/32)&VOC2007+2012&73.2\citeyearpar{10.1145/3410338.3412340}\\
&\cellcolor{LightCyan}ResNet-34&\cellcolor{LightCyan}Full Precision&\cellcolor{LightCyan}(32/32)&\cellcolor{LightCyan}VOC2007+2012&\cellcolor{LightCyan}77.8\citeyearpar{Xu_2021_CVPR}\\

&ResNet-34&Bi-Real\citeyearpar{liu2018bi}&(1/1)&VOC2007+2012&63.1\citeyearpar{Xu_2021_CVPR}\\

&\cellcolor{LightCyan}ResNet-34&\cellcolor{LightCyan}ASDA-FRCNN\citeyearpar{10.1145/3410338.3412340}&\cellcolor{LightCyan}(1/1)&\cellcolor{LightCyan}VOC2007+2012&\cellcolor{LightCyan}65.5\citeyearpar{10.1145/3410338.3412340}\\

&ResNet-34&BiDet\citeyearpar{wang2020bidet}&(1/1)&VOC2007+2012&65.8\citeyearpar{Xu_2021_CVPR}\\

&\cellcolor{LightCyan}ResNet-34&\cellcolor{LightCyan}ReActNet\citeyearpar{liu2020reactnet}&\cellcolor{LightCyan}(1/1)&\cellcolor{LightCyan}VOC2007+2012&\cellcolor{LightCyan}72.3\citeyearpar{Xu_2021_CVPR}\\

&ResNet-34&LWS-Det\citeyearpar{Xu_2021_CVPR}&(1/1)&VOC2007+2012&75.8\citeyearpar{Xu_2021_CVPR}\\

\cline{2-6}
\\[-0.9em]
\cline{2-6}
\\[-8pt]
&\cellcolor{LightCyan}ResNet-50&\cellcolor{LightCyan}Full Precision&\cellcolor{LightCyan}(32/32)&\cellcolor{LightCyan}VOC2007+2012&\cellcolor{LightCyan}79.5\citeyearpar{Xu_2021_CVPR}\\
&ResNet-50&Bi-Real\citeyearpar{liu2018bi}&(1/1)&VOC2007+2012&65.7\citeyearpar{Xu_2021_CVPR}\\

&\cellcolor{LightCyan}ResNet-50&\cellcolor{LightCyan}ReActNet\citeyearpar{liu2020reactnet}&\cellcolor{LightCyan}(1/1)&\cellcolor{LightCyan}VOC2007+2012&\cellcolor{LightCyan}73.1\citeyearpar{Xu_2021_CVPR}\\

&ResNet-50&LWS-Det\citeyearpar{Xu_2021_CVPR}&(1/1)&VOC2007+2012&76.9\citeyearpar{Xu_2021_CVPR}\\

\hline

\end{tabular}

 \caption{\label{tab:table-name}Benchmark results on VOC2007 test dataset}

  \begin{tablenotes}
          %\footnotesize   %% If you want them smaller like foot notes
   
            \item Note:  mAP: Mean Average, \textbf{${}^1$}:  BiDet(SC)
means the proposed method with extra shortcut for the architectures
Precision,\textbf{${}^2$}: DIR-Net: journal version of IR-Net{\citep{ qin2020forward}}
        \end{tablenotes}
    \end{threeparttable}
\end{table}

\begin{table}[h!]
 \begin{threeparttable}
\centering

\begin{tabular}{llllll}

\hlineB{2.5} 
  \rowcolor{Gray}
Framework & BackBone&Method&(W/A) &Trained data& mAP(\%) \\
\hlineB{2.5}

{SSD300\citeyearpar{liu2016ssd}}&VGG-16&Full Precision&(32/32)&VOC2007+2012&72.4\citeyearpar{wang2020bidet}\\
&\cellcolor{LightCyan}VGG-16&\cellcolor{LightCyan}Full Precision&\cellcolor{LightCyan}(32/32)&\cellcolor{LightCyan}VOC2007+2012&\cellcolor{LightCyan}74.3\citeyearpar{Xu_2021_CVPR}\\

&VGG-16&BNN\citeyearpar{courbariaux2016binarized}&(1/1)&VOC2007+2012&42.0\citeyearpar{wang2020bidet}\\

&\cellcolor{LightCyan}VGG-16&\cellcolor{LightCyan}XNOR-Net\citeyearpar{rastegari2016xnor}&\cellcolor{LightCyan}(1/1)&\cellcolor{LightCyan}VOC2007+2012&50.2\citeyearpar{wang2020bidet}\\

&VGG-16&Bi-Real\citeyearpar{liu2018bi}&(1/1)&VOC2007+2012&63.8\citeyearpar{wang2020bidet}\\

&\cellcolor{LightCyan}VGG-16&\cellcolor{LightCyan}BiDet\citeyearpar{wang2020bidet}&\cellcolor{LightCyan}(1/1)&\cellcolor{LightCyan}VOC2007+2012&\cellcolor{LightCyan}52.4\citeyearpar{wang2020bidet}\\

&VGG-16&BiDet(SC)\footnotemark[1]\citeyearpar{wang2020bidet}&(1/1)&VOC2007+2012&66.0\citeyearpar{wang2020bidet}\\

&\cellcolor{LightCyan}   VGG-16&\cellcolor{LightCyan}   AutoBiDet\citeyearpar{wang2021learning}&\cellcolor{LightCyan}   (1/1)&\cellcolor{LightCyan}   VOC2007+2012&\cellcolor{LightCyan}   53.5
\citeyearpar{wang2021learning}\\

&   VGG-16&   AutoBiDet(SC)\footnotemark[1]\citeyearpar{wang2021learning}&   (1/1)&   VOC2007+2012&   67.5\citeyearpar{wang2021learning}\\

&\cellcolor{LightCyan}   VGG-16&\cellcolor{LightCyan}   DIR-Net\footnotemark[2]\citeyearpar{qin2022distribution}&\cellcolor{LightCyan}   (1/1)&\cellcolor{LightCyan}   VOC2007+2012&\cellcolor{LightCyan}   67.1\citeyearpar{qin2022distribution}\\

&   VGG-16&   AdaBin\citeyearpar{tu2022adabin}&   (1/1)&   VOC2007+2012&   64.0\citeyearpar{tu2022adabin}\\

&\cellcolor{LightCyan}   VGG-16&\cellcolor{LightCyan}   AdaBin(SC)\footnotemark[1]\citeyearpar{tu2022adabin}&\cellcolor{LightCyan}   (1/1)&\cellcolor{LightCyan}   VOC2007+2012&\cellcolor{LightCyan}   69.4\citeyearpar{tu2022adabin}\\

&VGG-16&ReActNet\citeyearpar{liu2020reactnet}&(1/1)&VOC2007+2012&68.4\citeyearpar{Xu_2021_CVPR}\\

&\cellcolor{LightCyan}VGG-16&\cellcolor{LightCyan}LWS-Det\citeyearpar{Xu_2021_CVPR}&\cellcolor{LightCyan}(1/1)&\cellcolor{LightCyan}VOC2007+2012&\cellcolor{LightCyan}71.4\citeyearpar{Xu_2021_CVPR}\\

&   VGG-16&   TA-BiDet\citeyearpar{pu2022ta}&   (1/1)&   VOC2007+2012&   74.6\citeyearpar{pu2022ta}\\
\cline{2-6}
\\[-1em]
\cline{2-6}
\\[-8pt]

&\cellcolor{LightCyan}MobileNetV1&\cellcolor{LightCyan}Full Precision&\cellcolor{LightCyan}(32/32)&\cellcolor{LightCyan}VOC2007+2012&\cellcolor{LightCyan}68.0\citeyearpar{wang2020bidet}\\
&MobileNetV1&XNOR-Net\citeyearpar{rastegari2016xnor}&(1/1)&VOC2007+2012&48.9\citeyearpar{wang2020bidet}\\

&\cellcolor{LightCyan}MobileNetV1&\cellcolor{LightCyan}BiDet&\cellcolor{LightCyan}(1/1)&\cellcolor{LightCyan}VOC2007+2012&\cellcolor{LightCyan}51.2\citeyearpar{wang2020bidet}\\

\hline
Yolo\citeyearpar{redmon2016you}&VGG-16&Full Precision&(32/32)&VOC2007+2012&66.4\citeyearpar{Bethge_2019_ICCV}\\
\hline
{SSD512\citeyearpar{liu2016ssd}}&\cellcolor{LightCyan}VGG-16&\cellcolor{LightCyan}Full Precision&\cellcolor{LightCyan}(32/32)&\cellcolor{LightCyan}VOC2007+2012&\cellcolor{LightCyan}76.8\citeyearpar{Bethge_2019_ICCV}\\
&BinaryDenseNet37\citeyearpar{Bethge_2019_ICCV}&BinaryDenseNet&(1/1)&VOC2007+2012&66.4\citeyearpar{Bethge_2019_ICCV}\\
&\cellcolor{LightCyan}BinaryDenseNet45\citeyearpar{Bethge_2019_ICCV}&\cellcolor{LightCyan}BinaryDenseNet&\cellcolor{LightCyan}(1/1)&\cellcolor{LightCyan}VOC2007+2012&\cellcolor{LightCyan}68.2\citeyearpar{Bethge_2019_ICCV}\\

\hline

\end{tabular}

 \caption{\label{tab:table-name}Continue Table 23: Benchmark results on VOC2007 test dataset.  }

  \begin{tablenotes}
          %\footnotesize   %% If you want them smaller like foot notes
   
            \item Note:  mAP: Mean Average, \textbf{${}^1$}:  BiDet(SC), 
means the proposed method with extra shortcut for the architectures
Precision,\textbf{${}^2$}: DIR-Net: journal version of IR-Net{\citep{ qin2020forward}}
        \end{tablenotes}
    \end{threeparttable}
\end{table}
%%%%%%%%%%%%%%%%%%%%%%%%%%%%%%%%%%%%%%%%%%%%%%%%%%%%COCO

\begin{table}[!h]
 \begin{threeparttable}
\centering

\begin{tabular}{llllll}
\hlineB{2.5} 
\rowcolor{Gray}

Framework & BackBone&Method&(W/A) & mAP@.5(\%)&mAP@[.5,.95](\%) \\
\hlineB{2.5} 
{Faster-RCNN\citeyearpar{long2015fully}}&\cellcolor{LightCyan}ResNet-18&\cellcolor{LightCyan}Full Precision&\cellcolor{LightCyan}(32/32)&\cellcolor{LightCyan}42.7\citeyearpar{10.1145/3410338.3412340}&\cellcolor{LightCyan}21.9\citeyearpar{10.1145/3410338.3412340}\\
&ResNet-18&Full Precision&(32/32)&44.8\citeyearpar{wang2020bidet}&26.0\citeyearpar{wang2020bidet}\\

&\cellcolor{LightCyan}ResNet-18&\cellcolor{LightCyan}Full Precision&\cellcolor{LightCyan}(32/32)&\cellcolor{LightCyan}53.8\citeyearpar{wang2020bidet}&\cellcolor{LightCyan}32.2\citeyearpar{wang2020bidet}\\

&ResNet-18&ASDA-FRCNN\citeyearpar{10.1145/3410338.3412340}&(1/1)&37.5\citeyearpar{10.1145/3410338.3412340}&19.4\citeyearpar{10.1145/3410338.3412340}\\

&\cellcolor{LightCyan}ResNet-18&\cellcolor{LightCyan}BNN\citeyearpar{courbariaux2016binarized}&\cellcolor{LightCyan}(1/1)&\cellcolor{LightCyan}14.3\citeyearpar{wang2020bidet}&\cellcolor{LightCyan}5.6\citeyearpar{wang2020bidet}\\

&ResNet-18&XNOR-Net\citeyearpar{rastegari2016xnor}&(1/1)&10.4\citeyearpar{wang2020bidet}&21.6\citeyearpar{wang2020bidet}\\

&\cellcolor{LightCyan}ResNet-18&\cellcolor{LightCyan}Bi-Real\citeyearpar{liu2018bi}&\cellcolor{LightCyan}(1/1)&\cellcolor{LightCyan}29.0\citeyearpar{wang2020bidet}&\cellcolor{LightCyan}14.4\citeyearpar{wang2020bidet}\\

&ResNet-18&Bi-Real\citeyearpar{liu2018bi}&(1/1)&33.1\citeyearpar{Xu_2021_CVPR}&17.4\citeyearpar{Xu_2021_CVPR}\\

&\cellcolor{LightCyan}ResNet-18&\cellcolor{LightCyan}BiDet\citeyearpar{wang2020bidet}&\cellcolor{LightCyan}(1/1)&\cellcolor{LightCyan}24.8\citeyearpar{wang2020bidet}&\cellcolor{LightCyan}12.1\citeyearpar{wang2020bidet}\\

&ResNet-18&BiDet(SC)\footnotemark[1]\citeyearpar{wang2020bidet}&(1/1)&31.0\citeyearpar{wang2020bidet}&15.7\citeyearpar{wang2020bidet}\\

&\cellcolor{LightCyan}   ResNet-18&\cellcolor{LightCyan}   DIR-Net\footnotemark[2]\citeyearpar{qin2022distribution}&\cellcolor{LightCyan}   (1/1)&\cellcolor{LightCyan}   31.5\citeyearpar{qin2022distribution}&\cellcolor{LightCyan}   16.1\citeyearpar{qin2022distribution}\\

&ResNet-18&ReActNet\citeyearpar{liu2020reactnet}&(1/1)&38.5\citeyearpar{Xu_2021_CVPR}&21.1\citeyearpar{Xu_2021_CVPR}\\

&\cellcolor{LightCyan}ResNet-18&\cellcolor{LightCyan}LWS-Det\citeyearpar{Xu_2021_CVPR}&\cellcolor{LightCyan}(1/1)&\cellcolor{LightCyan}44.9\citeyearpar{Xu_2021_CVPR}&\cellcolor{LightCyan}26.9\citeyearpar{Xu_2021_CVPR}\\
%\citeyearpar{Xu_2021_CVPR}

\cline{2-6}
\\[-1em]
\cline{2-6}
\\[-8pt]
&ResNet-34&Full Precision&(32/32)&57.6\citeyearpar{Xu_2021_CVPR}&35.8\citeyearpar{Xu_2021_CVPR}\\

&\cellcolor{LightCyan}ResNet-34&\cellcolor{LightCyan}Bi-Real\citeyearpar{liu2018bi}&\cellcolor{LightCyan}(1/1)&\cellcolor{LightCyan}37.1\citeyearpar{Xu_2021_CVPR}&\cellcolor{LightCyan}20.1\citeyearpar{Xu_2021_CVPR}\\

&ResNet-34&BiDet\citeyearpar{wang2020bidet}&(1/1)&41.8\citeyearpar{Xu_2021_CVPR}&21.7\citeyearpar{Xu_2021_CVPR}\\

&\cellcolor{LightCyan}ResNet-34&\cellcolor{LightCyan}ReActNet\citeyearpar{liu2020reactnet}&\cellcolor{LightCyan}(1/1)&\cellcolor{LightCyan}43.3\citeyearpar{Xu_2021_CVPR}&\cellcolor{LightCyan}23.4\citeyearpar{Xu_2021_CVPR}\\

&ResNet-34&LWS-Det\citeyearpar{Xu_2021_CVPR}&(1/1)&49.2\citeyearpar{Xu_2021_CVPR}&29.9\citeyearpar{Xu_2021_CVPR}\\

\cline{2-6}
\\[-1em]
\cline{2-6}
\\[-8pt]
&\cellcolor{LightCyan}ResNet-50&\cellcolor{LightCyan}Full Precision&\cellcolor{LightCyan}(32/32)&\cellcolor{LightCyan}59.3\citeyearpar{Xu_2021_CVPR}&\cellcolor{LightCyan}37.7\citeyearpar{Xu_2021_CVPR}\\

&ResNet-50&Bi-Real\citeyearpar{liu2018bi}&(1/1)&40.0\citeyearpar{Xu_2021_CVPR}&22.9\citeyearpar{Xu_2021_CVPR}\\

&\cellcolor{LightCyan}ResNet-50&\cellcolor{LightCyan}ReActNet\citeyearpar{liu2020reactnet}&\cellcolor{LightCyan}(1/1)&\cellcolor{LightCyan}47.7\citeyearpar{Xu_2021_CVPR}&\cellcolor{LightCyan}26.1\citeyearpar{Xu_2021_CVPR}\\

&ResNet-50&LWS-Det\citeyearpar{Xu_2021_CVPR}&(1/1)&52.1\citeyearpar{Xu_2021_CVPR}&31.7\citeyearpar{Xu_2021_CVPR}\\
\hline

\end{tabular}

 \caption{\label{tab:table-name} Benchmark results on MS-COCO minival. }

  \begin{tablenotes}
          %\footnotesize   %% If you want them smaller like foot notes
   
            \item Note:  mAP: Mean Average (AP), mAP@.5 : mAP for Intersection over Union (IoU) = 0.5, mAP@[.5, .95] : mAP for IoU $\in$ [0.5 : 0.05 : 0.95], \textbf{${}^1$}:  BiDet(SC)
means the proposed method with extra shortcut for the architectures
Precision,\textbf{${}^2$}: DIR-Net: journal version of IR-Net{\citep{ qin2020forward}}
        \end{tablenotes}
    \end{threeparttable}
\end{table}

\begin{table}[!h]
 \begin{threeparttable}
\centering

\begin{tabular}{llllll}

\hlineB{2.5} 
  \rowcolor{Gray}
Framework & BackBone&Method&(W/A) & mAP@.5(\%)&mAP@[.5,.95](\%) \\
\hlineB{2.5}

{SSD300\citeyearpar{liu2016ssd}}&VGG-16&Full Precision&(32/32)&41.2\citeyearpar{wang2020bidet}&23.2\citeyearpar{wang2020bidet}\\

&\cellcolor{LightCyan}VGG-16&\cellcolor{LightCyan}BNN\citeyearpar{courbariaux2016binarized}&\cellcolor{LightCyan}(1/1)&\cellcolor{LightCyan}15.9\citeyearpar{wang2020bidet}&\cellcolor{LightCyan}6.2\citeyearpar{wang2020bidet}\\

&VGG-16&XNOR-Net\citeyearpar{rastegari2016xnor}&(1/1)&19.5\citeyearpar{wang2020bidet}&8.1\citeyearpar{wang2020bidet}\\

&\cellcolor{LightCyan}VGG-16&\cellcolor{LightCyan}Bi-Real\citeyearpar{liu2018bi}&\cellcolor{LightCyan}(1/1)&\cellcolor{LightCyan}26.0\citeyearpar{wang2020bidet}&\cellcolor{LightCyan}11.2\citeyearpar{wang2020bidet}\\

&VGG-16&BiDet\citeyearpar{wang2020bidet}&(1/1)&22.5\citeyearpar{wang2020bidet}&9.8\citeyearpar{wang2020bidet}\\

&\cellcolor{LightCyan}VGG-16&\cellcolor{LightCyan}BiDet(SC)\footnotemark[1]\citeyearpar{wang2020bidet}&\cellcolor{LightCyan}(1/1)&\cellcolor{LightCyan}28.3\citeyearpar{wang2020bidet}&\cellcolor{LightCyan}13.2\citeyearpar{wang2020bidet}\\

&   VGG-16&   DIR-Net\footnotemark[2]\citeyearpar{qin2022distribution}&   (1/1)&   29.7\citeyearpar{qin2022distribution}&   14.0\citeyearpar{qin2022distribution}\\

&\cellcolor{LightCyan}   VGG-16&\cellcolor{LightCyan}   AutoBiDet\citeyearpar{wang2021learning}&\cellcolor{LightCyan}   (1/1)&\cellcolor{LightCyan}   30.3\citeyearpar{wang2021learning}&\cellcolor{LightCyan}   14.3\citeyearpar{wang2021learning}\\

&ResNet-50&ReActNet\citeyearpar{liu2020reactnet}&(1/1)&30.0\citeyearpar{Xu_2021_CVPR}&15.3\citeyearpar{Xu_2021_CVPR}\\

&\cellcolor{LightCyan}ResNet-50&\cellcolor{LightCyan}LWS-Det\citeyearpar{Xu_2021_CVPR}&\cellcolor{LightCyan}(1/1)&\cellcolor{LightCyan}32.9\citeyearpar{Xu_2021_CVPR}&\cellcolor{LightCyan}17.1\citeyearpar{Xu_2021_CVPR}\\

&   ResNet-50&   TA-BiDet\citeyearpar{pu2022ta}&   (1/1)&   37.2\citeyearpar{pu2022ta}&   19.9\citeyearpar{pu2022ta}\\

\hline 

\end{tabular}

 \caption{\label{tab:table-name} Continue Table 25: Benchmark results on MS-COCO minival. }

  \begin{tablenotes}
          %\footnotesize   %% If you want them smaller like foot notes
   
            \item Note:  mAP: Mean Average (AP), mAP@.5: mAP for Intersection over Union (IoU) = 0.5, mAP@[.5, .95] : mAP for IoU $\in$ [0.5 : 0.05 : 0.95], \textbf{${}^1$}:  BiDet(SC)
means the proposed method with extra shortcut for the architectures
Precision,\textbf{${}^2$}: DIR-Net: journal version of IR-Net{\citep{ qin2020forward}}
        \end{tablenotes}
    \end{threeparttable}
\end{table}

\newpage
\subsection{Semantic Segmentation}
Group-Net \citep{zhuang2019structured} reports their method can be successfully applied to Semantic Segmentation task. The author proposes Binary Parallel Atrous Convolution (BPAC) to further improve the BNN model performance mIOU which is measured regarding averaged pixel
intersection-over-union. Their used dataset for testing semantic segmentation is PASCAL VOC 2012 \citep{everingham2010pascal}, which contains 20 foreground object classes and one background class. The dataset cconsists of 1464 (train), 1449
(val) and 1456 (test) images.   Table 27 lists the latest BNN benchmark results on PASCAL VOC 2012 test dataset.\\
\begin{table}[h!]

\begin{tabular}{llll}

\hlineB{2.5} 
  \rowcolor{Gray}
Based Model & BackBone&Method/(W/A) & mIOU \\
\hlineB{2.5}

  \rowcolor{LightCyan}
 Faster-RCNN\citeyearpar{long2015fully}, FCN-16s&{ResNet-18}&Full Precision(32/32)&67.3\\
  \rowcolor{LightCyan}
&&Group-Net(5 x 1/1)\citeyearpar{zhuang2019structured}&62.7\\
 \rowcolor{LightCyan}
&&Group-Net + BPAC(5 x 1/1)\citeyearpar{zhuang2019structured}&66.3\\
 \rowcolor{LightCyan}
&ResNet-18**\citeyearpar{liu2018bi}&Group-Net + BPAC(5 x 1/1)\citeyearpar{zhuang2019structured}&67.7\\

 Faster-RCNN\citeyearpar{long2015fully}, FCN-32s&{ResNet-18}&Full Precision(32/32)&64.9\\
&&Group-Net(5 x 1/1)\citeyearpar{zhuang2019structured}&60.5\\
&&Group-Net + BPAC(5 x 1/1)\citeyearpar{zhuang2019structured}&63.8\\
&ResNet-18**\citeyearpar{liu2018bi}&Group-Net + BPAC(5 x 1/1)\citeyearpar{zhuang2019structured}&65.1\\
 \rowcolor{LightCyan}
Faster-RCNN\citeyearpar{long2015fully}, FCN-32s&{ResNet-34}&Full Precision(32/32)&72.7\\
 \rowcolor{LightCyan}
&&Group-Net(5 x 1/1)\citeyearpar{zhuang2019structured}&68.2\\
 \rowcolor{LightCyan}
&&Group-Net + BPAC(5 x 1/1)\citeyearpar{zhuang2019structured}&71.2\\
 \rowcolor{LightCyan}
&ResNet-34**\citeyearpar{liu2018bi}&Group-Net + BPAC(5 x 1/1)\citeyearpar{zhuang2019structured}&72.8\\

 Faster-RCNN\citeyearpar{long2015fully} , FCN-32s&{ResNet-50}&Full Precision(32/32)&73.1\\
&&Group-Net(5 x 1/1)\citeyearpar{zhuang2019structured}&67.2\\
&&Group-Net + BPAC(5 x 1/1)\citeyearpar{zhuang2019structured}&70.4\\

\hline
\end{tabular}

 \caption{\label{tab:table-name}PASCAL VOC 2012 testing results }
 \vspace{-0.5cm}
\end{table}

\subsection{Natural Language Processing}}

\noindent Natural Language Processing (NLP) is one of the most important applied fields of AI. BNN can be applied to promote the applications of NLP model on edge devices in the real
world. \citet{jain2020end} explore and proposes a BNN method on text classification. \citet{baibinarybert} firstly apply BNN methods to BERT \citeyearpar{devlin2018bert} on NLP tasks. \citet{qinbibert} firstly propose a full binarization of BERT, called BiBERT, on NLP tasks.  \citet{liubit} propose an innovative pipeline to quantize transformers for extremely low precision (1-2) bits, while reducing the performance gap of previous methods to full precision.GLUE \citeyearpar{wangglue} is a popular benchmark with diverse NLP tasks. Table 28 lists the information for NLP task corpus. Table 29 presents the last BNN benchmark results on GLUE.

\clearpage

\begin{table}[h!]

\begin{tabular}{lllll}

\hlineB{2.5} 
  \rowcolor{Gray}
  \multicolumn{5}{c}{   Sigle-Sentence Tasks}\\
  \rowcolor{Gray}
   Corpus \;  \;\;\;\;\;\;   &    Train Data Size&   Test Data Size &    Task &   Metrics\\
\hlineB{2.5}

   CoLA &    8.5k&   1k&   acceptability &   Matthews correlation \\
   SST-2&   67k&   1.8k&   sentiment&    accuracy\\

\hline
\\[-1em]
\hlineB{2.5} 
  \rowcolor{Gray}
  \multicolumn{5}{c}{   Similarity and Paraphrase Tasks}\\
  \rowcolor{Gray}
   Corpus &    Train Data Size&   Test Data Size &    Task &   Metrics\\
\hlineB{2.5} 
   MRPC &    3.7k&   1.7k&   paraphrase&   accuracy/F1\\
   STS-B &    7k&   1.4k&   sentence similarity&   Pearson/Spearman correlation\\
   QQP &   364k&   391k&   paraphrase&   accuracy/F1\\

\hline
\\[-1em]
\hlineB{2.5} 
  \rowcolor{Gray}
  \multicolumn{5}{c}{   Inference Tasks }\\
  \rowcolor{Gray}
   Corpus &    Train Data Size&   Test Data Size &    Task &   Metrics\\
\hlineB{2.5} 
   MNLI &    393k&   20k&   NLI&   match/mismatched accuracy\\
   QNLI &    105k&   5.4k&   QA/NLI&   accuracy\\
   RTE &   2.5k&   3k&   NLI&   accuracy\\
   WNLI &   634&   146&   conference/NLI&   accuracy\\

\hline
\end{tabular}

 \caption{\label{tab:table-name}   Diverse NLP tasks on GLUE Benchmark}
   \begin{tablenotes}
          %\footnotesize   %% If you want them smaller like foot notes
   
            \item    Note:  CoLA: Corpus of Linguistic Acceptability,  SST-2: Stanford Sentiment Treebank, 
 MRPC: Microsoft Research Paraphrase Corpus,  STS-B: Semantic Textual Similarity Benchmark,  QQP: Quora Question Pairs,  MNLI: Multi-Genre Natural Language Inference, QNLI: Question Natural Language Inference,  RTE: Recognizing Textual Entailment,  WNLI: Winograd Natural Language Inference, QA/NLI: Question and Answer/ Natural Language Inference 
        \end{tablenotes}
 \vspace{-0.5cm}
\end{table}

\begin{table}[h!]

\begin{tabular}{l@{\hskip 0.1in}l@{\hskip 0.1in}l@{\hskip 0.1in}l@{\hskip 0.1in}l@{\hskip 0.1in}l@{\hskip 0.1in}l@{\hskip 0.1in}l@{\hskip 0.1in}l@{\hskip 0.1in}l@{\hskip 0.1in}l@{\hskip 0.1in}l@{\hskip 0.1in}l@{\hskip 0.1in}l}

\hlineB{2.5} 

\rowcolor{Gray}
   \tiny Method &   \tiny\#Bits (E-W-A) &   \tiny Size\tiny(MB)&   \tiny FLOPs\tiny(G)&   \tiny MNLI\tiny(m/mm)&   \tiny QQP&   \tiny QNLI&   \tiny SST-2&   \tiny CoLA&   \tiny STS-B&   \tiny MRPC&   \tiny RTE&   \tiny Avg.\\
\hlineB{2.5} 

   \tiny BERT\citeyearpar{devlin2018bert}&   \tiny 32-32-32&   \tiny418&   \tiny22.5 &   \tiny 84.9/85.5 &   \tiny 91.4&   \tiny 92.1&   \tiny93.2&   \tiny 59.7 &   \tiny 90.1&   \tiny 86.3 &   \tiny 72.2&   \tiny 83.9\\

\hline
\\[-1.2em]
\hline
 \multicolumn{13}{c}{   Without data augmentation}\\
\hline
 \rowcolor{LightCyan}
   \tiny BinaryBERT\citeyearpar{baibinarybert}&   \tiny 1-1-1&   \tiny 16.5&   \tiny 0.4&   \tiny 35.6/35.3&   \tiny 66.2&   \tiny 51.5&   \tiny 53.2&   \tiny 0&   \tiny 6.1 &   \tiny68.3&   \tiny 52.7 &   \tiny41.0\\
   \tiny BiBERT\citeyearpar{qinbibert}&   \tiny 1-1-1&   \tiny 13.4&   \tiny 0.4&   \tiny 66.1/67.5&   \tiny 84.8&   \tiny 72.6&   \tiny 88.7&   \tiny 25.4&   \tiny 33.6&   \tiny 72.5&   \tiny 57.4 &   \tiny63.2\\
 \rowcolor{LightCyan}
   \tiny BiT\ddag\citeyearpar{liubit}&   \tiny1-1-1&   \tiny 13.4&   \tiny 0.4&   \tiny 77.1/77.5&   \tiny 82.9&   \tiny 85.7&   \tiny 87.7&   \tiny 25.1&   \tiny 71.1&   \tiny 79.7&   \tiny 58.8&   \tiny 71.0\\
   \tiny BiT\citeyearpar{liubit}&   \tiny 1-1-1&   \tiny 13.4&   \tiny 0.4 &   \tiny79.5/79.4&   \tiny 85.4&   \tiny 86.4&   \tiny 89.9&   \tiny 32.9&   \tiny 72.0&   \tiny 79.9&   \tiny 62.1&   \tiny 73.5\\

\hline
\\[-1.2em]
\hline
 \multicolumn{13}{c}{   With data augmentation}\\
\hline
 \rowcolor{LightCyan}

   \tiny BinaryBERT\citeyearpar{baibinarybert}&   \tiny 1-1-1&   \tiny 16.5&   \tiny 0.4&   \tiny 35.6/35.3*&   \tiny 66.2* &   \tiny66.1&   \tiny 78.3&   \tiny 7.3&   \tiny 22.1&   \tiny 69.3&   \tiny 57.7&   \tiny 48.7\\

   \tiny BiBERT\citeyearpar{qinbibert}&   \tiny 1-1-1&   \tiny 13.4&   \tiny 0.4&   \tiny 66.1/67.5*&   \tiny 84.8*&   \tiny 76.0 &   \tiny90.9 &   \tiny37.8&   \tiny 56.7&   \tiny 78.8&   \tiny 61.0&   \tiny 68.8\\

 \rowcolor{LightCyan}

   \tiny BiT\ddag\citeyearpar{liubit}&   \tiny1-1-1&   \tiny 13.4&   \tiny 0.4&   \tiny 77.1/77.5*&   \tiny 82.9* &   \tiny85.0&   \tiny 91.5&   \tiny 32.0 &   \tiny84.1&   \tiny 88.0&   \tiny 67.5&   \tiny 76.0\\
   \tiny BiT\citeyearpar{liubit}&   \tiny 1-1-1&   \tiny 13.4&   \tiny 0.4 &   \tiny79.5/79.4*&   \tiny 85.4* &   \tiny86.5&   \tiny 92.3&   \tiny 38.2&   \tiny 84.2 &   \tiny88.0&   \tiny 69.7&   \tiny 78.0\\

\hline
\end{tabular}

 \caption{\label{tab:table-name}   BNN Bert benchmark results on NLP tasks of GLUE}

  \begin{minipage}{15.5cm}%
       Note:  E-W-A: the quantization level of embeddings, weights and activations, m/mm: match/mismatch,  \ddag: distilling binary models
using full-precision teacher without using multi-distill technique\ddag\citeyearpar{liubit}, 
*: Data augmentation is not needed for MNLI and QQP.%
  \end{minipage}%

 \vspace{-1cm}
\end{table}

\subsection{Unsupervised and Semi-supervised Learning
}

 \noindent Unsupervised learning is one of the most popular cutting edge research topics in machine learning where the model is not given labeled data to learn from, but instead, it must find patterns and structure in the data on its own. Self-supervised learning(SSL) is one subfield of unsupervised learning where the model is trained to predict information from the input data itself in a first stage, which are then used for some supervised learning task in the second and later stages. Semi-supervised learning is a machine learning technique where the model is trained on a mixture of labeled and unlabeled data. 
 There are a few BNN published research works on SSL and semi-supervised areas. \citet{shen2021s2} propose a novel guided learning paradigm, called s2-BNN, to distill the target binary network training 
from real-valued. \citet{kim2022unsupervised} design a framework, called BURN, that jointly trains
the Full-Precision classifier and the binary network for the unsupervised representation
learning as well as a
feature similarity loss, a dynamic loss balancing and modified multi-stage training to further improve the accuracy. Table 30-31 list the last benchmark results for unsupervised and semi-supervised learning.
 \\

\clearpage

\begin{table}[!h]
 {
\begin{threeparttable}

 \caption{\label{tab:table-name}   Bechmark results of linear evaluation and
semi-supervised fine-tuning on ImageNet}
\begin{tabular}{lllllllll}
\hlineB{2.5}

{\multirow{4}{*}{   Method}}\;\;\;\;\;\;&&\;\;\;\;\;\;{\multirow{2}{*}{   Linear Evaluation}} \;\;\;\;\;\;&&
  \multicolumn{5}{c}{   \;\;\;\;\;\; Semi-Supervised Fine-tuning\;\;\;\;\;\;}\\
  
&&&&\multicolumn{2}{c}{   1\% Labels}&&\multicolumn{2}{c}{\;\;\;\;\;   10\% Labels}\\
\cline{3-3}\cline{5-9}\\

&&\;\;\;\;\;\;\;\;\;\;\;\;   Top-1 (\%)  &&   Top-1 (\%)&    Top-5 (\%)& &   Top-1 (\%) &    Top-5 (\%)\\
  \hline
  
   Supervised Pre-training&&\;\;\;\;\;\;\;\;\;\;\;\;    64.10&&    42.96&    69.10 &&   53.07&    77.40\\
 \hline

 \rowcolor{LightCyan}
     S2-BNN\citeyearpar{shen2021s2}&&\;\;\;\;\;\;\;\;\;\;\;\;    61.50&&    36.08&    61.83&&    45.98 &   71.11\\

   BURN\citeyearpar{kim2022unsupervised} &&\;\;\;\;\;\;\;\;\;\;\;\;    62.29&&    39.75 &   67.13&&    49.96&    75.52\\

    \hline

\end{tabular}
 \begin{tablenotes}   
          %\footnotesize   %% If you want them smaller like foot notes
         \item Note:  Linear evaluation (top-1) and semi-supervised fine-tuning (1\% labels or 10\% labels) on ImageNet after pretraining.
        \end{tablenotes}
    \end{threeparttable}
}
\end{table}

\begin{table}[!h]
 {
\begin{threeparttable}

 \caption{\label{tab:table-name}   Bechmark results of  Transfer learning on object-centric and scene-centric dataset}
\begin{tabular}{llllllll}
\hlineB{2.5}

   {\multirow{2}{*}{Method}}\;\;\;\;\;\;\;\;\;&&   \;\;\;\;\;\;Scene-Centric \;\;\;\;\;\;&&
  \multicolumn{4}{c}{   \;\;\;\;\;\;Object-Centric\;\;\;\;\;\;}\\
  
&&   \;\;\;\;\;\;\;\;\;Places205
&&   CIFAR10&    CIFAR100 &   CUB-200-2011&    Birdsnap\\
  \hline
     Supervised Pre-training&&\;\;\;\;\;\;\;\;\;\;\;\;    46.38 &&    78.30&    57.82 &   54.64&    36.90\\

 \hline

 \rowcolor{LightCyan}
     S2-BNN\citeyearpar{shen2021s2}&&\;\;\;\;\;\;\;\;\;\;\;\;    46.58&&    82.70 &   61.90&    47.50&    34.10\\

   BURN\citeyearpar{kim2022unsupervised} &&\;\;\;\;\;\;\;\;\;\;\;\;    47.22&&    84.60&    61.99&    49.62&    34.48\\

    \hline

\end{tabular}
 \begin{tablenotes}
          %\footnotesize   %% If you want them smaller like foot notes
            \item Note: Transfer learning (top-1) on either object-centric or scene-centric datasets after pretraining. CIFAR10, CIFAR100, CUB-200-2011, and Birdsnap are used as the object-centric datasets while Places205 is used as the scene-centric dataset.
        \end{tablenotes}
    \end{threeparttable}
}
\end{table}

\subsection{Other Tasks}

There are other tasks that take the capacities of BNN. \citet{bulat2017binarized} propose a new BNN network for human pose estimation and face alignment. \citet{fasfous2021binarycop} present BinaryCoP which is a BNN  classifier for correct facial-mask wear and positioning on edge devices. To speed-up large-scale image retrieval search with low storage cost, 
\citet{zhang2021binary} propose a novelty hashing method called Binary Neural Network Hashing (BNNH) which combines BNN with hashing technique.\citet{li2021efficient} design a 3D BNN to recognize human actions. Their BNN can achieve 89.2\% accuracy with 384 frames per second on the dataset KTH \citep{schuldt2004recognizing}.   \citet{penkovsky2020memory} propose methods to apply BNN to biomedical signals tasks such as electrocardiography (ECG) and electroencephalography (EEG), which can enable smart autonomous healthcare devices.  \citep{xiang2017binary}, \citep{qian2019binary} and \citep{9516334} are different BNN methods on speech recognition. BiFSMN \citeyearpar{qin2022bifsmn} and its improved version BiFSMN2 \citeyearpar{qin2022bifsmnv2} present the BNN applications on keyword spotting. \citet{bahri2021binary} explore the field of designing BNN-based graph neural networks (GNNs) by evaluating different strategies for the binarization of GNNs.
\citet{frickenstein2020binary} propose Binary DAD-Net which was the first BNN-based semantic segmentation network on driveable area detection in the field of autonomous driving.
%\citet{bahri2020binarygnn} investigate different schemes to binary graph neural network and proposes  \\

 \subsection{Summary}

Although BNN has many successful applications, a few potential opportunities and challenges remain an open issue.

For a given application, what binary neural network architecture should be used? How automatically search architecture or create a 2D or 3D BNN network with higher accuracy and lower OPs. In general, all the layers (except the input and output layers) of a BNN are binarized CNN layers, which are one of the primary sources of losing information. This situation will be more difficult in the deeper layer because the performance drop is accumulated from the previous layers. Also, the information-lose level of different layers inside BNN should not be the same, and we should not be able to treat the information-loss issue equally. Besides, compared to information loss from binarization in 2D tasks, more severe information- loss from binarization can be generated in 3D tasks cases due to more dimensional information availability. Currently, there are few papers for solving 3D tasks.

How to develop transformer-based BNN models for vision tasks? In the past, deep neural networks mostly used costly convolution operations. Moreover, it is the reason and motivation to create the BNN that lowers expansive computing costs in convolution operations. Late last year, (Dosovitskiy et al., 2020) introduced a novel network structure called vision transformer (ViT). The idea for ViT came from the concept of transformer developed in natural language processing (NLP) applications. Instead of using the convolution operations, ViT splits an input image into custom fixed-size patches and feeds the linear projections of these patches along with their image position into a transformer encoder network. More and more transformer-based or a combination of transformer and convolution variants models were published. They reported that their performance could beat CNN-based models with the same weight size. Therefore, how to effectively develop a transformer-based BNN can be a new challenge and a hot research opportunity.

\section{Conclusions}

As mentioned above, since 2016, BNN techniques have drawn increasing research interest because of their capability to deploy models on resource-limited devices. BNN can significantly reduce storage, network complexity and energy consumption to make neural networks more efficient in embedded settings.  However, binarization unavoidably causes a significant performance drop. In this paper, the literature on BNN techniques has been rigorously explored and discussed. For the first time, we solely focus on reviewing mainly 1-bit activations and weights networks that decrease the network memory usage and computational cost. 

Furthermore, a comparative classification of these techniques has been performed and discussed under multiple network components: quantization function, activations/weights distribution, loss function improvement, gradient approximation,  network topology structure, and training strategy and tricks. Additionally, we present popular efficient platforms for BNN and investigate current BNN applications progress. Also, we discussed and identified the research gap and the best methods available in the literature review. Finally, we provide several recommendations and research directions for future exploration. We firmly believe that such an intricate field of BNN is just starting to permeate a broad range of artificial intelligence communities and tiny resource-constraint systems and will soon be taught to students and professionals as an essential topic in computer vision and deep learning.

%addons 638 springer
%addons 450 under review

%122 \runningtitle{\textit{}} % Example Article
%addons 572 \def\copyright@text{\textcopyright\ %Springer \textbullet\textbullet\textbullet\textbullet}

%page number

%%%%%%%%%%%%%%%%%%%%%%%%%%%%%%%%%%%%%%%%%%%%%%%%%%%%%%%%%%%%%%%%%%%%%%%%%%%

%%% BIBLIOGRAPHY %%%%%%%%%%%%%%%%%%%%%%%%%%%%%%%%%%%%%%%%%%%%%%%%%%%%%%%%%%%

     % format of references provided by the journal(.bst)
     \newcommand{\bibcommenthead}{}
\bibliographystyle{sn-chicago}
     % name your Bibtex file containing your references(.bib)
     
      {
\bibliography{sola_bibliography_example}  
}

     % Checking: look if the file containing the ``\bibitem'' exits
     %           so check if the .bbl file exist(bibTeX compilation)
\IfFileExists{\jobname.bbl}{} {\typeout{}
\typeout{****************************************************}
\typeout{****************************************************}
\typeout{** Please run "bibtex \jobname" to obtain} \typeout{**
the bibliography and then re-run LaTeX} \typeout{** twice to fix
the references !}
\typeout{****************************************************}
\typeout{****************************************************}
\typeout{}

}

\end{article}

\end{document}